\documentclass[ 3p]{elsarticle}
\pdfoutput=1
\usepackage{ccicons}
\usepackage{amsmath}
\usepackage{amsthm}
\usepackage[dvipsnames]{xcolor}
\usepackage[hidelinks]{hyperref}

\usepackage{graphicx}
\usepackage{tikz}
\usepackage{pgfplots}
\usepgfplotslibrary{statistics}

\usepackage{amsmath,nccmath,bm}
\usepackage{mathrsfs}
\usepackage{amsfonts}
\usepackage{amssymb}
\usepackage[multiple]{footmisc}
\usepackage{algorithm,algorithmicx}
\usepackage[noend]{algpseudocode}
\newcommand*\Let[2]{\State #1 $\gets$ #2}
\algrenewcommand\algorithmicrequire{\textbf{Precondition:}}
\algrenewcommand\algorithmicensure{\textbf{Postcondition:}}

\newenvironment{myexpcont}
{\begin{myexp}\hspace{-0.5em}{\it \arabic{running}.}\hspace{0.5em}}
  {{\unskip\nobreak\hfil\penalty50\hskip1em\null\nobreak\hfil$\lozenge$\parfillskip=0pt\finalhyphendemerits=0\endgraf}\end{myexp}\addtocounter{myexp}{-1}\addtocounter{running}{1}}

\usepackage{mathtools}

\DeclareMathOperator{\posi}{posi}
\DeclareMathOperator{\E}{E}
\DeclareMathOperator{\cone}{cone}

\newcommand{\R}{\mathbb{R}}

\newcommand{\N}{\mathbb{N}}
\newcommand{\V}{\mathscr{V}}
\newcommand{\Q}{\mathscr{Q}}

\newcommand{\X}{\mathcal{X}}

\newcommand{\precc}{<}
\newcommand{\precceq}{\le}

\newcommand{\OO}{\mathcal{O}}
\newcommand{\OOO}{\mathbb{O}}
\newcommand{\Ass}{\mathcal{A}}
\newcommand{\A}{\mathcal{G}_{\Ass}}

\DeclarePairedDelimiterX\cset[2]{\{}{\}}{\,#1 \;\colon #2\,}

\usepackage[capitalise,noabbrev]{cleveref}

\newenvironment{proofof}[1]{\par\noindent{\bfseries\upshape Proof of #1\ }}{\qed\medskip}

\usepackage{enumitem}
\newlist{axiom}{enumerate}{2}     
\crefname{axiomi}{Axiom}{Axioms} 
\crefname{axiomii}{Axiom}{Axioms} 

\def\maxenumi{0}
\def\saveenumi{0}

\makeatletter
\AtEndDocument{%
  \immediate\write\@auxout{%
    \string\gdef\string\maxenumi{\saveenumi}}}
\makeatother
\setlist[axiom,1]{%
  leftmargin=*,
  widest=\maxenumi,
  after=\ifnum\value{enumi}>\saveenumi\xdef\saveenumi{\the\value{enumi}}\fi}

  \newlist{property}{enumerate}{2}     
\crefname{propertyi}{Property}{Properties} 
\crefname{propertyii}{Property}{Properties} 

\setlist[property,1]{%
  leftmargin=*,
  widest=\maxenumi,
  after=\ifnum\value{enumi}>\saveenumi\xdef\saveenumi{\the\value{enumi}}\fi}

\newcommand{\EE}{\mathrm{E}}

\newcommand{\itrr}{m}

\newcommand{\fromto}[2][1]{#1\kern-0.1em\colon\kern-0.2em #2}

\newcommand{\Echoice}[1][\Ass]{\smash{C^{\EE}_{\mathcal{P}(\Ass)}}}

\makeatletter
\newcommand{\labeltext}[3][]{%
    \@bsphack%
    \csname phantomsection\endcsname
    \def\tst{#1}%
    \def\labelmarkup{\emph}
    \def\refmarkup{}%
    \ifx\tst\empty\def\@currentlabel{\refmarkup{#2}}{\label{#3}}%
    \else\def\@currentlabel{\refmarkup{#1}}{\label{#3}}\fi%
    \@esphack%
    \labelmarkup{#2}
}
\makeatother

\newcommand{\Eu}{\underline{\E}}

\usepackage{thmtools}
\usepackage{hyperref}
\theoremstyle{definition}
\newtheorem{theorem}{Theorem}[section]
\newtheorem{example}[theorem]{Example}
\newtheorem{definition}[theorem]{Definition}
\newtheorem{proposition}[theorem]{Proposition}
\newtheorem{corollary}[theorem]{Corollary}
\newtheorem{lemma}[theorem]{Lemma}
\newtheorem{myexp}[theorem]{Running Example}

\newcommand{\cl}{\textrm{cl}}
\newcommand{\NatExt}{C_{\Ass}}
\newcommand{\D}{\mathcal{H}}
\newcommand{\G}{\mathcal{G}}
\newcommand{\HH}{\mathcal{H}}
\newcommand{\NN}{\mathcal{N}}
\newcommand{\Qe}{\Q_{\emptyset}}
\newcommand{\isF}{\textsc{IsFeasible}}

\renewcommand{\algorithmicrequire}{\textbf{Input:}}
\renewcommand{\algorithmicensure}{\textbf{Output:}}
\newcommand{\myTikzMark}[1]{\tikz[baseline=-0.9ex]{\node[mark size=0.7ex]{\pgfuseplotmark{#1}};}}
\newcommand{\colorMark}[3]{\tikz[baseline=-0.9ex]{\node[mark size=0.7ex,mark options={#2,solid}](B){\pgfuseplotmark{#1}};\draw[dash pattern=on #3pt off 1pt]([xshift=-0.15cm]B.west)--([xshift=0.15cm]B.east);}}
\definecolor{LimeGreen}{rgb}{0.55, 0.71, 0.0}
\definecolor{indigo}{rgb}{0.29, 0.0, 0.51}
\theoremstyle{definition}
\newcommand{\aantaloptionsetsexp}{10 }
\renewcommand{\Eu}{\underline{\mathrm{E}}}
\renewcommand{\E}{\mathrm{E}}

\crefalias{statement}{enumi}
\crefname{statement}{statement}{statements}
\Crefname{statement}{statement}{statements}

\usepackage{booktabs}

\crefname{myexp}{Running Example}{Running Examples}
\usepackage{caption}

\pgfplotsset{select coords between index/.style 2 args={
    x filter/.code={
        \ifnum\coordindex<#1\fi
        \ifnum\coordindex>#2\fi
    }
}}

\newcommand{\OOOO}{\mathbf{O}}

\usepackage{chngcntr}

\journal{International Journal of Approximate Reasoning}

\makeatletter
\newcommand*{\algrule}[1][\algorithmicindent]{%
  \makebox[#1][l]{%
    \hspace*{.2em}
    \vrule height .75\baselineskip depth .25\baselineskip
  }
}

\newcount\ALG@printindent@tempcnta
\def\ALG@printindent{%
    \ifnum \theALG@nested>0
    \ifx\ALG@text\ALG@x@notext
    \else
    \unskip
    \ALG@printindent@tempcnta=1
    \loop
    \algrule[\csname ALG@ind@\the\ALG@printindent@tempcnta\endcsname]%
    \advance \ALG@printindent@tempcnta 1
    \ifnum \ALG@printindent@tempcnta<\numexpr\theALG@nested+1\relax
    \repeat
    \fi
    \fi
}
\patchcmd{\ALG@doentity}{\noindent\hskip\ALG@tlm}{\ALG@printindent}{}{\errmessage{failed to patch}}
\patchcmd{\ALG@doentity}{\item[]\nointerlineskip}{}{}{} 
\makeatother
\usepackage{subcaption}
\pgfplotsset{compat=1.18}
\usepackage[dont-mess-around]{fnpct}
\let\oldFootnote\footnote
\newcommand\nextToken\relax
\renewcommand\footnote[1]{%
    \oldFootnote{#1}\futurelet\nextToken\isFootnote}
\newcommand\isFootnote{%
    \ifx\footnote\nextToken\textsuperscript{,}\fi}

\begin{document}

\begin{frontmatter}

\title{Extending choice assessments to choice functions: An algorithm for computing the natural extension}

\author{Arne Decadt\corref{mycorrespondingauthor}}
\ead{arne.decadt@ugent.be}
\cortext[mycorrespondingauthor]{Corresponding authors}
\author{Alexander Erreygers}
\author{Jasper De Bock\corref{mycorrespondingauthor}}
\ead{jasper.debock@ugent.be}
\address{Ghent University, Belgium}
\tnotetext[t1]{\ccCopy~2024. This manuscript version is made available under the \href{https://creativecommons.org/licenses/by-nc-nd/4.0/}{CC-BY-NC-ND 4.0} \ccbyncnd~license}



%


\begin{abstract}
We study how to infer new choices from prior choices using the framework of choice functions, a unifying mathematical framework for decision-making based on sets of preference orders.
In particular, we define the natural (most conservative) extension of a given choice assessment to a coherent choice function---whenever possible---and use this natural extension to make new choices.
We provide a practical algorithm for computing this natural extension and various ways to improve scalability.
Finally, we test these algorithms for different types of choice assessments.
\end{abstract}


\end{frontmatter}


\section{Introduction}
\label{sec:intro}
In classical probability theory, decisions are typically made by maximising expected utility.
This leads to a single optimal decision, or a set of optimal decisions all of which are equivalent.
In the theory of imprecise probabilities, where multiple probabilistic models are considered simultaneously, this decision rule can be generalised in multiple ways;
Troffaes \cite{troffaes2007decision} provides a nice overview.
A typical feature of the resulting decision rules is that they will not always yield a single optimal decision, as a decision that is optimal in one probability model may for example be suboptimal in another.

We here take this generalisation yet another step further by adopting the theory of choice functions: a mathematical framework for decision-making that incorporates several (imprecise) decision rules as special cases, including the classical approach of maximising expected utility \cite{de2020archimedean,de2019interpreting,seidenfeld2010coherent}.
An important feature of this framework of choice functions is that it allows one to impose axioms directly on the decisions that are represented by such a choice function \cite{de2019interpreting,seidenfeld2010coherent,van2018natural}.
We here adopt the coherence axioms that were put forward by De Bock and De Cooman \cite{de2019interpreting}.
We do not use these axioms directly though, but instead consider an alternative definition of coherence that is based on sets of preference orders and show that it is equivalent.

As we will explain and demonstrate in this contribution, we can use these coherent choice functions to infer new choices from previous choices.
In particular, for any given assessment of previous choices that is compatible with coherence, we will achieve this by introducing the so-called natural extension of this assessment: the unique most conservative coherent choice function that is compatible with the assessment.

We start in \cref{sec:choicef} with an introduction to choice functions and coherence.
\cref{sec:assessments} then first defines choice assessments, their consistency and their natural extension, and then goes on to reformulate these concepts in terms of coherent sets of desirable options.
Next, in \cref{sec:evalNatExt}, we show how this reduces the problems of checking the consistency of an assessment and computing its natural extension to something that we can solve practically and algorithmically.
The running time of the algorithm depends rather heavily on the size of the assessment that is provided though.
To reduce this running time, \cref{sec:simp,sec:simpGen} present several methods that can be used to replace an assessment by an equivalent object that contains the same information, but can be more efficiently used to check consistency and calculate the natural extension.
In \cref{sec:experiment}, we test our algorithms and examine how the size and imprecision of the assessment impact the time required to determine the natural extension.
\cref{sec:conc} concludes the paper and provides some suggestions for future work.

\section{Choice functions}\label{sec:choicef}
A choice function $C$ is a function that, when applied to a set of options, may reject one or more---but not all---options from that set.
The options that are not rejected are then said to be `chosen'.
Usually the options are actions that have a corresponding reward.
This reward furthermore depends on the state of an unknown---uncertain---variable $X$ that takes values in a set $\X$.
We will assume that the rewards can be represented by real numbers, on a real-valued utility scale.
In this setting, an option $u$ is thus a function from states $x$ in $\X$ to $\R$.
We will denote the set of all possible options by $\V\subseteq\R^{\X}$ and require that this forms a real vector space with pointwise vector addition and scalar multiplication. 
Bounded options are sometimes also called gambles in the literature.
Moreover, we endow $\V$ with the partial order $\precceq$: for all $u,v\in\V$,  $u\precceq v$ if and only if $ u(x)\le v(x)$ for all $x\in \X$; $\precc$ is the corresponding strict version, so $u\precc v$ if $u\precceq v$ and $u\neq v$.\footnote{In principle, our results in \cref{sec:choicef,sec:assessments,sec:simp,sec:simpGen} (except \cref{sec2:examples}) can also be adapted to work for any ordered vector space over the real numbers, i.e. a vector space \(\mathcal{W}\) and a partial order \(\triangleleft\) on \(\mathcal{W}\) for which for any vectors \(u,v,w\in \mathcal{W}\) and real number \(r > 0\), \(u\triangleleft v\) implies \(u+w\triangleleft v+w\) and \(r u\triangleleft r v\); we restrict ourselves to $(\V,
\precc)$ for didactic purposes.}
We also let \(2^{\V}\) denote the power set of \(\V\).
To make all of this more tangible, we consider the following toy problem as a running example.
\begin{myexp}\label{ex:1.1}
A farming company cultivates tomatoes and they have obtained a large order from a foreign client.
However, due to government regulations they are not sure whether they can deliver this order.
So the state space $\X$ is $\{$order can be delivered, order cannot be delivered$\}$.
The company now has multiple options to distribute their workforce.
They can fully prepare the order, partially prepare the order or not prepare the order at all.
Since $\X$ only has two elements, we can identify the options with vectors in $\R^2$.
We will let the first component of these vectors correspond to the reward if the order can be delivered.
For example, the option of fully preparing the order could correspond to the vector $v_1\coloneqq(5,-3)$: if the order goes through, then the company receives a payment---or utility---of $5$ for that order;
however, if the order does not go through, the company ``receives'' a negative reward $-3$, reflecting the large amount of resources that they spent on an order that could not be delivered in the end.
\hfill\pushQED{$\lozenge$}\popQED\end{myexp}

Finiteness of sets will be important throughout this paper, and we therefore introduce the symbol \(\Subset\) to mean that a set is a non-empty finite subset.
A first example is that we will restrict ourselves to choices from finite sets of options. 
That is, the domain of our choice functions will be $\Q\coloneqq\{A\colon A\Subset\V\}$, the set of all finite subsets of $\V$ excluding the empty set.
We also let \(\Qe\coloneqq \Q\cup \{\emptyset\}\).
Formally, a \emph{choice function} is then any function~$C\colon \Q\to\Q$ such that $C(A)\subseteq\nobreak A$ for all $A\in\Q$.
We will also consider the corresponding rejection function $R_C\colon \Q\to\Qe\colon A\mapsto A\setminus C(A)$. 

\par\label{par:interpretation} We will give the following interpretation to these choice functions.
For every set $A\in \Q$ and option $u\in A$, we take $u\in C(A)$---$u$ is `chosen'---to mean that there is no other option in $A$ that is preferred to $u$.
Equivalently, $u\in R_C(A)$---$u$ is rejected from $A$---if there is an option in $A$ that is preferred to $u$.

\begin{myexp}\label{ex:1.2}
We will let the choice function $C$ correspond to choices that the strategic advisor of the company makes or would make for a given set of options, where these choices can be multivalued whenever he does not single out a unique best option. 
Suppose for example that he has rejected $v_3$ and $v_4$ from a set $A_1\coloneqq\{v_1,v_2,v_3,v_4\}$, with
$
v_1\coloneqq(5,-3),$ $v_2\coloneqq(3,-2),$ $v_3\coloneqq(1,-1),$ and $v_4\coloneqq(-2,1),
$
but remains undecided about whether to choose $v_1$ or $v_2$.
This corresponds to the statement $C(A_1)=\{v_1,v_2\}$, or equivalently, $R_C(A_1)=\{v_3,v_4\}$.
\hfill\pushQED{$\lozenge$}\popQED\end{myexp}

\citet{de2019interpreting} define coherent choice functions by imposing properties for rationality on the corresponding rejection functions.
An example of such a property is their axiom \(\mathrm{R}_4\), which is analogous to Sen's Alpha \cite{sen1971choice,sen1977social}: 
 if some option is rejected from a set, then it will also be rejected from any of its supersets.
We here opt to not define coherent choice functions in terms of such properties though, but to instead use an equivalent characterisation in terms of sets of preference orders.

In particular, these preference orders are taken to be strict partial vector orders on $\V$, typically denoted by  $\prec$, that extend the original strict order $<$.
This means that they should have the following properties {\cite[$\triangleright_0-\triangleright_4$]{de2019interpreting}}: for all $u,v,w\in\V$ and $\lambda>0$, 
\begin{axiom}[label=$\mathrm{\prec}_{\arabic*}$., ref=$\mathrm{\prec}_{\arabic*}$,start=0]
\item\label{ax:Oantisym} $u\not\prec u$, \hfill (irreflexivity)
\item\label{ax:Otransi} if $u\prec v$ and $v\prec w$ then also $u\prec w$, \hfill (transitivity)
\item\label{ax:Otransla} if $u\prec v$ then also $u+w\prec v+w$, \hfill (translation invariance)
\item\label{ax:Oscal} if $u\prec v$ then also $\lambda u\prec \lambda v$, \hfill (scaling invariance)
\item\label{ax:Oext}  if $u< v$ then $u\prec v$. \hfill (extends \(<\))
\end{axiom}
We call orderings that satisfy \cref{ax:Otransla,ax:Otransi,ax:Oext,ax:Oscal,ax:Oantisym} \emph{(coherent) preference orders} and let \(\mathbb{O}\) denote the set of all preference orders on \(\V\).
Note that thanks to \cref{ax:Otransla} preference orders are fully determined by their set of desirable options~\(G_{\prec}\coloneqq \{u\in \V \colon 0\prec u\}\), where desirable means preferred to the status quo \(0\).
These sets of desirable options will be important later on and are useful because they reduce the `pairs' of options necessary to describe a partial vector order to just `single' options.
We will call a set of desirable options \(G\subseteq\V\) \emph{coherent} if it satisfies the following properties:
for all \(u,v\in G\), \(w\in \V\) and \(\lambda>0\),
\begin{axiom}[label=$\mathrm{d}_{\arabic*}$., ref=$\mathrm{d}_{\arabic*}$,start=0]
  \item\label{ax:Dantisym} $0\notin G$, \hfill (irreflexivity)
  \item\label{ax:Dtransla} $u+v\in G$, \hfill (additivity)
  \item\label{ax:Dscal}  $\lambda u\in G$, \hfill (scaling invariance)
  \item\label{ax:Dext}  if $0<w$ then $w\in G$. \hfill (extends \(<\))
  \end{axiom}
  We furthermore use \(\overline{\mathrm{\mathbf{G}}}\) to denote the set of all coherent sets of desirable options.
  As our next result shows, if \(\prec\) is a preference order, then \(G_{\prec}\) is  coherent.
  Furthermore, for any set of options \(G\subseteq \V\), if we define the binary relation~\(\prec_G\), for all \(u,v\in \V\) by 
  \[
  u\prec_G v \Leftrightarrow v-u\in G,
  \]
  then the coherence of \(G\) implies that \(\prec_G\) is a preference order.
  Working with coherent sets of desirable options, or with preference orders, is therefore equivalent.
\begin{lemma}\label{lem:axD}
For any preference order \(\prec\), \(G_{\prec}\) is a coherent set of desirable options.
Moreover, the map \(\prec\mapsto G_{\prec}\) is a bijection between the set of preference orders and the set of coherent sets of desirable options, with inverse \(G\mapsto \prec_G\).
\end{lemma}
\begin{proof}
  First we prove that \(G_{\prec}\) is coherent for any preference order \(\prec\).
  Take any preference order~\(\prec\).
  Then by \cref{ax:Oantisym} we have \(0\not\prec 0\) which implies \cref{ax:Dantisym}.
  If \(0\prec u\) and \(0\prec v\) by \cref{ax:Otransla} we have \(0\prec u\prec u+v\), so by \cref{ax:Otransi} we have \(0\prec u+v\), which implies \cref{ax:Dtransla}.
  Similarly, \cref{ax:Dscal} follows from \cref{ax:Oscal} and \cref{ax:Dext} from \cref{ax:Oext}.

  Next we prove that the map \(\prec\mapsto G_{\prec}\) is a bijection.
  First we prove that this map is injective. 
  Assume that we have two preference orders \(\prec\) and \(\prec'\) such that \(G_{\prec}=G_{\prec'}\).
  Then for all \(u,v\in \V\), we have \(v-u\in G_{\prec}\) if and only if \(v-u\in G_{\prec'}\).
  So, \(0\prec v-u\) if and only if \(0\prec' v-u\) and therefore, by \cref{ax:Otransla},  \(u\prec v\) if and only if \(u\prec' v\) for all \(u,v\in \V\), or equivalently, \(\prec=\prec'\).

  To prove that the map \(\prec\mapsto G_{\prec}\) is surjective and has inverse \(G\mapsto \prec_G\), assume that we have some coherent \(G\subseteq \V\) and consider \(\prec_G\).
    Then \cref{ax:Otransla} is satisfied trivially for \(\prec_G\) and \cref{ax:Oantisym,ax:Oscal,ax:Oext} follow immediately from respectively \cref{ax:Dantisym,ax:Dscal,ax:Dext}.
  For \cref{ax:Otransi}, assume that \(u\prec_G v\) and \(v\prec_G w\).
  Then \(v-u\in G\) and \(w-v\in G\), so \(w-u=(w-v)+(v-u)\in G\) by \cref{ax:Dtransla}, hence \(u\prec_G w\).
  So we conclude that \(\prec_G\) is a preference order.
  Furthermore, for all \(u\in \V\), we have by definition that \(u\in G\) if and only if \(0\prec_G u\), so by definition \(G=G_{\prec_G}\), which shows that \(G\mapsto \prec_G\) is indeed the inverse of \(\prec \mapsto G_{\prec}\).
\end{proof}

For any preference order \({\prec}\in \mathbb{O}\) on \(\V\), inspired by our interpretation for choice functions, we consider the corresponding choice function \(C_{\prec}\) defined by
\[
  C_{\prec}(A)\coloneqq \cset{u\in A}{ (\forall a\in A) u\not\prec a}\text{ for all \(A\in \Q\).}
\]
The corresponding rejection function \(R_{\prec}\) \cite[Equation~(1)]{de2019interpreting} is then given by
\[
  R_{\prec}(A)\coloneqq \cset{u\in A}{ (\exists a\in A) u\prec a}\text{ for all \(A\in \Q\).}
\]
Crucially, however, such a preference order $\prec$ need not be known.
Instead, in its full generality, our definition will allow for the use of a set of preference orders \(\OO\subseteq \OOO\). 
One interpretation is that there is some true preference order that we do not know exactly, but of which we only know that it belongs to some set \(\OO\) of candidate orders.
Another interpretation is that we aim to study the consequences of a set of preference orders \(\OO\), for example  when a group of people with different preferences try to make a common decision.
Any such set of preference orders \(\OO\subseteq \OOO\) corresponds to a function \(C_{\OO}\colon \Q\to \Qe\) defined for all \(A\in \Q\) by
\begin{equation}\label{eq:oorsprdef}
  C_{\OO}(A)\coloneqq \cset{u\in A}{ (\exists {\prec}\in \OO) (\forall a\in A) u\not\prec a}=\bigcup_{{\prec} \in \OO} C_{\prec}(A);
\end{equation}
it represents the choices---or rather, the rejections---that are compatible with each of the orders \(\prec \in \OO\).
Whenever \(C_{\OO}\) is a choice function, the corresponding rejection function \(R_{\OO}\) is given by
\begin{equation}\label{eq:Rdef}R_{\OO}(A)\coloneqq A\setminus C_{\OO}(A) = \cset{u\in A}{ (\forall {\prec}\in \OO) (\exists a\in A) u\prec a}\text{ for all \(A\in \Q\).}\end{equation}
By definition, we already have that \(C_{\OO}(A)\subseteq A\) for all \(A\in \Q\).
For \(C_{\OO}\) to be a choice function---a map from \(\Q\) to \(\Q\)---we also need that \(C_{\OO}(A)\neq\emptyset\) for all \(A\in \Q\). 
This will be the case whenever \(\OO\) is not empty, as implied by the following proposition.
\begin{proposition}\label{prop:niet-leeg}
Fix some (possibly empty) set of preference orders~\(\OO\subseteq \OOO\).
Then for any \(A\in \Q\), \(C_{\OO}(A) = \emptyset\) if and only if \(\OO = \emptyset\).
\end{proposition}
We will prove this using the following well-known lemma, which will come in handy in our proof of \cref{lem:bestekomenuitV} further on as well.
We state it without proof since the result is essentially well-known from lattice theory; see for example {\cite[I.3~Theorem~3]{birkhoff1940lattice}}.\footnote{Note that in this book partial orders are originally defined to be non-strict, but thanks to Lemma 1 in Section I.1 of \cite{birkhoff1940lattice}, strict partial orders \(\prec\) are connected to non-strict partial orders \(\preceq\) by \(u\prec v \Leftrightarrow u\preceq v \text{ and } u\neq v\).
Moreover, the definition of maximal elements and the proof of this theorem uses strict partial orders.}
\begin{lemma}\label{lem:undominated}
Consider any strict preorder \(\prec\), so an irreflexive (\(\prec_0\)) and transitive \((\prec_1)\) binary relation, on a finite set \(A\). Then for any element \(a\in A\) there is an element \(a^*\in A\), with \(a\prec a^* \) or \(a=a^*\), such that \(a^*\) is undominated in \(A\), meaning that there is no \(b\in A\) for which \(a^*\prec b\).
\end{lemma}
\begin{proofof}{\cref{prop:niet-leeg}}
The reverse implication follows from \cref{eq:oorsprdef}.
We prove the direct implication by contraposition.
Let \(\OO\neq \emptyset\) and fix any \({\prec} \in \OO \).
Take any \(A\in \Q\) and \(u\in A\).
Then by \cref{lem:undominated} there is an undominated \(u^*\in A\).
But then \(u^*\in C_{\OO}(A)\neq \emptyset\) by definition.
\end{proofof}
This allows us to define coherence for choice functions as follows.

\begin{definition}\label{def:Ccoh}
We call a choice function $C\colon\Q\to\Q$ \emph{coherent} if there is some non-empty set of preference orders~\(\OO\subseteq \OOO\) such that \(C=C_{\OO}\). 
\end{definition}
Alternatively, as explained earlier, and as we formally prove in \ref{appsec:equivAx}, this definition of coherence can also be equivalently characterised using five axioms \citep[$\mathrm{R}_0-\mathrm{R}_4$]{de2019interpreting} for rejection functions, but our characterisation in terms of preference orders is more convenient in the present setting.

We note also that the set of preference orders \(\OO\) in \cref{def:Ccoh} need not be unique.
It follows from \cref{eq:oorsprdef} that the largest set of preference orders \(\OO\) for which a coherent choice function \(C\) is equal to \(C_{\OO}\) is
\[
  \OO_C=\{{\prec}\in \OOO\colon (\forall A\in \Q) C_{\prec}(A)\subseteq C(A)\}=\bigcap_{A\in \Q} \{{\prec}\in \OOO\colon C_{\prec}(A)\subseteq C(A)\}. 
\]


\subsection{Examples}\label{sec2:examples}
Let us now try to get a bit more feel for the previous theory by looking at some well-known (sets of)  orderings.
We will show, for example, that due to the connection between preference orders and so-called coherent lower expectations, decision-making with maximality or E-admissibility \cite{troffaes2007decision}, or by maximising expected utility, all fit in our framework.
To do this, we will assume in this \cref{sec2:examples} that \(\V\) is the subset of \(\R^\X\) that contains all bounded options (that is, all gambles).

\subsubsection{Lower expectations}\label{sec3:lowerexp}
A coherent lower expectation \(\Eu\) is a functional on \(\V\) that satisfies the following properties: for all \(f,g\in \V\) and \(\lambda\geq 0\) \citep[Section~2.2.1]{augustin2014introduction}\footnote{They are also called coherent lower previsions, and are then typically denoted by \(\underline{\mathrm{P}}\).},
\begin{enumerate}
  \item \(\Eu(f)\geq \inf(f)\), \hfill (boundedness)
  \item \(\Eu(f+g)\geq \Eu(f)+\Eu(g)\), \hfill (super-linearity) 
  \item \(\Eu(\lambda f)=\lambda \Eu(f)\). \hfill (non-negative homogeneity)
\end{enumerate}
They are relevant to our framework because, as discussed by Quaeghebeur \citep[Section~1.6.3]{augustin2014introduction}, every preference order \(\prec\)---or, equivalently, every corresponding coherent set of desirable options \(G_{\prec}\coloneqq \{u\in \V \colon 0\prec u\}\)---determines a coherent lower expectation
\[
  \Eu_{\prec} \colon \V \to \R \colon u \mapsto \Eu_{\prec}(u) \coloneqq \sup \{\alpha\in \R \colon  \alpha \prec u \}=\sup\{\alpha\in \R \colon u-\alpha \in G_{\prec}\},
\]
where we identify any real number \(\alpha\) with the option that takes on the constant value \(\alpha\).
The lower expectation \(\Eu_{\prec}\) `forgets' only the `border structure' of the preference order, in the sense that the original preference order can be retrieved up to its boundary behaviour.
A pseudoinverse that we consider here is
\[
  \prec_{\Eu}\coloneqq \{(u,v) \colon u<v \text{ or }0 <\Eu(v-u)\},
\]
as this is the most conservative preference order that is compatible with a given lower expectation~\(\Eu\), i.e. the smallest preference order \(\prec\) for which \(\Eu=\Eu_{\prec}\).
All possible pseudoinverses are preference orders \(\prec\) for which \(\prec_{\Eu}\subseteq \prec \subseteq \{(u,v) \colon 0 \leq \Eu(v-u) \text{ and } u\neq v\}\);
for each such pseudoinverse~\(\prec\), including \(\prec_{\Eu}\), we have that \(\Eu_{\prec}=\Eu\).

In the special case that \(\Eu(u)=-\Eu(-u)\)  for all \(u\in \V\), \(\Eu\) is a linear functional  \citep[Section~2.2.2]{augustin2014introduction} and is therefore called a linear expectation.
Such a linear expectation is typically denoted simply as \(\E\).
When \(\X\) is finite, these linear expectations have a unique corresponding so-called probability mass function (pmf) \(p\colon \X \to [0,1]\) with \(\sum_{x\in \X} p(x)=1\) such that \(\E(u)=\E_p(u) \coloneqq \sum_{x\in \X} p(x) u(x)\)  for every option~\(u\in \V\).
This makes such linear expectations particularly easy to specify and evaluate.

An important observation, that will prove convenient in \cref{sec:experiment}, is that a general coherent lower expectation \(\Eu\) can also be interpreted as a lower envelope of a set of such linear expectations \cite[Propositions~2.3]{augustin2014introduction}.
In particular, if \(\X\) is finite, then for every coherent lower expectation \(\Eu\) there is a closed convex set \({P}\) of pmfs such that \(\Eu(u)=\min_{p\in {P}} \E_p(u)\) for every option~\(u\in \V\).
In practice this set \({P}\) will furthermore often have a finite number of extreme points \(p_1,...,p_m\), in which case we can evaluate \(\Eu\) efficiently because then \(\Eu(u)=\min_{j=1}^m  \E_{p_j}(u)\) for every option~\(u\in \V\).

\subsubsection{Decision-making based on lower expectations}
Since preference orders are related to coherent lower expectations, it should not come as a surprise that the same is true for choice functions, or decision-making.

The classical way of decision-making is by maximising expected utility, as for example described in \cite[Section~2]{troffaes2007decision}, in which case we start from a single linear expectation \(\E\) that determines everything.
For any option set \(A\in \Q\), we then map every option \(u\in A\) to its expected utility \(\E(u)\) and choose the option that maximises this expected utility, or in case of a tie, the set of options that maximise it and are undominated with respect to \(<\).
Since \(\E\) is a linear functional, we have that `\(\E(u)<\E(v)\) or \(u<v\)' if and only if \(u\prec_{\E} v\).
So the chosen options are the options that are undominated by \(\prec_\E\), and therefore the options chosen by \(C_{\prec_\E}\).

A second important way of decision-making, which starts from a single lower expectation \(\Eu\), is by maximality, as for example described by \citet[Section~3.2]{troffaes2007decision}.
When choosing with maximality, one compares options pairwise: an option \(v\) is preferred over an option \(u\) if \(v\) dominates \(u\) (so \(u<v\)) or if one would pay a positive amount of utility to swap \(u\) for \(v\), in the sense that \(\Eu(v-u)>0\).
This is exactly what \(\prec_{\Eu}\) expresses.
Maximality then chooses those options that are undominated with respect to this ordering \(\prec_{\Eu}\).
So we see that
 \(
  C_{\prec_{\Eu}}
  \)
   corresponds to choosing with maximality for the lower expectation \(\Eu\).

A third important way of decision-making is E-admissibility, as for example described in \cite[Section~3.4]{troffaes2007decision}.
In that case, one starts with a set \(\mathcal{E}\) of linear expectations on \(\V\).
The chosen options are the ones  for which there is at least one \(\E\in \mathcal{E}\) for which \(C_{\prec_{\E}} \) chooses them.
So if we consider the set of preference orders \(\OOO(\mathcal{E})\coloneqq \{\prec_{\E} \colon \E \in \mathcal{E}\}\),
then the corresponding choice function \(C_{\OOO(\mathcal{E})}\) chooses with E-admissibility because, for any \(A\in \Q\), we have that
\(
  C_{\OOO(\mathcal{E})}(A)= \bigcup_{ \prec\in \OOO(\mathcal{E})} C_{\prec}(A) =\bigcup_{ \E\in \mathcal{E}} C_{\prec_{\E}}(A) 
\).

Our framework however also captures a generalisation of the previous methods (although this generalisation is still not as general as our full framework). 
Consider a set \(\mathcal{E}\) of lower expectations.
Then we have a set \(\OOO(\mathcal{E})\coloneqq\{\prec_{\Eu} \colon \Eu\in \mathcal{E}\}\) of preference orders that defines a choice function \(C_{\mathcal{E}}\coloneqq C_{\OOO(\mathcal{E})}\), with
\begin{align*} 
  C_{\mathcal{E}}(A)&= C_{\OOO(\mathcal{E})}(A)\\
  &=\left\{u \in A \colon (\exists \prec_{\Eu}\in \OOO(\mathcal{E}))(\forall v \in A) u\not\prec_{\Eu} v  \right\}\\
  &=\left\{u \in A \colon (\exists {\Eu}\in \mathcal{E})(\forall v \in A)  {\Eu}(v-u) \leq 0 \text{ and }u\not< v   \right\}
\end{align*}
for every \(A\in \Q\), where the last equality follows from the definition of \(\prec_{\Eu}\).
If all expectations in \(\mathcal{E}\) are linear then we choose with E-admissibility and if \(\mathcal{E}\) is a singleton, then we choose with maximality.
And if  \(\mathcal{E}\) contains a single linear expectation, we end up maximising expected utility.

\section{Consistency and the natural extension of a choice assessment}
\label{sec:assessments}
Now that we are familiar with choice functions, we move on to the topic of this paper: how to extend a partial choice assessment to a coherent choice function.
In this endeavour, we consider a decision-maker and assume that there is some coherent choice function~\(C\) that represents her preferences.
However, we (or she) may not fully know this function.
Our partial information about~\(C\) comes in the form of preferences regarding some---so not necessarily all---option sets.
In particular, we assume that for some option sets~\(A\in\Q\), we know that the decision-maker rejects all options in~\(W\subseteq A\), meaning that \(C(A)\subseteq A\setminus W\).
An equivalent way of expressing this, which will be more convenient for our purposes, is to state that \(C(V\cup W)\subseteq V\), with \(V\coloneqq A\setminus W\), or equivalently, that \(W \subseteq R_{C}(V\cup W)\).
We will represent such information by an \emph{assessment}: a set~\(\Ass\subseteq \Q\times\Qe\) of pairs~\((V,W)\) of disjoint option sets---so \(V\cap W=\emptyset\)---with the interpretation that, for all~\((V,W)\in\Ass\), the options in~\(W\) are definitely rejected from~\(V\cup W\).
Note that we do not allow \(V=\emptyset\) because this cannot represent partial information about a coherent choice function due to \cref{def:Ccoh,prop:niet-leeg}.
Also note that \(W=\emptyset\) is uninformative, since it simply states that \(C(V)\subseteq V\), but is nevertheless allowed.

To make this idea more concrete, let us go back to the example.

\begin{myexp}\label{ex:1.3}
Suppose that the strategic advisor of the farming company has previously rejected the options $v_3$ and $v_4$ from the option set $A_1$, as 
in \cref{ex:1.2}, 
and has chosen $v_6$ from $A_2\coloneqq\{v_5,v_6\}$, where
$
v_5\coloneqq(3,1)$ and $v_6\coloneqq(-4,8).
$
This corresponds to the assessment
\[
\Ass=\{(\{v_1,v_2\},\{v_3,v_4\}),(\{v_6\},\{v_5\})\}.
\]
Suppose now that the company's strategic advisor has fallen ill and the company is faced with two new decision problems that amount to choosing from the set $A_3=\{(-3,4),(0,1),(4,-3)\}$ and from \(A_4=\{(-2,2),(5,-4)\}\).
Since no such choices were made before, the conservative option is to make the completely uninformative statements $C(A_3)\subseteq A_3$ and $C(A_4)\subseteq A_4$.
However, perhaps the company can make more informative choices by taking into account the advisor's previous choices?
\hfill\pushQED{$\lozenge$}\popQED\end{myexp}
If previous choices are not readily available, we can of course also elicit such assessments from experts. 
In that case, a practical benefit of eliciting such types of assessment is that the expert is not required to have knowledge about the underlying uncertainty model, since a choice assessment is agnostic about that.
\subsection{Introducing consistency and natural extension}

Given an assessment~$\Ass$ that we would like to extend, a first important question is whether there is a coherent choice function~\(C\) that agrees with it. 
\begin{definition}\label{def:consistentAss}
An assessment~\(\Ass\) is \emph{consistent} if there is a coherent choice function~\(C\) such that \(C(V\cup W)\subseteq V\) for all~\((V,W)\in\Ass\).
\end{definition}
To characterise this notion of consistency, we recall that a choice function~\(C\) is coherent whenever there is some non-empty set of preference orders~\(\OO\) such that \(C=C_{\OO}\), and observe that it follows from~\cref{eq:oorsprdef} that \(C_{\OO}\) satisfies the conditions in \cref{def:consistentAss} if and only if
\[
\OO
\subseteq\OOO(\Ass)
\coloneqq 
\cset[\big]{{\prec}\in \mathbb{O}}{(\forall(V,W)\in\Ass)\;C_{\prec}(V\cup W)\subseteq V}.
\]
Hence, \(\Ass\) is consistent if and only if \(\OOO(\Ass)\neq\emptyset\).
To illustrate this concept, we go back to our running example.
\begin{myexp}\label{ex:1.4a}
  To show that the assessment from \cref{ex:1.3} is consistent, we can for example consider the probability mass function \((\frac49,\frac59)\) and the corresponding linear expectation defined by \(\EE((a,b))=\frac49 a + \frac59 b\) for any \((a,b)\in \R^2\).
  Then, as we know from \cref{sec3:lowerexp}, \(\prec \coloneqq \prec_{\EE}\) is a coherent preference order.
    To show that the preference order~\(\prec\) agrees with the assessment, we verify that for all~\((V,W)\in\Ass\), we have that \(C_{\prec}(V\cup W)\subseteq V\).
    For \(A_1=\{v_1,v_2,v_3,v_4\}=\{(5,-3),(3,-2),(1,-1),(-2,1)\}\), we see that \(v_4\prec v_3 \prec v_2 \prec v_1\) and therefore, indeed, \(C_{\prec}(\{v_1,v_2\}\cup\{v_3,v_4\})=\{v_1\}\subseteq \{v_1,v_2\}\).
    For \(A_2=\{v_5,v_6\}=\{(3,1),(-4,8)\}\) we see that \(v_5\prec v_6\) and therefore, indeed, \(C_{\prec}(\{v_6\}\cup\{v_5\})=\{v_6\}\subseteq \{v_6\}\).
    Hence, \(\prec\in \OOO(\Ass)\neq\emptyset\), so the assessment is consistent.
  \hfill\pushQED{$\lozenge$}\popQED\end{myexp}

If an assessment~\(\Ass\) is consistent and there is more than one coherent choice function that agrees with it, then the question remains which one we should use.
A careful decision-maker would only want to reject options if this is implied by the assessment.
So she wants a most conservative agreeing coherent choice function: one that rejects the fewest number of options.
Since larger sets of preference orders lead to more conservative choice functions, this most conservative agreeing choice function then clearly exists, and is then equal to~\(C_{\OOO(\Ass)}\). 
For notational convenience, for any assessment \(\Ass\), we denote this function by~\(\NatExt\coloneqq C_{\OOO(\Ass)}\), regardless of whether \(\Ass\) is consistent; the term natural extension though, we reserve for the consistent case. 

\begin{definition}\label{def:NatExt}
Whenever an assessment \(\Ass\) is consistent, we call \(\NatExt\) the \emph{natural extension} of~\(\Ass\).
\end{definition}

Once more, we illustrate this concept using our running example.
\begin{myexp}\label{ex:1.4b}
  As mentioned in \cref{ex:1.3}, we want to choose from the sets \(A_3=\{(-3,4),(0,1),(4,-3)\}\) and \(A_4=\{(-2,2),(5,-4)\}\).
  Since we know from \cref{ex:1.4a} that our assessment is consistent, this amounts to evaluating the following natural extensions: \(C_{\Ass}(A_3)\) and \(C_{\Ass}(A_4)\).
  A first step could be to use the preference order \(\prec\) from \cref{ex:1.4a} and evaluate \(C_{\prec}(A_3)\) and \(C_{\prec}(A_4)\).
  We see that \( (4,-3) \prec (0,1)\) and \((0,1)\prec (-3,4)\), so \(C_{\prec}(A_3)=\{(-3,4)\}\); similarly \((5,-4)\prec (-2,2)\), so \(C_{\prec}(A_4)=\{(-2,2)\}\).
  However, this would not yield the most conservative---and hence natural---extension, as we will show now.
  Consider for example the preference order \(\prec'\coloneqq \prec_{\EE'}\) that corresponds to the linear expectation \(\EE'\) that is associated with the probability mass function \((\frac{9}{19},\frac{10}{19})\), defined by  \(\EE'((a,b))=\frac{9}{19} a + \frac{10}{19} b\) for any \((a,b)\in \R^2\).
  This is a preference order in \(\OOO(\Ass)\) for the same reasons as \(\prec\): \(v_4\prec' v_3 \prec' v_2 \prec' v_1\) and \(v_5 \prec' v_6\).
  We can furthermore check that \(C_{\prec'}(A_3)=\{(-3,4)\}\) and \(C_{\prec'}(A_4)=\{(5,-4)\}\).
  Hence, since both \(C_{\prec}(A_4)=\{(-2,2)\}\) and \(C_{\prec'}(A_4)=\{(5,-4)\}\) are contained in the union \(C_{\Ass}(A_4)=\bigcup_{\prec\in \OOO(\Ass)} C_\prec(A_4)\), we know that \(C_{\Ass}(A_4)=A_4\).
  For \(A_3\), however, we only know that \(\{(-3,4)\}\subseteq C_{\Ass}(A_3)\), but cannot exclude the possibility that there are preference orders in \(\OOO(\Ass)\) that choose other options than \((-3,4)\).
  Finding \(C_{\Ass}(A_3)\) therefore requires us to check more---and possibly all---preference orders in \(\OOO(\Ass)\).
  \hfill\pushQED{$\lozenge$}\popQED\end{myexp}

As these examples demonstrate, checking consistency and evaluating the natural extension of a consistent assessment is no easy task.
That said, we do see that the consistency and the natural extension of an assessment \(\Ass\) are both entirely characterised by \(\OOO(\Ass)\).
Given the importance of this set, we will now investigate its structure in more detail.

\subsection{Alternative characterisations of \texorpdfstring{\(\OOO(\Ass)\)}{the set of preference orders of the assessment}}\label{sec:charactOOOAss}
As a first step, we consider the following more practical expression for~\(\OOO(\Ass)\).


\begin{proposition}\label{prop:asss}
Consider an assessment~\(\Ass\).
Then \[\OOO(\Ass)=\cset[\big]{{\prec} \in  \mathbb{O}}{(\forall (V,W)\in\Ass )(\forall w\in W)(\exists v\in V) w\prec v}.\]
\end{proposition}
To prove this, it suffices to apply the following lemma to every \((V,W)\in \Ass\).
\begin{lemma}\label{lem:bestekomenuitV}
For any preference order \(\prec\) and any disjoint \(V\in\Q\) and \(W\in \Qe\) the following statements are equivalent:
\begin{equation}\label{eq:caral1}
 C_{\prec}(V\cup W)\subseteq V
\end{equation}
 and
 \begin{equation}\label{eq:caral2}
 (\forall w\in W)(\exists v\in V)w \prec v.
 \end{equation}
\end{lemma}
\begin{proof}
Both statements are trivially true if \(W=\emptyset\), so it suffices to consider the case \(W\neq \emptyset\).
First we prove that \cref{eq:caral1} implies \cref{eq:caral2}.
From \cref{eq:caral1} and the fact that \(V\) and \(W\) are disjoint, it follows that \(w\notin C_{\prec}(V\cup W)\) for all~\(w\in W\).
This means by definition that 
\begin{equation}\label{eq:oorsprk}
(\forall w\in W)(\exists a\in V\cup W)w \prec a.
\end{equation}
We will now show that this implies \cref{eq:caral2}.
Take any option~\(w\in W\).
By \cref{lem:undominated} there is some option~\(w^*\in W\) that is undominated in \(W\) with respect to~\(\prec\) such that \(w\prec w^*\) or \(w=w^*\). 
Since \(w^*\in W\), we know from \cref{eq:oorsprk} that there is some~\(a^*\in V\cup W\) such that \(w^*\prec a^*\).
Since \(w^*\) is undominated in~\(W\) with respect to~\(\prec\), it is impossible that \(a^*\in W\),
so it must be that \(a^*\in V\).
Thus, we have found some \(a^*\) in \(V\) such that \(w\prec w^*\prec a^*\) or \(w=w^*\prec a^*\) and therefore, in any case, \(w \prec a^*\).
As this holds for any option~\(w\in W\), this proves~\cref{eq:caral2}.

Next we prove that \cref{eq:caral2} implies \cref{eq:caral1}.
Take any option~\(w \in W\).
Since \(V\subseteq V\cup W\), we have from \cref{eq:caral2} and the definition of~\(C_{\prec}\) that \(w\notin C_{\prec}(V\cup W)\).
Since this holds for any~\(w\in W\), it follows that \(C_{\prec}(V\cup W)\subseteq V\), and this is \cref{eq:caral1}.
\end{proof}

\begin{proofof}{\cref{prop:asss}}
This follows immediately from the definition of~\(\OOO(\Ass)\) and \cref{lem:bestekomenuitV}.
\end{proofof}

Taking a closer look at the expression in \cref{prop:asss}, and since we know from \cref{ax:Otransla} that \(w\prec v\) is equivalent to \(0\prec v-w\), we see that 
\[
  \OOO(\Ass)= \{ \prec \in \OOO\colon (\forall H\in \D_{\Ass})(\exists h \in H) 0\prec h  \},
\]
with \(\D_{\Ass}\coloneqq \{\{v-w\colon v\in V\}\colon (V,W)\in \Ass,w\in W\}\); we call this \(\D_{\Ass}\) the \emph{conjunctive generator}.\footnote{This name is chosen because of similarities with the conjunctive normal form; see \cref{eq:OOODunieOverOd}. As for our choice of letter, since \(\G\) is reserved for the disjunctive generator that will be introduced next, we opted for the next letter in the alphabet. Moreover, in West Flemish the letter `h' is also called `g upwards'. }
So we see that \(\OOO(\Ass)\) is a specific instance of a set of preference orders of the form
\[
  \OOO(\D)\coloneqq \{\prec \in \OOO\colon (\forall H\in \D)(\exists h \in H) 0 \prec h\},  
\]
with \(\D\subseteq \Qe\) a set of option sets; in particular \(\OOO(\Ass)=\OOO({\D_\Ass})\).
To analyse this expression further, let us introduce for any option \(h\in \V\) a corresponding set of preference orders~\(\OOO_h\coloneqq \{\prec \in \OOO \colon 0\prec h\}\) and for any option set \(H\in \Qe\) the notation 
\begin{equation}\label{eq:OOOHdefVierkanteHakenSterSter}
  \OOO[H]\coloneqq \{\prec \in \OOO \colon (\exists h \in H) 0 \prec h\} = \bigcup_{h\in H} \OOO_h,
\end{equation}
this enables us to write 
\begin{equation}\label{eq:OOODunieOverOd}
  \OOO(\D)=\bigcap_{H\in \D} \bigcup_{h\in H} \OOO_h = \bigcap_{H\in \D} \OOO[H] .
\end{equation}
We will now proceed to transform \cref{eq:OOODunieOverOd} into a union of intersections.
In particular, since every preference order in the set \(\OOO({\D})\) prefers at least one option~\(h\) in each option set~\(H\in \D\) to zero, we will split the set \(\OOO(\D)\) in (possibly overlapping) subsets of preference orders, according to which options \(h\) in each \(H\) they prefer to zero.

To formalise this, for any \(\D\subseteq \Qe\), we let \(\Phi(\D)\) be the set of selection functions, so those maps \(\phi\colon \D \to \V\) such that \(\phi(H)\in H\) for every \(H\in \D\).
Then 
\begin{equation}\label{eq:generatordef}
  \G(\D)\coloneqq\{\{\phi(H)\colon H\in \D \}\colon  \phi \in \Phi({\D}) \}
\end{equation}
is the set of all sets that can be obtained by selecting one option from each \(H\in \D.\)\footnote{\label{foot:4} A noteworthy case is \(\G(\emptyset)=\{\emptyset\}\). This follows from the fact that if there are no sets to select from, selecting one option from ``each'' of these sets yields the empty set. More formally, it follows from the fact that there is a single unique function that maps ``every'' element of \(\emptyset\) to an element of \(\V\). Another noteworthy case is that \(\G(\{\emptyset\})=\emptyset\).}
The case where \(\D=\{H_1,...,H_m\}\) is finite might make this more intuitive, because then 
\begin{equation}\label{eq:FiniteGdef}
  \G(\D)=\{\{h_1,...,h_m\}\colon (h_1,...,h_m)\in \times_{k=1}^m H_k\}.
\end{equation}
Since \(\OOO({\D})\) consists of all preference orders that prefer at least one \(h\in H\) to zero for each \(H\in \D\), we now find that 
\begin{align*}
  \OOO (\D) &=\{\prec \in \OOO \colon (\exists G\in \G(\D))(\forall g\in G) 0 \prec g \}
  = \bigcup_{G\in \G(\D)} \bigcap_{g\in G} \OOO_g
  = \bigcup_{G\in \G(\D)} \OOOO[G], \addtocounter{equation}{1}\tag{\theequation} \label{eq:OOODisUnieOverGD}
\end{align*}
with, for every set of options \(G\in 2^{\V}\), 
\begin{equation}\label{eq:Pd1mdef}
  \OOOO[G]\coloneqq \bigcap_{g\in G} \OOO_g = \cset{{\prec} \in \mathbb{O}}{(\forall g\in G) \; 0\prec g }.
  \end{equation}
So we see that \(\OOO(\D)\) is a particular instance of a set of preference orders of the form
\begin{equation}\label{eq:OOOOGdefRondehakenSter}
  \OOOO(\G)\coloneqq \bigcup_{G\in \G} \OOOO[G],
\end{equation}
with \(\G\subseteq 2^\V\) a set of option sets.
In particular, \(\OOO(\D)=\OOOO(\G(\D))\).
We will refer to any such set \(\G\) as a \emph{disjunctive generator}, or simply a generator whenever it is clear from the context what type of generator we are referring to.
We will call an option set \(G\) inside a generator \(\G\) a \emph{generator set}.
Of particular importance is the generator \(\G_{\Ass}\coloneqq \G(\D_{\Ass})\) because it characterises \(\OOO(\Ass)\).




\begin{corollary}\label{th:Nat}
Consider an assessment~\(\Ass\subseteq \Q\times \Qe\).
Then \(\OOO(\Ass)= \OOOO(\A) \).
\end{corollary}
\begin{proof}
  Follows immediately from the fact that \(\OOO(\Ass)=\OOO({\D_\Ass})\) and \cref{eq:OOODisUnieOverGD}.
\end{proof}

To illustrate these new concepts and their relation, let us look at what happens in our running example.
\begin{myexp}\label{ex:1.4}
The conjunctive generator is 
\[
  \D_{\Ass}=\{\{v_1-v_3,v_2-v_3\},\{v_1-v_4,v_2-v_4\},\{v_6-v_5\}\}.
\]
To improve readability, let us define \(h_1\coloneqq v_1-v_3=(4,-2)\), \(h_2\coloneqq v_2-v_3=(2,-1)\), \(h_3\coloneqq v_1-v_4=(7,-4)\), \(h_4\coloneqq v_2-v_4=(5,-3)\) and \(h_5\coloneqq v_6-v_5=(-7,7)\).
Then \(\D_{\Ass}=\{\{h_1,h_2\},\{h_3,h_4\},\{h_5\}\}\) and the corresponding (disjunctive) generator is
\[
  \A=\G(\D_{\Ass})=\{\{h_1,h_3,h_5\},\{h_1,h_4,h_5\},\{h_2,h_3,h_5\},\{h_2,h_4,h_5\}\}.\tag*{$\lozenge$}
\]
\end{myexp}

\subsection{Consistency and natural extension for generators}\label{subsec2:consNatGen}

For a given assessment \(\Ass\), we have by \cref{th:Nat} that
\(
  \OOO({\Ass})  = \OOOO(\G_\Ass)
\).
However, as we will see in \cref{sec:simp,sec:simpGen}, the same is often true for other, simpler generators \(\G\).
For that reason, rather than focus on \(\G_{\Ass}\) in particular, we will consider arbitrary disjunctive generators \(\G\subseteq 2^{\V}\), the sets \(\OOOO(\G)\) of preference orders that are generated by them and the corresponding operators \(C^{\G}\coloneqq C_{\OOOO(\G)}\).
We start by defining consistency and natural extension for such generators.
\begin{definition}\label{def:generatedChoiceFunction}
A disjunctive generator~\(\G\subseteq 2^{\V}\) is called consistent if \(\OOOO(\G)\neq \emptyset\).
Whenever it is consistent, we call the corresponding choice function \(C^{\G}\) its natural extension.
\end{definition}
Note that, for the particular case of \(\G_{\Ass}\), the consistency and natural extension of \(\G_{\Ass}\) is equivalent to that of \(\Ass\): \(\OOO(\Ass)\neq \emptyset\Leftrightarrow  \OOOO(\G_{\Ass})\neq \emptyset\) and \(C^{\G_\Ass}=C_{\OOOO(\G_\Ass)} = C_{\OOO(\Ass)}=C_{\Ass}\).

Since \(\OOOO(\G)=\bigcup_{G\in \G}\OOOO[G]\), we see that a generator \(\G\) is consistent if and only if \(\OOOO[G]\neq \emptyset\)  for at least one \(G\in \G\).
Alternatively, due to \cref{prop:niet-leeg}, \(\G\) is consistent if and only if \(0\in C^{\G}(\{0\})\).
For the operator \(C^{\G}\) itself, on the other hand, we see from the definitions that
\begin{equation}\label{eq:CFgeneratordef}
  C^{\G}(A) = C_{\OOOO(\G)}(A)=\{u \in A\colon (\exists G \in \G)(\exists \prec \in \OOOO[G])(\forall a \in A) u \not\prec a\},
\end{equation}
for any \(A\in \Q\).
For these reasons, it will be useful to find a convenient way of checking, for any \(G\subseteq \V\), whether \(\OOOO[G]\neq \emptyset\) and whether there is some \(\prec\in \OOOO[G]\) such that \(u\not\prec a\) for all \(a\in A\).
As we will now show, both problems are closely related to well-known concepts from the theory of coherent sets of desirable options \cite{van2018natural}. 

\subsection{Consistency and natural extension for generator sets}\label{subsec:choose}

Since we know from \cref{lem:axD} that there is a one-to-one correspondence between preference orders and coherent sets of desirable options, studying preference orders \(\prec\in \OOOO[G]\) is equivalent to studying their corresponding coherent sets of desirable options \(G_\prec \in \overline{\mathrm{\mathbf{G}}}\).
More explicitly, it follows from the definition of \(\OOOO[G]\) and \(G_{\prec} \) that, for any preference order \(\prec\), 
\begin{equation}
  \label{lem:percsetEquivSODO}
  \prec \in \OOOO[G] \Leftrightarrow G\subseteq G_{\prec}.
\end{equation}


  So we see that working with \(\OOOO[G]\) is equivalent to working with
  \(
    \overline{\mathrm{\mathbf{G}}}_{G}\coloneqq \cset{D\in \overline{\mathrm{\mathbf{G}}} }{G\subseteq D}.
  \)
  Conveniently, this set of compatible coherent sets of desirable options is well studied, allowing us to borrow some results from the theory of coherent sets of desirable options.

  First, an option set \(G\subseteq \V\) is called \emph{consistent} if \(\overline{\mathrm{\mathbf{G}}}_{G}\neq \emptyset\), meaning that it can be extended to a coherent set of desirable options.
  \citet[Theorem~2,(i)\(\Leftrightarrow\)(ii)]{van2018natural} provide the following alternative characterisation: 
  \begin{equation}\label{eq:posiZeroVanArthur}
    \overline{\mathrm{\mathbf{G}}}_{G}\neq \emptyset \Leftrightarrow \posi(G)\cap \{v \in \V \colon v\leq 0\}=\emptyset
  \end{equation}
where, for any \(V\subseteq \V\),
\[
    \posi(V)\coloneqq\cset*{\sum_{j=1}^n \lambda_j v_j}{ n\in \N,v_j\in V, \lambda_j>0}.
\]
Note that \(\posi\) is a closure operator on \(\V\), which is an operator \(\cl\colon 2^\V \to 2^\V\) for which the following properties hold for any \(A,B\in 2^\V\):
\begin{property}[label=$\mathrm{cl}_{\arabic*}$., ref=$\mathrm{cl}_{\arabic*}$,start=0]
  \item\label{ax:CLext}  \(A\subseteq \cl(A)\);
  \item\label{ax:CLincreasing} if \(A\subseteq B\), then \(\cl(A)\subseteq \cl(B)\);
  \item \label{ax:CLundo} \(\cl(\cl(A))=\cl(A)\).
\end{property}
If \(G\) is consistent, then by \cite[Theorem~2,(v)]{van2018natural} we furthermore have that among all the compatible \(D\in \overline{\mathrm{\mathbf{G}}}_G\), there is a least informative---smallest---coherent set of desirable options that contains \(G\).
It is called the natural extension of \(G\) and given by \(\NN(G)\coloneqq\posi(G\cup \V_{>0})\), where \(\V_{>0}\coloneqq \{u\in \V \colon 0 < u\}\).
More formally: for any consistent \(G\), we have that \(\NN(G)\in \overline{\mathrm{\mathbf{G}}}_{G}\) and \(\NN(G)\subseteq D\) for all \(D\in \overline{\mathrm{\mathbf{G}}}_{G}\). 

For non-consistent \(G\), we also let \(\NN(G)\coloneqq\posi(G\cup \V_{>0})\), but we then no longer call it the natural extension of \(G\).
This allows for the following alternative characterisation of consistency.
\begin{lemma}\label{lem:posiZero} 
  A set of options~\(G\subseteq \V\) is consistent if and only if \(0\notin\NN(G)\).
\end{lemma}
\begin{proof}
  If \(G\) is consistent, then \(\NN(G)\in \overline{\mathrm{\mathbf{G}}}_G\) is coherent by \cite[Theorem~2,(v)]{van2018natural}, so it follows from \cref{ax:Dantisym} that \(0\notin \NN(G)\).
  So it remains to prove that \(0\in \NN(G)\) for any \(G\) that is not consistent.
  If \(G\) is not consistent, then by \cref{eq:posiZeroVanArthur}, we know that \(\posi(G)\cap \{v \in \V \colon v\leq 0\}\neq \emptyset\).
  Take any \(u\in \posi(G)\cap \{v \in \V \colon v\leq 0\}\).
  Then there are some \(n\in \N\), \(v_j\in G\) and \(\lambda_j>0\) such that \(u=\sum_{j=1}^n \lambda_j v_j \leq 0\).
  If \(u=0\) then we are done because \(0=u= \sum_{j=1}^n \lambda_j v_j \in \posi(G\cup \V_{>0})= \NN(G)\) by definition.
  If \(u\neq 0\), then \(p = -u \in \V_{>0}\) such that \(0=p+u=p+\sum_{j=1}^n \lambda_j v_j\).
  Since \(p\in \V_{>0}\), we have that \(0\in \posi(G\cup \V_{>0})= \NN(G)\).
\end{proof}
These results for coherent sets of desirable options can be immediately used in our context.
First, due to \cref{lem:percsetEquivSODO}, \(G\) is consistent if and only if \(\OOOO[G]\neq \emptyset\).
It therefore follows from \cref{lem:posiZero} that
\begin{equation}\label{eq:equivOGDGNNG}
  \OOOO[G]\neq \emptyset \Leftrightarrow \overline{\mathrm{\mathbf{G}}}_{G} \neq \emptyset  \Leftrightarrow 0\not\in \NN(G).
\end{equation}
Secondly, we can also use the \(\NN\) operator to reformulate the condition that appears in \cref{eq:CFgeneratordef}.
\begin{lemma}\label{lem:testG}
  Let \(A\in \Q\) be an option set and consider an option~\(u\in A\) and a set of options \(G\subseteq \V\).
  Then the following statements are equivalent 
  \begin{enumerate}[label=(\roman*), ref=(\roman*)]
  \item\label[statement]{equiv3} there is a preference order~\({\prec}\in\OOOO[G]\) such that for all options~\(a\in A\) we have \(u\not\prec a \);
  \item\label[statement]{equiv2} there is some \(D\in \overline{\mathrm{\mathbf{G}}}_{G}\) such that \((A-u)\cap D =\emptyset\);
  \item\label[statement]{equiv1}\((A-u)\cap \NN(G)= \emptyset\). 
  \end{enumerate}
  \end{lemma}
  \begin{proof}
    If \(G\) is not consistent then all statements are trivially false by \cref{eq:equivOGDGNNG} because \(0=u-u\in A-u\).
    So we only have to handle the case where \(G\) is consistent.
    We give a circular proof of the negation of all statements.

    First we prove that the negation of \cref{equiv3} implies the negation of \cref{equiv2}.
    Assume that for all \({\prec}\in\OOOO[G]\) there is some \(a\in A\) such that \(u\prec a\).
    Then for any \(D\in \overline{\mathrm{\mathbf{G}}}_{G}\),
    by \cref{lem:axD}, \(\prec_D\) is a preference order such that \(G_{\prec_D}=D\).
    It therefore follows from \cref{lem:percsetEquivSODO} that \(\prec_D \in \OOOO[G]\).
    Therefore, there is some \(a\in A\) such that \(u\prec_D a\).
    Hence, \(a-u\in D\), so that \((A-u)\cap D\neq \emptyset\).

    Next we prove that the negation of \cref{equiv2} implies the negation of \cref{equiv1}.
    Assume that for all \(D\in \overline{\mathrm{\mathbf{G}}}_{G}\) we have \((A-u)\cap D \neq \emptyset\).
    By \cite[Theorem~2]{van2018natural}, since \(G\) is consistent, we have that \(\NN(G)\in \overline{\mathrm{\mathbf{G}}}_{G}\), which immediately implies \cref{equiv1}.

    Finally, we prove that the negation of \cref{equiv1} implies the negation of \cref{equiv3}.
    Assume that \((A-u)\cap \NN(G)\neq \emptyset\).
    Take any \(\prec \in \OOOO[G]\).
    Since \(G\) is consistent, we have that \(\NN(G)\) is the smallest coherent set of desirable options in \(\overline{\mathrm{\mathbf{G}}}_{G}\) \cite[Theorem~2,(v)]{van2018natural}.
    Moreover, by \cref{lem:percsetEquivSODO} we have that \(G\subseteq G_{\prec}\) and, by \cref{lem:axD} that \(G_\prec \in \overline{\mathrm{\mathbf{G}}}\), so \(G_{\prec}\in \overline{\mathrm{\mathbf{G}}}_G\).
    Whence, \(\NN(G)\subseteq G_\prec\).
    Therefore, since \((A-u)\cap \NN(G)\neq \emptyset\), there is some \(a\in A\) such that \(a-u\in G_{\prec}\).
    By definition and \cref{ax:Otransla} this means that \(a\prec u\).
  \end{proof}

\subsection{Connecting back to generators}\label{sec2:ConnectionBackToGenerators}
From these results, we can now check the consistency of a generator in a straightforward way.

  \begin{lemma}\label{cor:testcons}
  A generator~\(\G\subseteq  2^\V\) is consistent if and only if there is some \(G\in \G\) such that \( 0\notin \NN(G)\).
  \end{lemma}
  \begin{proof}
    By definition, the consistency of \(\G\) is equivalent to there being some \(G\in \G\) such that \(\OOOO[G]\neq \emptyset\).
    By \cref{lem:posiZero} this is equivalent to there being some \(G\in \G\) such that \(0\notin \NN(G)\).
  \end{proof}

  Similarly, the lemmas above also lead up to the following theorem that allows us to evaluate \(C^{\G}\) for any finite option set in terms of checks on \(\NN(G)\) for the generator sets \(G\in \G\).

\begin{theorem}\label{th:natextnieuw}
  Consider any generator~$\G\subseteq 2^{\V}$.
  For any option~set~\(A\in \Q\) and option~\(u\in A\), $u\in C^{\G}(A)$ if and only if there is some \(G\in \G\) such that $(A-u) \cap \NN(G) = \emptyset$.
\end{theorem}
\begin{proof}
  By \cref{eq:CFgeneratordef}, $u\in C^{\G}(A)$ is equivalent to 
  \[(\exists G\in \G)(\exists {\prec}\in \OOOO[G])(\forall a\in A) u\not\prec a,\] 
  and this is by \cref{lem:testG} equivalent to \((\exists G\in \G)(A-u)\cap \NN(G)=\emptyset\).
\end{proof}
Interestingly, this result does not only characterise the natural extension of consistent \(\G\).
Since we know from \cref{subsec2:consNatGen} that \(\G\) is consistent if and only if \(0\in C^{\G}(\{0\})\), this result also includes \cref{cor:testcons} as a special case.


\section{Practical methods for finite generator sets}\label{sec:evalNatExt}

We've just seen that both the problem of checking the consistency of an assessment and computing its natural extension reduce to checking for multiple options \(v\in \V\) and option sets \(G\in \G\) whether \(v\) belongs to \(\NN(G)\).
This can be done in various ways, but in this section we will proceed to propose one way for the case where \(G\) is finite, and show how this leads to practical algorithms for checking the consistency and evaluating \(C^{\G}\) for generators \(\G\) that consist of a finite number of such finite generator sets \(G\).
\subsection{Checking if an option belongs to \texorpdfstring{\(\NN(G)\)}{the natural extension} for finite \texorpdfstring{\(G\)}{G}}\label{subsec:finiteG}
If an assessment \(\Ass\) is finite, which will often be the case in practice, then there are only a finite number of option sets in \(\D_{\Ass}\) each containing a finite number of options.
The disjunctive generator \(\G_{\Ass}\) will then also consist of a finite number of option sets---at most \(\prod_{H\in \D_{\Ass}} |H|\), as can be seen from \cref{eq:FiniteGdef}---each of which contains only a finite number of options.
The following proposition gives a more practical way of checking, for one such \emph{finite} option set \(G\), whether an option belongs to \(\NN(G)\).
We use the convention that for all non-negative integers \(n\geq 0\), \((\lambda_{1},\ldots,\lambda_{n})>0\) means that \((\lambda_{1},\ldots,\lambda_{n})\) is an \(n\)-tuple of real numbers such that \(\lambda_j\geq 0\) for all \(j\in \{1,...,n\}\) and \(\lambda_j>0\) for at least one \(j\in \{1,...,n\}\); in particular, for \(n=0\), no such \((\lambda_1,...,\lambda_n)>0\) exists.

\begin{proposition}\label{th:testingg}
For any option set \(G=\{g_1,...,g_\itrr\}\in \Qe\) and option~\(v\in \V\), $v\in \NN(G)$ if and only if at least one of the following two conditions holds:
\begin{enumerate}[label=(\roman*), ref=(\roman*)]
  \item\label{cond:1} \(0 < v\); 
  \item\label{cond:2} there is some $(\lambda_1,\ldots,\lambda_{\itrr}) >0$ such that $\sum_{j=1}^{\itrr} \lambda_j g_j \leq v$.
\end{enumerate}
\end{proposition}
\begin{proof}
First we prove the implication to the left.
If \ref{cond:1} holds, we have that \(v\in \V_{>0} \subseteq \posi(G\cup\V_{>0}) = \NN(G)\).
If \ref{cond:2} holds---which implies that \(\itrr>0\)---we consider two cases.
If $\sum_{j=1}^{\itrr} \lambda_j g_j = v$, then we are done by definition of the \(\posi\) operator.
Otherwise, let \(p \coloneqq v-\sum_{j=1}^{\itrr} \lambda_j g_j\in \V_{>0}\) and \(\lambda_{m+1}\coloneqq 1\).
Then \(v=\sum_{j=1}^{m} \lambda_j g_j + \lambda_{m+1} p \in \posi(G\cup\V_{>0})=\NN(G)\).

Next, we prove the implication to the right.
Since \(v\in \NN(G)\), \(v\) is a finite positive linear combination of elements of \(\{g_1,...,g_m\}\cup \V_{>0}\). 
By summing terms with the same \(g_i\) together, and adding terms \(\lambda_i g_i\) with \(\lambda_i=0\) for those \(g_i\) that do not appear in this positive linear combination, we see that there is a natural number \(\ell\geq 1\) and there are \((\lambda_1,\ldots,\lambda_{m+\ell})> 0\) and \(p_1,\ldots,p_{\ell}\in \V_{>0}\) such that  \(v= \sum_{j=1}^{m} \lambda_j g_j +\sum_{j=1}^{\ell} \lambda_{m+j} p_j\).
If \(\lambda_{1}=\ldots=\lambda_m=0\), we infer from this that \((\lambda_{m+1},\ldots,\lambda_{m+\ell})>0\) and \(0< \sum_{j=1}^{\ell} \lambda_{m+j} p_j =v\), which is \ref{cond:1}.
In the other case we have that $(\lambda_1,\ldots,\lambda_{\itrr}) >0$ and, since \(0\leq \sum_{j=1}^{\ell} \lambda_{m+j} p_j\), also that \(\sum_{j=1}^m \lambda_j g_j = v - \sum_{j=1}^{\ell} \lambda_{m+j} p_j \leq v\), which is \ref{cond:2}.
\end{proof}
For option sets that are singletons, this condition can be simplified even further.
\begin{corollary}\label{cor:testingg}
  Consider any two options \(u,v\in \V\) with \(v\neq 0\).
  Then \(v \in \NN(\{u\})\) if and only if there is some \(\lambda\geq0\) such that \(\lambda u\leq v\).
\end{corollary}
\begin{proof}
  First we prove the implication to the right.
  By \cref{th:testingg}, we have that either \(0<v\) or there is some \(\lambda>0\) such that \(\lambda u\leq v\).
  If \(0<v\), then we can choose \(\lambda=0\) and in the other case we are done.

  Next we prove the implication to the left.
  If \(\lambda=0\), then \(0<v\) since \(v\neq 0\), which is sufficient by \cref{th:testingg}~\ref{cond:1}.
  If \(\lambda>0\), then this is sufficient by \cref{th:testingg}~\ref{cond:2}.
\end{proof}
\subsection{Consistency and natural extension in practice}
With \cref{th:testingg} at our disposal, the methods in \cref{sec2:ConnectionBackToGenerators} for checking consistency and evaluating \(C^{\G}\) can now be simplified even further, at least if \(\G\) consists of finite option sets.

For consistency, we then know from \cref{cor:testcons} that \(\G\) is consistent if and only if there is some \(G\in \G\) such that \(0\notin \NN(G)\).
Reformulating \(0\notin \NN(G)\) with \cref{th:testingg}, we arrive at the following result.
\begin{proposition}\label{prop:testingconsPractice}
  Consider a generator~\(\G\subseteq \Qe\) consisting of finite option sets.
  Then \(\G\) is consistent if and only if there is some \(G=\{g_1,...,g_\itrr\}\in \G\) such that \(\sum_{j=1}^{\itrr} \lambda_j g_j \not\leq 0\) for all \((\lambda_1,...,\lambda_{\itrr})>0\).
\end{proposition}
\begin{proof}
Combining \cref{cor:testcons,th:testingg}, we see that \(\G\) is consistent if and only if there is some \(G=\{g_1,...,g_\itrr\}\in \G\) for which \cref{th:testingg}~\ref{cond:1} and \ref{cond:2} are both false for \(v=0\). 
We see immediately that \ref{cond:1} is always false because \(0\not<0\), which leaves us with the requirement that \ref{cond:2} should be false for \(v=0\).
\end{proof}
To illustrate this, we return to our running example.

\begin{myexp}\label{ex:1.5}
Although we have already found that the strategic advisor's assessment is consistent, in \cref{ex:1.4a}, it required us to come up with a specific preference order to demonstrate this.
With \cref{prop:testingconsPractice}, we now have a more systematic approach at our disposal.
Indeed, we know from \cref{subsec2:consNatGen} that \(\Ass\) is consistent if and only if \(\G_{\Ass}\) is.
Therefore, by \cref{prop:testingconsPractice}, we can demonstrate consistency by finding a generator set $G=\{g_1,...,g_m\} \in \A$ such that $\sum_{j=1}^m \lambda_j g_j\not\leq 0$ for every $(\lambda_{1},...,\lambda_{m})>0$.
We will use the particular generator set $G=\{g_1,g_2,g_3\}=\{h_2,h_4,h_5\}=\{(2,-1),(5,-3),(-7,7)\}\in \A$. 
Assume \emph{ex absurdo} that there is some $(\lambda_{1},\lambda_2,\lambda_{3})>0$ such that
\(
\lambda_1 g_1+\lambda_2 g_2 + \lambda_3 g_3\leq 0.
\)
Notice that $2g_2\leq 5g_1$, so if we let $\mu_1\coloneqq\frac25\lambda_1+\lambda_2$ and $\mu_2\coloneqq7\lambda_3$ then  
\[
\sum_{j=1}^3 \lambda_j g_j\geqslant\frac25\lambda_1 g_2+\lambda_2 g_2 +\lambda_3  g_3 =\mu_1 h_4 +\frac17 \mu_2  h_5=
(5 \mu_1 - \mu_2,
-3 \mu_1 + \mu_2)
.
\]
Since $\sum_{j=1}^3 \lambda_j g_j \leq 0$, this implies that $5\mu_1\leq\mu_2\leq 3\mu_1$ and thus $\mu_1\leq0$ and $\mu_2\leq0$.
This is impossible though because $(\lambda_{1},\lambda_2,\lambda_{3})>0$ implies that $\mu_1>0$ or $\mu_2>0$.
Hence, $\Ass$ is consistent, confirming our earlier findings of \cref{ex:1.4a}.
\hfill\pushQED{$\lozenge$}\popQED\end{myexp}

To determine $C^{\G}(A)$ for any option set $A\in\Q$, we can check for every individual $u\in A$ if $u\in C^\G(A)$.
Due to \cref{th:natextnieuw}, this requires us to check, for every $u\in A$, if there is some $G\in \G$ such that $(A-u)\cap \NN(G)=\emptyset$, or equivalently, such that \(v\notin \NN(G)\) for all \(v\in A-u\).
If the generator sets in \(\G\) are all finite, then we can reformulate this condition using \cref{th:testingg}.
\begin{proposition}\label{prop:testinggPractice}
For any finite generator~\(\G\Subset \Qe\), option set~\(A\in \Q\) and option~\(u\in A\), $u\in C^{\G}(A)$ if and only if 
\begin{enumerate}
  \item\label{cond:1prac} \(v\not>0\) for all \(v\in A-u\), and
  \item\label{cond:2prac} there is some \(G=\{g_1,...,g_m\}\in \G\) such that \(\sum_{j=1}^m\lambda_j g_j \not\leq v\) for all \(v\in A-u\) and \((\lambda_1,...,\lambda_m)>0\).
\end{enumerate}
\end{proposition}
\begin{proof}
As explained in the text preceding this result, it follows from \cref{th:natextnieuw} that \(u\in C^{\G}(A)\) if and only if there is some \(G=\{g_1,...,g_m\}\in \G\) such that \(v\notin \NN(G)\) for all \(v\in A-u\).
By \cref{th:testingg}, the condition \(v\notin \NN(G)\) is true if both \ref{cond:1}  \(v\not>0\) and  \ref{cond:2} \(\sum_{j=1}^m\lambda_j g_j \not\leq v\) for all \((\lambda_1,...,\lambda_m)>0\).
Since \ref{cond:1} does not depend on \(G\), this implies the stated result.
\end{proof}
We again illustrate this using our running example.
\begin{myexp}\label{ex:1.6}
We can now finally tackle the whole problem at the end of 
\cref{ex:1.3}: choosing from the sets $A_3=\{(-3,4),\allowbreak (0,1),\allowbreak (4,-3)\}$ and \(A_4=\{(-2,2),(5,-4)\}\) based on the expert's assessment.
Since \(C_{\Ass}(A_4)=A_4\) was already evaluated in \cref{ex:1.4b}, we focus on computing $\NatExt(A_3)$, or equivalently, as explained in \cref{subsec2:consNatGen}, computing $C^{\G_{\Ass}}(A_3)$.
To see if $u=(4,-3)$ is chosen from $A_3$, as we know from \cref{prop:testinggPractice}, we need to check two things. 
First, whether \(v\not>0\) for all \(v \in A_3-u =\{(-7,7),(-4,4),(0,0)\}\); this is clearly the case.
Second, whether there is some set~$G=\{g_1,...,g_m\} \in \A$ such that $\sum_{j=1}^m \lambda_j g_j \not\leq v$ for all $v\in A_3-u$ and \((\lambda_1,...,\lambda_{m})>0\).
This is not the case because for every \(G=\{g_1,g_2,g_3\}\in \A\),\footnote{We have already seen in \cref{ex:1.4} that in our example, \(m=3\) for all \(G\in \A\).} since \(g_3=h_5=(-7,7)\), we find for $(\lambda_1,\lambda_2,\lambda_{3})=(0,0,1)$ and $v=(-7,7)$ that $\sum_{j=1}^3 \lambda_j g_j = h_5 = (-7,7)= v$.
So $(4,-3)$ is not chosen from \(A_3\).
Checking if $(0,1)$ is chosen is analogous.
In this case \(u=(0,1)\) and \(A_3-u =\{(-3,3),(0,0),(4,-4)\}\), so a similar argument using $v=(-3,3)$ and $(\lambda_1,\lambda_2,\lambda_{3})=(0,0,\frac37)$ leads us to conclude that $(0,1)$ is not chosen either.
From \cref{ex:1.4b}---and because $\NatExt(A_3)$ must contain at least one option by the consistency of $\Ass$ (see \cref{ex:1.4a,ex:1.5}) and \cref{prop:niet-leeg}---it follows that $\NatExt(A_3)=\{(-3,4)\}$.
So based on the advisor's earlier decisions, it follows that the company should choose $(-3,4)$ from $A_3$.
In contrast, since \(C_{\Ass}(A_4)=A_4\), the earlier decisions contain insufficient information to determine which option should be chosen from $A_4$.
\hfill\pushQED{$\lozenge$}\popQED\end{myexp}
In this simple toy example, the assessment $\Ass$ was small and the conditions in \cref{prop:testingconsPractice,prop:testinggPractice} could be checked manually.
In realistic problems, this may not be the case though.
We therefore proceed to develop a more algorithmic approach.

\subsection{An algorithmic approach}\label{sec:algchoose}

Fix some finite generator~\(\G \Subset  \Qe\). 
Then as we've seen in the previous section, for each \(G=\{g_1,...,g_m\}\in \G\), the step `check if there is some tuple~\((\lambda_1,...,\lambda_m)>0\) for which \(\sum_{j=1}^m\lambda_j g_j \leq v\)' is an essential element of our methods for checking the consistency of \(\G\) and evaluating \(C^\G\).
For that reason, we introduce a boolean function \isF$\colon \Qe\times \V\to\{\text{true},\text{false}\}$ for it. For every $\{g_1,...,g_m\}\in\Qe$ and $v\in\V$, it returns $\text{true}$  if $\sum_{j=1}^m\lambda_j g_j \leq v$ for at least one $(\lambda_1,...,\lambda_m)>0$, and $\text{false}$ otherwise.

Since we will need to evaluate \isF  ~repeatedly, we first look at how we can do this in practice.
By definition, $(\lambda_1,\dots,\lambda_m)>0$ can be rewritten as $\lambda_j\geq0$ for all $j\in\{1,\dots,m\}$ and $\sum_{j=1}^m \lambda_j>0$, which are all linear constraints.
Since the condition $\sum_{j=1}^m \lambda_j g_j \leq v$ is linear as well, we have a linear feasibility problem to solve.
However, strict inequalities such as $\sum_{j=1}^m \lambda_j>0$ are problematic for software solvers for linear feasibility problems.
A quick fix is to choose some very small $\epsilon>0$ and impose the inequality $\sum_{j=1}^m \lambda_j\geq \epsilon$ instead, but since this is an approximation, it does not guarantee that the result is correct.
A better solution is to use the following alternative characterisation that, by introducing an extra free variable, avoids the need for strict inequalities.\footnote{This result is a special case of what Quaeghebeur did for his analysis preceding the {CONE}strip algorithm \cite[Section~2]{quaeghebeur2013conestrip}, where \(\NN(G)\) is in his notation \(\underline{\mathcal{R}}\).} 
\begin{proposition}\label{prop:hunt}
Consider any $v\in \V$ and any $G\coloneqq\{g_1,\dots,g_m\}\in\Qe$.
Then \isF$(G,v)=\text{true}$ if and only if there is some $(\mu_1,\dots,\mu_{m+1})\in \R^{m+1}$ such that $\sum_{k=1}^m \mu_k g_k \leq \mu_{m+1} v$, $\mu_k\geq 0$ for all $k\in\{1,\dots,m\}$, $\mu_{m+1}\geq 1$ and $\sum_{k=1}^{m} \mu_k\geq 1$.
\end{proposition}
\begin{proof}
  Suppose \isF$(G,v)=\text{true}$.
  Then there is some
  $(\lambda_1,...,\lambda_m)>0$ such that $\sum_{k=1}^m \lambda_k g_k \leq v$.
  Let
  \[
    (\mu_1,...,\mu_{m+1})\coloneqq
  \begin{cases}
  (\lambda_1, \dots , \lambda_m,1) &\text{if }\sum_{j=1}^m \lambda_j\geq 1,\\
  \left(\frac{\lambda_1}{\sum_{j=1}^m \lambda_j}, \dots , \frac{\lambda_m}{\sum_{j=1}^m \lambda_j},\frac{1}{\sum_{j=1}^m \lambda_j}\right) &\text{otherwise.}
  \end{cases}
  \] Then \((\mu_1,...,\mu_{m+1})\) satisfies the conditions in the statement.

  Next we prove the implication to the left.
  So suppose that there is some \((\mu_1,...,\mu_{m+1})\in \R^{m+1}\) such that $\sum_{k=1}^m \mu_k g_k \leq \mu_{m+1} v$, $\mu_k\geq 0$ for all $k\in\{1,\dots,m\}$, $\mu_{m+1}\geq 1$ and $\sum_{k=1}^{m} \mu_k\geq 1$.
  Let $(\lambda_1,...,\lambda_n)\coloneqq(\frac{\mu_1}{\mu_{m+1}},\dots,\frac{\mu_m}{\mu_{m+1}} )$. 
  Then clearly \((\lambda_1,...,\lambda_m)>0\) and \(\sum_{k=1}^m \lambda_k g_k \leq v\), so \textsc{IsFeasible}$(G,v)=\text{true}$.
  \end{proof}

Computing \isF$(\{g_1,...,g_n\},v)$ is therefore a matter of solving the following linear feasibility problem: 
\begin{equation*}
  \begin{aligned}
    \text{find}\quad        & \mu_1,\dots,\mu_{n+1}\in\R, \\
    \text{subject to\quad}  & \mu_{n+1} v(x)-\textstyle\sum_{k=1}^n \mu_k g_k(x)\geq 0  &&\text{for all } x\in\X,\\
                      & \textstyle\sum_{k=1}^n \mu_k\geq 1,\quad \mu_{n+1}\geq1, \\
                  \quad  & \mu_k\geq0    &&\text{for all } k\in\{1,\dots,n\}.
  \end{aligned}
\end{equation*}
If \(n=0\) the sum in the second constraint is empty and therefore zero, resulting in the requirement that \(0\geq 1\), which is always false.
We will assume henceforth that you have a method for solving such linear feasibility problems.
For finite $\X$, which we will consider in our experiments in \cref{sec:experiment}, such problems can easily be solved by standard linear programming methods; see for example \cite{linprog2006,vaidya1989speeding}.

Now, if the generator~\(\G\) itself is also finite, then we can use \isF~ to automate \cref{prop:testingconsPractice,prop:testinggPractice}.
Consider any finite generator \(\G\Subset\Qe\).
Then, by \cref{prop:testingconsPractice}, we can check the consistency of \(\G\) by iterating over all \(G\in \G\) and checking whether \isF$(G,0)$ is false; if we find at least one such \(G\), we conclude that \(\G\) is consistent.
This results in the pseudocode in \cref{alg:CheckConsistency}.
Similarly, we can evaluate \(C^{\G}\) using \cref{prop:testinggPractice}, as is done in \cref{alg:Choose}.
The algorithm starts by iterating over all \(v\in A-u\) and checking whether \(v>0\); if we find one such \(v\) we conclude that \(u\) is not chosen.
Next, it checks the second condition of \cref{prop:testinggPractice}.
So, it searches for some \(G\in \G\) for which \isF$(G,v)$ is false for all \(v\in A-u\). 
Once we find that \isF$(G,v)$ is true for some \(v\in A-u\), then we do not have to check the remaining options in \( A-u\) that we have not checked yet, and we can move on to the next \(G\).
If we have found a \(G\) for which \isF$(G,v)$ is false for all \(v\in A-u\), then we conclude that \(u\) is chosen.
If, in the end, we have checked all \(G\in \G\), and we did not find such a \(G\), then we conclude that \(u\) is not chosen.

In practice, we are however not interested in the consistency and natural extension of a generator but in that of an assessment~\(\Ass\).
To that end, as explained in \cref{subsec2:consNatGen}, it suffices to convert this assessment \(\Ass\) to the conjunctive generator \(\HH_{\Ass}\) and then to the corresponding disjunctive generator \(\A\) and consider the consistency and natural extension of the latter.
If \(\A\) consists of finitely many finite option sets ---which it will, as we know from the beginning of \cref{subsec:finiteG}, if \(\Ass\) is finite---we can then  apply \cref{alg:CheckConsistency,alg:Choose} to \(\A\) to check the consistency of \(\Ass\) or evaluate its natural extension \(C_{\Ass}\), respectively.
To transform \(\Ass\) to \(\HH_\Ass\) and \(\A\), we present two simple additional algorithms.
 \cref{alg:AssToConjAkaPerineumNaive} converts an assessment \(\Ass\) to the corresponding conjunctive generator \(\D_{\Ass}\) as defined in \cref{sec:charactOOOAss}.
\cref{alg:ConjoToDisjNaive} converts a conjunctive generator \(\HH\)---such as \(\HH_{\Ass}\)---to the corresponding disjunctive generator \(\G(\HH)\), as given by \cref{eq:FiniteGdef}.
For \(\HH=\emptyset\)---which is the case for \(\HH_{\Ass}\) if \(\Ass=\emptyset\), or may happen as a result of the simplifications in \cref{sec:simp}---we know from  \cref{foot:4} that \(\G(\HH)=\{\emptyset\}\), which is why \cref{alg:ConjoToDisjNaive} initialises \(\G\) as such.
We call both algorithms naive because, as we will see in \cref{sec:simp,sec:simpGen}, both can be improved upon.

\begin{algorithm}
  \caption{Check if a generator is consistent \label{alg:CheckConsistency}}
  \begin{algorithmic}[1]
  \Require{finite generator \(\G\Subset\Qe\)}
  \Ensure{\(\text{true}\) if \(\G\) is consistent and \(\text{false}\) otherwise}
      \ForAll{$G\in \G$}
        \If{\textbf{not }\isF$(G,0)$}
          \State\Return{$\text{true}$}
        \EndIf
      \EndFor
      \State\Return{$\text{false}$}
  \end{algorithmic}
  \end{algorithm}

\begin{algorithm}
  \caption{Decide if an option~\(u\) is chosen from \(A\) by \(C^{\G}\) \label{alg:Choose}}
  \begin{algorithmic}[1]
  \Require{option set $A\in  \Q$, option \(u\in A\), finite generator \(\G\Subset\Qe\)}
  \Ensure{\(\text{true}\) if \(u\in C^{\G}(A)\) and \(\text{false}\) otherwise}
        \ForAll{$v\in A-u$}
        \If{$v>0$}
          \State\Return{$\text{false}$}
        \EndIf
      \EndFor
      \ForAll{$G\in \G$}
        \Let{Found-v}{$\text{false}$}
        \ForAll{$v\in A-u$}
            \If{\isF$(G,v)$}
              \State{Found-v $\gets \text{true}$}
              \State{\textbf{break}} 
            \EndIf
        \EndFor
        \If{\textbf{not} Found-v} 
          \State\Return{$\text{true}$}
        \EndIf
      \EndFor
      \State\Return{$\text{false}$}
  \end{algorithmic}
  \end{algorithm}

\begin{algorithm}
  \caption{From assessment to conjunctive generator \label{alg:AssToConjAkaPerineumNaive}}
  \begin{algorithmic}[1]
  \Require{finite assessment \(\Ass\Subset \Q\times \Qe\)}
  \Ensure{finite conjunctive generator \(\D\coloneqq \D_{\Ass}\Subset \Q\)}
  \Function{AssessmentToDisjunctiveNaive}{$\Ass$}
      \Let{$\D$}{$\emptyset$}
      \ForAll{$(V,W)\in \Ass$}
       \ForAll{$w\in W$}
          \Let{$\D$}{$\D \cup \{V - w\}$}
        \EndFor
      \EndFor
      \State\Return{\(\D\)}
  \EndFunction
  \end{algorithmic}
  \end{algorithm}

  \begin{algorithm}
    \caption{From conjunctive to disjunctive generator \label{alg:ConjoToDisjNaive}}
    \begin{algorithmic}[1]
    \Require{finite conjunctive generator \(\D\Subset \Qe\)}
    \Ensure{finite disjunctive generator \(\G\coloneqq \G(\D)\Subset \Qe\)}
    \Function{ConjunctiveToDisjunctiveNaive}{$\D$}
        \Let{$\G$}{$\{\emptyset\}$}
        \ForAll{$H \in \D$}
          \Let{$\G^*$}{$\emptyset$}
          \ForAll{$G \in \G$}
            \ForAll{$h \in H$}
              \Let{$\G^*$}{$\G^*\cup \{G\cup \{h\}\}$}
            \EndFor
          \EndFor
          \Let{$\G$}{$\G^*$}
        \EndFor
      \State\Return{$\G$}
    \EndFunction
    \end{algorithmic}
    \end{algorithm}

\subsection{Practical considerations}\label{subsec:practicalConsiderations}
If we want to implement \cref{alg:CheckConsistency,alg:Choose,alg:AssToConjAkaPerineumNaive,alg:ConjoToDisjNaive} in the programming language of your choice, there are some practical aspects to take into account.

A first such aspect is which data type to use to store generators in all the previous algorithms.
One option, that is closest to the mathematical descriptions in our pseudocode, is to use the set data type, which is implemented in most common programming languages.
However, we opt to use arrays; that is, we store option sets as arrays and sets of option sets as arrays of arrays.
One reason is that arrays are usually optimised for iteration, which is what we do in \cref{alg:CheckConsistency,alg:Choose}.
Another advantage of arrays is that they use slightly less memory when no duplicates are present, because set-like structures save some additional information to maintain their hash table.
The advantages of set-like structures, such as faster look-up and removal, on the other hand, are not of use in our algorithms.
Moreover, the only difference is that for arrays it is possible that duplicates are present, and it can be seen from \cref{prop:testingconsPractice,prop:testinggPractice} that even when there are duplicate option sets, this would just lead us to check the same option set twice, which will not alter the result.
Similarly, inside the option sets \(G\in \G\) it will not matter to have the same option twice, as the \(\lambda\)'s for the same options can be lumped together.
Since it is often a reasonable assumption that duplicates will be very rare, this leads us to think that the advantages of arrays outweigh the disadvantages here.

A second particular aspect is that \cref{alg:ConjoToDisjNaive} will lead to memory explosion for large conjunctive generators if we save the disjunctive generator in memory.
This is because \(|\G(\HH)|=|\times_{H\in \HH} H|\) for a given conjunctive generator \(\HH\) (when using arrays).
Therefore, it is in practice better to only save the conjunctive generator in memory and do the iteration over the disjunctive generator on the fly\footnote{In Julia and Python this is done by iterating over `Itertools.product' of \(\HH\), for example.} when we need it in \cref{alg:CheckConsistency,alg:Choose}.

We will take both considerations into account in our implementation when we will test our algorithms in \cref{sec:experiment}.
However, the exponential explosion can still manifest itself in the running time of \cref{alg:CheckConsistency,alg:Choose}.
To address this potential issue, in the next two sections, we will develop methods to reduce the size of a (conjunctive or disjunctive) generator, without altering the corresponding set of preference orders.
We start with conjunctive generators.

\section{Simplifying conjunctive generators}\label{sec:simp}

For any given conjunctive generator \(\HH\Subset \Qe\)---and for \(\HH_{\Ass}\) in particular---the consistency and natural extension of the corresponding generator \(\G(\HH)\)---and that of \(\A=\G(\HH_\Ass)\) in particular---are fully determined by the corresponding set of preferences \(\OOOO(\G(\HH))=\OOO(\HH)\).
Therefore, if we can simplify \(\HH\) without altering \(\OOO(\HH)\), this will reduce the running time of \cref{alg:CheckConsistency,alg:Choose,alg:ConjoToDisjNaive} without altering the result.
The aim of this section is to achieve such simplifications, either by removing option sets from \(\HH\) or by removing options inside the individual option sets in \(\HH\).
At the end we also present an algorithm that implements these simplifications.


\subsection{Removing option sets containing a positive option}
First we show that we can sometimes remove a whole option set.
In particular, for any conjunctive generator \(\D\), we can remove all option sets \(H\in \D\) for which \(\OOO[H]=\OOO\), as doing so does not alter \(\OOO(\D)=\bigcap_{H\in \D} \OOO[{H}]\).
The following lemma gives a simple necessary and sufficient condition for this to happen.
\begin{lemma}\label{prop:simpDer2}
  Consider an option set~\(H\in \Q\). Then \(\OOO[H]=\OOO\) if and only if there is an option \(h\in H\) such that \(0<h\).
\end{lemma}
\begin{proof}
  First we prove the implication to the right.
  Suppose that \(\OOO[H]=\OOO\), then \(<\in \OOO=\OOO[H]\).
  So by definition there is some \(h\in H\) such that \(0<h\).

  Next we prove the implication to the left.
  Since \(0<h\) and any \({\prec} \in \OOO\) extends \(<\) by \cref{ax:Oext}, we have \(\OOO_h=\OOO\).
  Therefore, by \cref{eq:OOODunieOverOd}, we have that \(\OOO[{H}]=\bigcup_{h\in H} \OOO_h = \OOO\).
\end{proof}



 
\subsection{Removing negative options}
The next simplification is based on the idea that some options \(u\in H\) are non-informative.
In particular, if an option \(u\in H\) is nonpositive (\(u\leq 0\)), then it cannot be preferred to zero by a preference order since preference orders extend \(<\), so \(\OOO_u=\emptyset\).
Removing \(u\) from \(H\) will therefore not change \(\OOO[H]\).
This is made formal in the following lemma.
\begin{lemma}\label{prop:simpDer1}
  Consider an option set~\(H\in \Q\) and an option \(u\in H\). 
  If \(u\leq 0\), then \(\OOO[{H}]=\OOO[{H\setminus\{u\}}]\).
  \end{lemma}
  \begin{proof}
    We will first show that \(\OOO_u=\emptyset\) if \(u\leq 0\).
    Suppose \emph{ex absurdo} that there is some \(\prec\in \OOO_u\).
    Then by definition of \(\OOO_u\), \(0\prec u\).
    If \(u=0\) this is already a contradiction with \cref{ax:Oantisym}.
    If \(u<0\) then by \cref{ax:Oext} it follows that \(u\prec 0\prec u\).
    By \cref{ax:Otransi} then \(u\prec u\), which is a contradiction with \cref{ax:Oantisym}.

    Therefore, by \cref{eq:OOODunieOverOd}, we have that
    \[
      \OOO[{H}]=\bigcup_{h\in H} \OOO_h=\bigcup_{h\in H\setminus\{u\}} \OOO_h=\OOO[{H\setminus\{u\}}]. \qedhere
    \]
  \end{proof}

  It is possible that all options in \(H\) are nonpositive.
  We can then remove all of them, which leads to \(H=\emptyset\) and \(\OOO[H]=\bigcup_{h\in \emptyset} \OOO_h=\emptyset\).
  For the corresponding disjunctive generator~\(\G(\HH)\), this implies that 
  \begin{equation}\label{eq:inconsistentDerivedAssessment}
    \OOO(\G(\HH))=\OOO(\D)=\bigcap_{H\in \D} \OOO[{H}]=\emptyset,
  \end{equation}
  so we can conclude that \(\G(\HH)\) is not consistent.

  We can extend the idea of \cref{prop:simpDer1} of removing negative options to removing options that are `dominated' by another option in the same set; this is what we will look at next.

  \subsection{Removing dominated options}
  If we know that there are two options \(u,v\in H\) such that \(\OOO_u \subseteq \OOO_v\), then we can remove \(u\) from \(H\) because it does not contribute to the union in the definition of \(\OOO[H]\).
  The following lemma gives a practical sufficient condition for this to happen.
  \begin{lemma}\label{prop:simpDer}
    Consider an option set~\(H\in \Q\) and two distinct options \(u,v\in H\). 
    If \(v\in \NN(\{u\})\), then \(\OOO[{H}]=\OOO[{H\setminus\{u\}}]\).
  \end{lemma}

    
    
  A direct proof using \cref{cor:testingg} (and \cref{prop:simpDer1} if \(v=0\)) is fairly straightforward.
  We instead opt to prove it using two more abstract lemmas though, because these will prove useful later on as well.
  As shown by \citet{morgado1962characterization}, the first lemma implies that \(\NN\) is a closure operator, like \(\posi\).
  \begin{lemma}\label{lem:posiClosureOperatorEquivalence}
    Consider two sets of options \(A,B\subseteq \V\).
    Then \(A\subseteq \NN(B)\) if and only if \(\NN(A)\subseteq \NN(B)\).
  \end{lemma}
  \begin{proof}
    First, assume that \(A\subseteq \NN(B)\).
    Take any \(u\in \NN(A)\).
    Then by definition of \(\NN(A)\) there are \(n\in \N\) and \(v_j\in A\cup\V_{>0}\) and \(\lambda_j>0\) such that \(u=\sum_{j=1}^{n} \lambda_j v_j \).
    Since \(A\subseteq \NN(B)\), we have for every \(v_j\in A\), by definition of the \(\posi\) operator, that there are \(n_j \in \N\), \(w_{j,k}\in B\cup\V_{>0}\) and \(\lambda_{j,k}>0\) such that \(v_{j}=\sum_{k=1}^{n_j} \lambda_{j,k} w_{j,k}\). 
    The same is true for \(v_j\in \V_{>0}\) if we choose \(n_j=1\) and \(\lambda_{j,1}=1\).
    Therefore, we have that
    \[
      u=\sum_{j=1}^n \lambda_j \left( \sum_{k=1}^{n_j} \lambda_{j,k} w_{j,k}  \right)
      =\sum_{j=1}^n \sum_{k=1}^{n_j} (\lambda_j \lambda_{j,k}) w_{j,k}
    \]
    Since \(w_{j,k}\in B\cup \V_{>0}\) and \((\lambda_j \lambda_{j,k})>0\), we have by definition of the \(\posi\) operator that \(u\in \NN(B)\).

    Next, assume that \(\NN(A)\subseteq \NN(B)\).
    Take any \(a\in A\).
    Using \(n=1\), \(\lambda_1=1\) and \(a \in A\cup \V_{>0}\) in the definition of \(\NN(A)\) we have that \(a\in \NN(A)\subseteq \NN(B)\).
    Therefore, \(A\subseteq \NN(B)\).
  \end{proof}
\begin{lemma}\label{cor:MostConservativePrec}
  Consider two option sets \(G_1,G_2\subseteq 2^{\V}\). 
  If \(\NN(G_2)\subseteq \NN(G_1)\), then \(\OOOO[G_1]\subseteq \OOOO[G_2]\).
\end{lemma}
\begin{proof}
  Take any \(\prec \in \OOOO[G_1]\).
  By \cref{lem:percsetEquivSODO} we have that \(G_1\subseteq G_{\prec}\).
  Since \(G_\prec\) is coherent because of \cref{lem:axD}, this implies that \(G_1\) is consistent.
  By \cite[Theorem~2]{van2018natural} \(\NN(G_1)\) is therefore the smallest coherent set of desirable options that contains \(G_1\).
  Since \(G_1\subseteq G_{\prec}\) and \(G_\prec\) is coherent, this implies that \(\NN(G_1)\subseteq G_\prec\).
  On the other hand, since \(\NN(G_2)\subseteq \NN(G_1)\) we know from \cref{lem:posiClosureOperatorEquivalence} that \(G_2\subseteq \NN(G_1)\).
  Hence, \(G_2\subseteq \NN(G_1)\subseteq  G_{\prec}\).
  Therefore, \(\prec \in \OOOO[G_2]\) by \cref{lem:percsetEquivSODO}.
\end{proof}
  \begin{proofof}{\cref{prop:simpDer}}
    By \cref{lem:posiClosureOperatorEquivalence,cor:MostConservativePrec} we have that \(\OOOO[\{u\}]\subseteq \OOOO[\{v\}]\).
    By definition this is the same as \(\OOO_u\subseteq \OOO_v\).
    Therefore, by \cref{eq:OOODunieOverOd}, we have that
      \[
        \OOO[{H}]=\bigcup_{h\in H} \OOO_h=\bigcup_{h\in H\setminus\{u\}} \OOO_h=\OOO[H\setminus\{u\}],
      \]
      as required.
  \end{proofof}

  In the case of our running example, all the option sets in $\D_{\Ass}$ can even be reduced to singletons.
  \begin{myexp}\label{ex:1.7}
  In \cref{ex:1.4} we determined that $\D_{\Ass}=\{\{h_1,h_2\},\{h_3,h_4\},\{h_5\}\}$, where \(h_1=(4,-2)\), \(h_2=(2,-1)\), \(h_3=(7,-4)\), \(h_4=(5,-3)\) and \(h_5=(-7,7)\).
  For $\{h_1,h_2\}$ we see that $2 h_2 =h_1$ and in $\{h_3,h_4\}$ we see that $\frac75 h_4 \leq  h_3$.
  So, by \cref{prop:simpDer,cor:testingg}, we can simplify the conjunctive generator $\D_{\Ass}$ to 
  $\D\coloneqq\{\{h_2\},\{h_3\},\{h_5\}\}$ without altering the corresponding set of preferences \(\OOO(\D_{\Ass})=\OOO(\D)\).
  \hfill\pushQED{$\lozenge$}\popQED\end{myexp}

  The generator in this example is small, making it easy to check which options can be removed.
  In general, however, it would be useful to automate this process, which is what we now set out to do.
  
  Inspired by \cref{prop:simpDer}, we introduce the binary relation \(\trianglelefteq\) on \(\V\) defined by \(u\trianglelefteq v \Leftrightarrow v\in \NN(\{u\}) \).
  Since we know from \cref{lem:posiClosureOperatorEquivalence} that \(u\trianglelefteq v\) is equivalent to \(\NN(\{v\})\subseteq \NN(\{u\})\) the relation \(\trianglelefteq\) is a (non-strict) preorder since it is clearly reflexive and transitive.
  To describe  \(\trianglelefteq\), we use the function \textsc{optionOrd}$\colon \V^2 \to \{\text{$\text{true}$},\text{$\text{false}$}\}$ that returns $\text{true}$ for a given option pair \((u,v)\in \V^2\) whenever  \( u \trianglelefteq v\).
  Due to \cref{th:testingg}, \textsc{optionOrd}$(u,v)$ is true if \(0<v\) or is otherwise equal to \(\isF(\{u\},v)\);
  this allows us to easily evaluate it in practice.
  Since we know from \cref{prop:simpDer} that elements of \(H\) that are dominated according to \(\trianglelefteq\) can be removed from \(H\), the best we can do is replace \(H\) by a subset that contains no such dominated elements but still generates the same set of preference orders~\(\OOO[H]\).
  Such a set is minimal in the sense that it cannot be reduced further by removing more options from it while keeping the set of preference orders identical.

  \cref{alg:Max} is one way to construct such a minimal set.
  The algorithm works for any (non-strict) preorder, allowing us to use it for other purposes in \cref{sec:simpGen} as well.
  To obtain a minimal set, it selects exactly one option from every equivalence class of options that are dominated by each other but by no other option, where two elements \(u,v\) are equivalent with respect to a preorder \(\trianglelefteq\) whenever \(u \trianglelefteq v\)  and \(v\trianglelefteq u\).
  It works by starting with an empty set \(S_{\text{max}}\) and adding options to it, one by one, that are not dominated by any other option in \(S_{\text{max}}\).
  Every time we add such an option to \(S_{\text{max}}\), we also remove the other options from \(S_{\text{max}}\) that are dominated by this new option.
  Applied to a set \(S\) with \(n\) elements, \cref{alg:Max} requires at most \({n(n-1)}\) binary comparisons between elements of \(S\).

  In particular, starting from a conjunctive generator \(\D\), we can replace it by the simplified conjunctive generator~\(\{\textsc{Max}(H,\textsc{optionOrd})\colon H\in \D\}\), where,  for every option set \(H\in  \D\), \(\textsc{Max}(H,\textsc{optionOrd})\) uses \textsc{optionOrd} at most \({|H|(|H|-1)}\) times.
    \begin{algorithm}
      \caption{Find `maximal' elements of a preordered set}\label{alg:Max}
      \begin{algorithmic}[1]
      \Require{ a preordered finite set \(S\), a comparison function \(f\colon S\times S \to \{\text{true},\text{false}\}\) such that \(f(s,t)=\text{true}\)  whenever \(s\) is dominated by \(t\) in the preorder}
      \Ensure{`maximal' elements of \(S\) with only one of every equivalence class}
      \Function{Max}{$S$, $f$}
        \Let{\(S_\text{max}\)}{\(\emptyset\)}
        \ForAll{$s\in S$}
          \Let{IsMaximal}{$\text{true}$}
          \ForAll{$t\in S_{\text{max}}$} 
            \If{$f(s,t)$}
              \Let{IsMaximal}{$\text{false}$}
              \State{\textbf{break}}
            \EndIf
          \EndFor
          \If{IsMaximal}
            \Let{\(S_{\text{new}}\)}{\(\{s\}\)}
            \ForAll{$t\in S_{\text{max}}$} 
              \If{not $f(t,s)$}
                \Let{$S_{\text{new}}$}{$S_{\text{new}}\cup\{t\}$}
              \EndIf
            \EndFor
            \Let{$S_{\text{max}}$}{$S_{\text{new}}$}
          \EndIf
        \EndFor
        \State\Return{$S_{\text{max}}$}
      \EndFunction
      \end{algorithmic}
      \end{algorithm}
  
  Due to \cref{prop:simpDer}, this will not alter the corresponding set of preferences.
  \begin{corollary}\label{prop:simpDerForReal}
    For any option set \(H\in \Qe\),  \[\OOO[H]=\OOO[\textsc{Max}(H,\textsc{optionOrd})].\]
  \end{corollary}
  \begin{proof}
    Since \cref{alg:Max} only removes options that are dominated according to \(\trianglelefteq\), this follows from repeated application of \cref{prop:simpDer}. 
  \end{proof}

We end this section on simplifying conjunctive generators by presenting an algorithm that uses these simplifications to simplify the conjunctive generator \(\D_{\Ass}\) of an assessment \(\Ass\). 
Combining the definition of \(\D_{\Ass}\) and the simplifications in \cref{prop:simpDer1,prop:simpDer2,prop:simpDer}, we get the procedure in  \cref{alg:CreateDer}.
It works by iterating over all pairs \((V,W)\in \Ass\) and all \(w\in W\).
For each such \(w\) it iterates over, the algorithm collects all \(v-w\not\leq 0\), with \(v\in V\), into an option set \(H\)---because if \(v-w\leq 0\), we can remove it by \cref{prop:simpDer1}---unless \(v-w>0\) for some \(v\in V\), in which case the option set \(H\) can be removed by \cref{prop:simpDer2}.
After creating such an option set~\(H\) we simplify it further by removing dominated options, using \cref{alg:Max}, and add it to \(\HH\).
It could be that we add \(H=\emptyset\) at some point---if \(v-w\leq 0\) for all \(v\in V\)---in which case, whatever the other option sets in \(\HH\) are, we know from \cref{eq:inconsistentDerivedAssessment} that \(\OOO(\HH)=\emptyset\) and therefore that \(\Ass\) is inconsistent.
Since the other option sets do not matter, we then directly return the simplest conjunctive generator \(\HH\) for which \(\OOO(\HH)=\emptyset\), which is \(\{\emptyset\}\). 

Here too the use of arrays instead of sets is not a problem, because it will only make us check some things multiple times.
It is even advisable, for the reasons explained in \cref{subsec:practicalConsiderations}.
\begin{algorithm}
  \caption{From assessment to conjunctive generator, with simplifications\label{alg:CreateDer}}
  \begin{algorithmic}[1]
  \Require{finite assessment \(\Ass\Subset\Q\times\Qe\)}
  \Ensure{$\D\Subset \Qe$ such that \(\OOO(\D)=\OOO(\D_{\Ass})\)}
  \Function{AssessmentToConjunctive}{$\Ass$}
    \Let{$\D$}{$\emptyset$}
      \ForAll{$(V,W) \in \Ass$}
        \ForAll{$w \in W$}
          \State {$H \gets \emptyset$}
          \Let{containsPositive}{$\text{false}$}
          \ForAll{$v\in V$}
            \If{$v-w > 0$}  \Comment{\cref{prop:simpDer2}}
              \Let{containsPositive}{$\text{true}$}
              \State{\textbf{break}}
            \EndIf
            \If{$v-w \not\leq 0$} \Comment{\cref{prop:simpDer1}}
              \Let{$H$}{ $H\cup\{v-w\}$}
            \EndIf
          \EndFor
          \If{\textbf{not} containsPositive}
            \If{$H=\emptyset$}
              \State{\textbf{Warning:} \(\Ass\) is not consistent.} \Comment{\cref{eq:inconsistentDerivedAssessment}}
              \State\Return{$\{\emptyset\}$}
            \EndIf
            \Let{ $\D$}{$\D\cup\{\textsc{Max}(H,\textsc{optionOrd})\}$}\Comment{\cref{prop:simpDerForReal}}
          \EndIf
        \EndFor
      \EndFor
      \State\Return{$\D$} 
  \EndFunction
  \end{algorithmic}
  \end{algorithm}


  \section{Simplifying disjunctive generators}\label{sec:simpGen}
  Even if we simplify \(\HH\) using the methods in \cref{sec:simp}, transforming \(\HH\) into \(\G(\HH)\) using \cref{alg:ConjoToDisjNaive} can still lead to an exponential explosion in the size of this generator.
  To address this issue, we will now proceed to develop a method to simplify a disjunctive generator \(\G\), and we will use it to simplify \(\G(\HH)\) without ever explicitly constructing \(\G(\HH)\) itself.
  If we do this without altering the corresponding set of preference orders~\(\OOOO(\G(\HH))\), then as before, this will reduce the running time of \cref{alg:CheckConsistency,alg:Choose} without altering the result.
We will achieve this by removing entire generator sets that do not contribute any new information, or individual uninformative options inside generator sets.
\subsection{Removing option sets from generators}
Our first simplifications involve removing option sets \(G\in \G\) that do not contribute any information, in the sense that \(\OOOO(\G\setminus\{G\})=\OOOO(\G)\).
The first such type of option sets are those that are inconsistent, or equivalently, due to \cref{lem:posiZero}, those for which \(0\in \NN(G)\).
\begin{lemma}\label{th:simp2}
  Consider a set of option sets $\mathcal{G}\subseteq 2^\V$ and an option set $G\in\G$.
  If \(0\in \NN(G)\), then \(\OOOO(\G \setminus\{G\})=\OOOO(\G)\).
  \end{lemma}
  \begin{proof}
    Follows straight from the definition \(\OOOO(\G)=\bigcup_{G\in \G} \OOOO[G]\) and the fact that, due to \cref{eq:equivOGDGNNG}, \(\OOOO[G]=\emptyset\) when \(0\in \NN(G)\). 
  \end{proof}
The next condition is related to pairs of generator sets and is therefore a bit more intensive to check.
\begin{lemma}\label{th:simp3}
Consider a set of option sets $\mathcal{G}\subseteq 2^{\V}$ and two distinct option sets $G_1,G_2\in\G$.
If \(G_2\subseteq \NN(G_1)\), then \(\OOOO(\G \setminus\{G_1\})=\OOOO(\G)\).
\end{lemma}
\begin{proof}
  Since \(G_2\subseteq \NN(G_1)\), we have \(\OOOO[G_1]\subseteq \OOOO[G_2]\) by \cref{lem:posiClosureOperatorEquivalence,cor:MostConservativePrec}.
  Because \(\OOOO(\G)=\bigcup_{G\in \G} \OOOO[G]\), it follows that \(\OOOO(\G)=\OOOO(\G \setminus\{G_1\})\).
\end{proof}

To show its relevance we will use an example with \(\V=\R^3\).
\begin{example}
Consider the generator \(\G=\{G_1,G_2\}\), where
  \(G_1=\{(2,2,-1),(2,-1,2),(-1,2,2)\}\) and \(G_2=\{(5,-1,-1),(-1,5,-1),(-1,-1,5)\}\).
It is fairly straightforward to verify that this generator cannot be simplified by \cref{th:simp2}.
However, we do have that \(G_1\subseteq \posi(G_2\cup\V_{>0})=\NN(G_2)\); for example 
\(
  (2,2,-1)=\frac12 (5,-1,-1)+ \frac12(-1,5,-1).
\)
It therefore follows from \cref{th:simp3} that \(\OOOO(\{G_1,G_2\})=\OOOO(\{G_1\})\).
\end{example}

To automate this process
we define the function \textsc{G-Ord} that returns \(\text{true}\) for a given pair of option sets \((G_1,G_2)\in 2^\V \times 2^\V\) whenever \(G_2\subseteq \NN(G_1)\) and false otherwise.
Since we know from \cref{lem:posiClosureOperatorEquivalence} that this is equivalent to \( \NN(G_2) \subseteq \NN(G_1)\),
\textsc{G-Ord} defines a (non-strict) preorder \(\preccurlyeq\) on \(2^\V\)---defined by \(G_1\preccurlyeq G_2 \Leftrightarrow \textsc{G-Ord}(G_1,G_2) \)---as it is clearly reflexive and transitive.
This allows us to use \cref{alg:Max} once more, this time to select a minimal set \(\textsc{Max}(\G,\textsc{G-Ord})\) of undominated elements from a generator~\(\G\), at least if \(\G\) is finite and if we are able to evaluate \textsc{G-Ord}.\footnote{It should be noted that if we work with arrays, then duplicates can be present in the generator. In that case, only one of these duplicates will be kept because \cref{alg:Max} will retain exactly one option set from every equivalence class defined by \(\preccurlyeq\).}
\begin{corollary}\label{cor:simpGen}
  For any finite set of option sets \(\G\Subset 2^\V\),  \[\OOOO(\G)=\OOOO(\textsc{Max}(\G,\textsc{G-Ord})).\]
\end{corollary}
\begin{proof}
This follows immediately by repeated application of \cref{th:simp3}.
\end{proof}

Evaluating \textsc{G-Ord}$(G_1,G_2)$ for arbitrary \(G_1,G_2\in 2^\V\) is tricky, but if \(G_1\) and \(G_2\) are finite, then it can be evaluated as the conjunction of \isF\((G_1,g)\) over all nonpositive options~\(g\in G_2\setminus \V_{>0}\).
\begin{lemma}\label{lem:G-OrdasIsFeasible}
  Consider two finite option sets \(G_1,G_2\in \Qe\).
  Then 
  \[\textsc{G-Ord}(G_1,G_2) \Leftrightarrow (\forall g\in G_2\setminus \V_{>0})\;\textsc{IsFeasible}(G_1,g).\]
\end{lemma}
\begin{proof}
  \(\textsc{G-Ord}(G_1,G_2)\) is by definition true if and only if \(G_2\subseteq \NN(G_1)\), or equivalently, if \(g\in \NN(G_1)\) for all \(g\in G_2\). 
  Due to \cref{th:testingg} and the definition of \isF, this is equivalent to the condition that for every \(g\in G_2\), we have that \(g>0\) or that \(\textsc{IsFeasible}(G_1,g)\) is true.
  This is equivalent to the condition on the right.
\end{proof}

If an option set \(G_1\) belongs to a generator \(\G(\HH)\), where \(\HH\) is created using \cref{alg:CreateDer}, then there is no need to check if \(g>0\) because such \(g\) will have been removed based on \cref{prop:simpDer2}.


\subsection{Simplifying generator sets}

In contrast to the previous section, we will now look at which options inside an individual option set of a generator do not contain any information.
The following lemma provides a sufficient condition for two generator sets---one of which can be a subset of the other---to generate the same set of preference orders.

\begin{lemma}\label{lem:equivNNenOO}
  For any two option sets \(G,G'\in 2^\V\), if \(\NN(G)=\NN(G')\), then also \(\OOOO[G]=\OOOO[G']\).
\end{lemma}
\begin{proof}
  This follows immediately from two applications of \cref{cor:MostConservativePrec}.
\end{proof}
  Inspired by this result, we will try to simplify a generator set by replacing it with a smallest subset~\(G'\subseteq G\) for which \(\NN(G)=\NN(G')\).
  That is, we want to come up with a function \textsc{MinConeSubset}\(\colon\Qe\to \Qe\) that, for every \(G\in \Qe\), returns some subset \(G'=\textsc{MinConeSubset}(G)\subseteq G\) for which \(\NN(G)=\NN(G')\neq \NN(G'\setminus\{u\})\) for all \(u\in G'\).
  Due to \cref{lem:equivNNenOO}, the resulting \(G'\) will then yield the same set of preferences.

\begin{proposition}\label{prop:MinConeSubset}
  Consider an option set \(G\in\Qe\).
  Then \[\OOOO[G]=\OOOO[\textsc{MinConeSubset}(G)].\]
\end{proposition}
\begin{proof}
This follows immediately from the properties that we require of \textsc{MinConeSubset}---i.e. \(\NN(G)=\NN(\textsc{MinConeSubset}(G))\)---and \cref{lem:equivNNenOO}.
\end{proof}

  A first way to obtain such a function is to use the following lemma to find such a \(G'\) iteratively.
\begin{lemma}\label{th:simp1}
For any option set~\(G\in 2^\V\) and option \(u\in G\), \(u\in   \NN(G\setminus\{u\})\) if and only if \(\NN(G)=\NN(G\setminus \{u\})\).
\end{lemma}
\begin{proof}
  The implication to the left is immediate from \(\NN(G)=\posi(G\cup\V_{>0})\), the definition of the \(\posi\) operator with \(n=1\) and \(u\in G\), so we prove the implication to the right.

  On the one hand, since \(\NN(G\setminus\{u\})\subseteq \NN(G\setminus\{u\})\), \cref{lem:posiClosureOperatorEquivalence} implies that \( G\setminus\{u\} \subseteq  \NN(G\setminus \{u\})\).
  This together with \(u \in \NN(G\setminus\{u\})\) implies that \(G\subseteq \NN(G\setminus\{u\})\).
  Therefore, by \cref{lem:posiClosureOperatorEquivalence}, we have that \(\NN(G)\subseteq \NN(G\setminus\{u\})\).

  On the other hand, since \(\NN(G)\subseteq \NN(G)\), \cref{lem:posiClosureOperatorEquivalence} implies that \(G\subseteq \NN(G)\).
  Therefore, \(G\setminus\{u\}\subseteq G\subseteq \NN(G)\) and hence, by \cref{lem:posiClosureOperatorEquivalence}, \(\NN(G\setminus\{u\})\subseteq \NN(G)\).
\end{proof}

This can again be applied to our running example.
\begin{myexp}\label{ex:1.8}
Let us start from the conjunctive generator $\D=\{\{h_2\},\allowbreak\{h_3\},\allowbreak\{h_5\}\}$
that we found 
in \cref{ex:1.7} with $h_2=(2,-1)$, $h_3=(7,-4)$ and $h_5=(-7,7)$.
The corresponding disjunctive generator is \(\G(\HH)=\{G\}\) with \(G=\{h_2,h_3,h_5\}\).
Since \(4 h_2=(8,-4)>(7,-4)=h_3\), we know that \(h_2=\frac14 h_3 + \frac14 p\) for \(p=(1,0)\in \V_{>0}\), which implies that \(h_2\in \NN(\{h_3,h_5\})\).
We can therefore remove $h_2$ from the set \(G\) in this generator due to \cref{th:simp1,lem:equivNNenOO}, resulting in \(G'=\{h_3,h_5\}\).
This leads to a generator \(\G'=\{G'\}\) that is equivalent to \(\G(\HH)\), in the sense that \(\OOOO(\G')=\OOOO(\G(\HH))=\OOO(\HH)\).
Since we already know from \cref{ex:1.7} that \(\OOO(\HH)=\OOO(\HH_{\Ass})=\OOOO(\G_\Ass)\), we find that instead of computing \(C^{\G_{\Ass}}=C_{\OOOO(\G_{\Ass})}\) to evaluate \(\NatExt\), as we did in \cref{ex:1.6}, we can now instead evaluate \(C^{\G'}=C_{\OOOO(\G')}\).
Since \(\G'\) is smaller, we know from \cref{alg:Choose} that this will be easier.
\hfill\pushQED{$\lozenge$}\popQED\end{myexp}

To obtain a smallest subset \(G'\) of \(G\) such that \(\NN(G')=\NN(G)\neq \NN(G'\setminus \{u\})\) for all \(u\in G'\), we can now keep on using \cref{th:simp1,lem:equivNNenOO} to remove options from \(G\) until we cannot remove any more.
This yields a method whose time complexity is polynomial in both the dimension of the state space and the number of options in \(G\).
To check the condition \(u\in \NN(G\setminus\{u\})\) in practice, due to \cref{th:testingg} we can first check if \(u>0\)---in which case it is always true, so \(u\) can be removed---and otherwise evaluate \isF\((G\setminus \{u\},u)\)---removing \(u\) if it is true.

There are other ways to come up with a function \textsc{MinConeSubset} that has the required properties though, because it is related to a well-known problem in polyhedral theory: redundancy removal.
To make the connection, we first need to introduce some definitions.
For any set of options \(V\subseteq \V\) the \emph{conic hull} of \(V\) is defined as
\[
  \cone(V)=\cset*{\sum_{j=1}^n \lambda_j v_j}{ n\in \N,v_j\in V, \lambda_j\geq 0},
\]
and turns the set \(V\) into a (pointed) \emph{cone}, which includes \(0\).
This corresponds to \citep[Eq.~(2.6)]{fukuda2020polyhedral} for finite \(V\).
It is easy to check that, similarly to the \(\NN\) and \(\posi\) operators, the \(\cone\) operator is a closure operator as well.
An option~\(v\in V\) is called \emph{redundant} if \(\cone(V\setminus \{v\})=\cone(V)\).
Removing redundant options is therefore clearly a problem that is closely related to what is done in \cref{th:simp1}.
If \(V\in \Qe\) is a finite set of options, then \emph{redundancy removal} means finding a subset \(V^*\) of options so that \(V^*\) does not contain any redundant options and \(\cone(V)=\cone(V^*)\), as described for example by \citet[Problem~7.2 and Section~7.4]{fukuda2020polyhedral}. 

Let us now make the connection between redundancy removal and \textsc{MinConeSubset} more formal.
It is clear that \(\posi(V)\cup \{0\}= \cone(V)\) for any non-empty \(V\in 2^\V\setminus\{\emptyset\}\).
Therefore, we see that \(\NN(G)\cup\{0\}=\posi(G\cup\V_{>0})\cup\{0\}=\cone(G\cup\V_{>0})\) is the conic hull of \(G\cup\V_{>0}\).
Hence, if \(0\notin \NN(G)\)---if it contained \(0\) then this \(G\) could be removed by \cref{th:simp2}---then \(\NN(G)=\cone(G\cup\V_{>0})\setminus\{0\}\).
The following lemma uses this connection between the \(\cone\) and the \(\posi\) operator to establish a connection between redundancy removal and \textsc{MinConeSubset}.
\begin{lemma}\label{lem:MinConeSubsetConnection}
  For any two option sets \(G,G^*\in 2^\V\), if \(0\notin \NN(G)\), \(0\notin G^*\) and \(\cone(G\cup \V_{>0})=\cone(G^*)\), then
  \begin{enumerate}[label=(\roman*),ref=(\roman*)]
  \item\label{stat:1Mincon} \(\NN(G)=\NN(G^*\setminus \V_{>0})\);
  \item\label{stat:2Mincon} for all \(u\in G^*\setminus\V_{>0}\) such that \(\cone(G^*)\neq \cone(G^*\setminus \{u\})\), we also have that \(\NN(G^*\setminus \V_{>0})\neq \NN((G^*\setminus \V_{>0})\setminus \{u\})\).
  \end{enumerate}
\end{lemma}
We first prove the following intermediate result.
\begin{lemma}\label{lem:tussenstap}
For any two option sets \(G,G^*\in 2^\V\), if \(\NN(G)=\posi(G^*)\), then \(\NN(G)=\NN(G^*\setminus \V_{>0})\). 
\end{lemma}
\begin{proof}
Since \(\NN(G^*\setminus \V_{>0})=\posi((G^*\setminus \V_{>0})\cup \V_{>0})=\posi(G^*\cup \V_{>0})=\NN(G^*)\), it suffices to prove that \(\NN(G)=\NN(G^*)\).

First, by \cref{ax:CLincreasing} for the \(\posi\) operator, \(\NN(G)=\posi(G^*)\subseteq \posi(G^*\cup \V_{>0})=\NN(G^*)\).

So it remains to prove that \(\NN(G^*)\subseteq \NN(G)\).
By \cref{lem:posiClosureOperatorEquivalence}, we know that this is equivalent to proving that \(G^* \subseteq \NN(G)\).
But we have by \cref{ax:CLext} for the \(\posi\) operator that \(G^*\subseteq \posi(G^*)= \NN(G)\).
\end{proof}
\begin{proofof}{\cref{lem:MinConeSubsetConnection}}
  For the first part it suffices by \cref{lem:tussenstap} to prove that \(\NN(G)=\posi(G^*)\).
  Since \(0\notin \NN(G)\), we know from the main text preceding \cref{lem:MinConeSubsetConnection} that \(\NN(G)=\cone(G\cup \V_{>0})\setminus\{0\}=\cone(G^*)\setminus\{0\}\).
  We will now prove that also \(\posi(G^*)=\cone(G^*)\setminus\{0\}\).
  From the definitions of the \(\posi\) and \(\cone\) operators it can be seen that it is sufficient to prove that \(0\notin \posi(G^*)\).
  Assume \emph{ex absurdo} that \(0\in \posi(G^*)\).
  Then there are \(n\in \N\), \(v_j\in G^*\), and \(\lambda_j>0\) such that \(0=\sum_{j=1}^n \lambda_j v_j\).
  It also follows from \cref{ax:CLext} for the \(\cone\) operator and the assumptions in the statement that \(G^*\subseteq \cone(G^*)=\cone(G\cup \V_{>0})=\posi(G\cup \V_{>0})\cup \{0\}\), which, since \(0\notin G^*\), implies that \(G^*\subseteq \posi(G\cup \V_{>0})\).
  So for every \(j\in \{1,...,n\}\) there are \(m_{j}\in \N\), \(w_{j,i}\in G\cup \V_{>0}\), and \(\mu_{j,i}>0\) such that \(v_j=\sum_{i=1}^{m_j} \mu_{j,i} w_{j,i}\).
  It therefore follows that 
  \(
  0=\sum_{j=1}^n \lambda_j v_j=\sum_{j=1}^n \sum_{i=1}^{m_j} (\lambda_j \mu_{j,i}) w_{j,i}\in \posi(G\cup \V_{>0})=\NN(G)\),
   which is a contradiction.
  Hence, \(0\notin \posi(G^*)\) and therefore \(\posi(G^*)=\cone(G^*)\setminus\{0\}=\NN(G)\).
  It now follows from \cref{lem:tussenstap} that \(\NN(G)=\NN(G^*\setminus\V_{>0})\).

  
  We prove the second part by contraposition.
  Assume that \(\NN(G^*\setminus \V_{>0})=\NN((G^*\setminus \V_{>0})\setminus \{u\})\) for some \(u\in G^*\setminus\V_{>0}\).
  We'll show that then also \(\cone(G^*)=\cone(G^*\setminus\{u\})\).
  The set inclusion \(\cone(G^*\setminus\{u\})\subseteq \cone(G^*)\) follows from \cref{ax:CLincreasing} for the \(\cone\) operator.
  We will now prove the other inclusion \(\cone(G^*)\subseteq \cone(G^*\setminus\{u\})\) by first proving some intermediate things related to \(u\).

  Since \(u\in G^*\setminus \V_{>0}\), it is clear from the definition of the \(\posi\) operator that 
  \begin{align*}
  u\in \posi((G^*\setminus\V_{>0})\cup \V_{>0})&=\NN(G^*\setminus \V_{>0})=\NN((G^*\setminus \V_{>0})\setminus \{u\})\\&=\posi(((G^*\setminus \V_{>0})\setminus\{u\})\cup \V_{>0})=\posi((G^*\setminus \{u\})\cup \V_{>0}).
  \end{align*}
  Therefore, there are \(n\in \N\), \(v_j\in (G^*\setminus \{u\})\cup \V_{>0}\) and \(\lambda_j>0\) such that \(u=\sum_{j=1}^n \lambda_j v_j\).
  Without loss of generality, there is some \(k\in \{0,...,n\}\) such that \(v_j\in G^*\setminus \{u\}\) for all \(j\leq k\) and \(v_j\in \V_{>0}\) for all \(j>k\).
  Then we have that \(u=\sum_{j=1}^k \lambda_j v_j + \sum_{j=k+1}^n \lambda_j v_j\).
  We have that \(p\coloneqq\sum_{j=k+1}^n \lambda_j v_j\) is either equal to \(0\), when \(k=n\), or belongs to \(\V_{>0}\) by the scaling and translation property of~\(<\).

  If \(p\in\V_{>0}\), then \(p\) clearly belongs to \(\posi(G\cup \V_{>0})=\NN(G)\), by \cref{ax:CLext} for the \(\posi\) operator, and therefore also belongs to \(\posi(G^*)\) since we already proved in the first part that \(\posi(G^*)=\NN(G)\).
  Since \(p\in \posi(G^*)\) and because we can rearrange and lump terms together, there are then \(\ell\in \N\) with \(\ell\geq 2\), \(\nu_1\geq 0\), \(y_r\in G^*\setminus\{u\}\) and \(\nu_r > 0\) for all \(r\geq 2\), such that \(p=\nu_1 u + \sum_{r=2}^\ell \nu_r y_r\), where \(\ell\) is at least \(2\) since \(u\notin \V_{>0}\).
  %
  If \(p=0\), we simply let \(p=\nu_1 u\) with \(\nu_1=0\). 
  We always have that \(\nu_1 < 1\), trivially for \(p=0\) and, for \(p\in\V_{>0}\), because otherwise \(0\in \posi(G^*)=\NN(G)\) because \(0=\sum_{j=1}^k \lambda_j v_j+(\nu_1-1) u + \sum_{r=2}^\ell \nu_r y_r\), which is not true by assumption.

  Therefore, in any case, there are \(\ell\in \N\), \(\nu_1\geq 0\) with \(\nu_1<1\), \(y_r\in G^*\setminus\{u\}\) and \(\nu_r > 0\) for all \(r\geq 2\), such that \(p=\nu_1 u + \sum_{r=2}^\ell \nu_r y_r\)---if \(\ell=1\), the empty sum in this expression is \(0\).
  So, we have that \(u=\sum_{j=1}^k \frac{\lambda_j}{1-\nu_1} v_j + \sum_{r=2}^\ell \frac{\nu_r}{1-\nu_1} y_r\) with \(v_j,y_r\in G^*\setminus\{u\}\) for all \(j\in \{1,...,k\}\) and \(r\in \{2,...,\ell\}\).
  This implies that \(u\in \cone(G^*\setminus\{u\})\) by definition.
  Since also \(G^*\setminus\{u\}\subseteq \cone(G^*\setminus\{u\})\) by \cref{ax:CLext}, we have that \(G^*\subseteq \cone(G^*\setminus\{u\})\), whence \(\cone(G^*)\subseteq \cone(\cone(G^*\setminus \{u\})) = \cone(G^*\setminus \{u\})\) by \cref{ax:CLincreasing,ax:CLundo}.
\end{proofof}

Now if \(G\cup \V_{>0}\) were finite, then this result would allow us to implement \textsc{MinConeSubset} using redundancy removal, applying it to reduce \(G\cup\V_{>0}\) to the set \(G^*\) that appears in \cref{lem:MinConeSubsetConnection}.

However, \(G\cup \V_{>0}\) is always infinite because \(\V_{>0}\) is.
We resolve this by replacing \(\V_{>0}\) with the set \(\{\mathbb{I}_x \colon x \in \X\}\), where, for every \(x\in \X\), the standard basis vector \(\mathbb{I}_x\colon \X \to \R\) maps \(x\) to 1 and all other states in \(\X\) to 0.
We formalise this in the following lemma, the proof of which we omit since its steps are straightforward and based on the fact that \(\V_{>0}\subseteq \cone(\{\mathbb{I}_x\colon x\in \X\})\).
\begin{lemma}\label{lem:extreme} 
  Let the state space \(\X\) be finite.
  For any option set \(G\in 2^\V\), we have that \(\cone(G\cup \V_{>0})=\cone(G\cup \{\mathbb{I}_x\colon x\in \X\})\).
\end{lemma}
For any \(G\in \Qe\) such that \(0\notin \NN(G)\), we can use this result to implement \(G'=\textsc{MinConeSubset}(G)\) as follows: we first apply redundancy removal to find a minimal subset \(G^*\) of \(G\cup \{\mathbb{I}_x\colon x\in \X\}\) for which \(\cone(G^*)=\cone(G\cup \{\mathbb{I}_x\colon x\in \X\})\) and then remove the positive options from \(G^*\) to obtain \(G'=G^*\setminus \V_{>0}\).
This is always possible because \(G\cup \{\mathbb{I}_x\colon x\in \X\}\) is finite.
To see why it works, we first observe that, due to \cref{lem:extreme}, \(\cone(G^*)=\cone(G\cup \V_{>0})\).
Furthermore, since \(0\notin \NN(G)\), we automatically have that \(0\notin G^*\) because \(G^*\subseteq G\cup \{\mathbb{I}_x\colon x\in \X\}\) and \(0\notin G\) since \(0\notin\NN(G)\).
Using \cref{lem:MinConeSubsetConnection} \ref{stat:1Mincon}, this already implies that \(\NN(G)=\NN(G')\).
Furthermore, since the minimality of \(G^*\) implies that \(\cone(G^*)\neq \cone(G^*\setminus \{u\})\) for all \(u\in G^*\), we know from \cref{lem:MinConeSubsetConnection} \ref{stat:2Mincon} that \(\NN(G')\neq \NN(G'\setminus \{u\})\) for all \(u\in G'\).
So we see that \(G'\) indeed satisfies the properties required of \textsc{MinConeSubset}\((G)\).

In practice, to find a minimal subset \(G^*\) of \(G\cup \{\mathbb{I}_x\colon x\in \X\}\) such that \(\cone(G^*)=\cone(G\cup \{\mathbb{I}_x\colon x\in \X\})\), we can for example use the function \texttt{removevredundancy} of the Julia library Polyhedra.jl \cite{legat2023polyhedral}.
Alternatively, such a minimal subset free of redundant options can also be found by calculating the dual representation, for example using the Quickhull \cite{barber1996quickhull} or the Double Description \cite{DoubleDes} methods, but the time complexity of these algorithms scale exponentially with the size of the state space; such methods are therefore usually only used for 2- and 3-dimensional vector spaces. 

To summarise, it is possible to implement \textsc{MinConeSubset} directly using \textsc{IsFeasible}, but it is also possible to use redundancy removal methods from polyhedral theory.
In our experiments in \cref{sec:experiment}, we will implement \textsc{MinConeSubset} using \texttt{removevredundancy} from Polyhedra.jl, but everything works regardless of which approach is used.
\subsection{Simplifying a generator}\label{subsec:simpGenNaive}
Having presented several techniques to simplify a disjunctive generator~\(\G\), \cref{alg:updateCGNaive} now combines them into a single algorithm.
For every \(G\in \G\), we first check if \(0\in \NN(G)\) because if this is the case, we know from \cref{th:simp2} that we can remove it.
As we know from \cref{th:testingg}, this happens if \isF$(G,0)$ is true.
If this is not the case, we simplify \(G\) by replacing it with \(\textsc{MinConeSubset}(G)\), as allowed by \cref{prop:MinConeSubset}.
Finally, we take the maximal elements of the generator with respect to \textsc{G-Ord}, which is allowed by \cref{cor:simpGen}.

\begin{algorithm}
  \caption{Simplify a disjunctive generator \label{alg:updateCGNaive}}
  \begin{algorithmic}[1]
  \Require{finite disjunctive generator $\G\Subset \Qe$.}
  \Ensure{finite disjunctive generator $\G'$ for which \(\OOOO(\G')=\OOOO(\G)\)}
  \Function{Simplify}{$\G$}
      \Let{$\G'$}{$\emptyset$}
      \ForAll{$G\in \G$}
          \If{\textbf{not} \isF\((G,0)\)} \Comment{\cref{th:simp2}}
            \Let{\(\G'\)\!}{\!$\G'\! \cup \{\text{\textsc{MinConeSubset}}(G)\}$} \Comment{\cref{prop:MinConeSubset}}
          \EndIf
      \EndFor
      \Let{$\G'$}{$\textsc{Max}(\G',\textsc{G-Ord})$}\Comment{\cref{cor:simpGen}}

    \State\Return{$\G$}
  \EndFunction
  \end{algorithmic}
  \end{algorithm}

\subsection{From conjunctive generator to disjunctive generator: step by step}\label{subsec:fromDerToGen}
\cref{alg:updateCGNaive} can be applied to any disjunctive generator, but in particular we will often be interested in applying it to a generator of the form \(\G(\HH)\).
The naive way to implement this would be to first create the generator \(\G(\D)\) and then use \cref{alg:updateCGNaive} to simplify it.
  
This is however not efficient.
Since the size of \(\G(\D)\) is exponential in the size of \(\D\), the mere act of saving it in memory would already use a lot of resources (both time and memory).
Fortunately though, there is no need to construct \(\G(\D)\) if all we want is to find a small \(\G\) such that \(\OOOO(\G)=\OOOO(\G(\D))\).
Instead, we can construct such a simplified generator \(\G\) iteratively from \(\D\).
Let us enumerate the option sets in \(\D\) as \(H_1,...,H_m\) and, for each \(k=0,...,m\), let \(\D_{1:k}=\{H_1,...,H_k\}\).
The idea is now to construct a generator \(\G_k\) such that \(\OOOO(\G_k)=\OOOO(\G(\D_{1:k}))\), for increasing values of \(k=1,...,m\).
For \(k=m\), since \(\D_{1:m}=\D\), we will then find the set \(\G=\G_m\) that we are after.

We start from \(\G_0=\{\emptyset\}\) because \(\OOOO(\{\emptyset\})=\OOOO(\G(\emptyset))=\OOO(\emptyset)=\OOO\) cf.  \cref{foot:4}.
Next, for each \(k\in\{1,...,m\}\), we construct \(\G_k\) from \(\G_{k-1}\) and \(H_k\), in such a way that \(\OOOO(\G_k)=\OOOO(\G(\D_{1:k}))\).
The following result shows that one way to achieve this is by letting \(\G_k \coloneqq \{G\cup\{h\}\colon G\in \G_{k-1}, h \in H_k\}\).
\begin{proposition}\label{th:crossprodOrderings}
  Let \(\D\subseteq \Qe\) be a conjunctive generator, \(\G\) a generator such that \(\OOOO(\G)=\OOOO(\G(\D))\) and consider any option set \(H\in \Qe\).
  Then 
  \[
    \OOOO(\G(\D\cup \{H\}))=\OOOO(\{G\cup\{h\}\colon G\in \G, h \in H\}).
  \]
\end{proposition}
\begin{proof}
  For every \(G \in \G\) and \(h \in H\) we have that \(\OOOO[G\cup \{h\}]=\OOOO[G]\cap \OOO_h\).
  Therefore, we have that
  \begin{align*}
    \OOOO(\{G\cup\{h\}\colon G\in \G, h \in H\}) &\overset{(\ref{eq:OOOOGdefRondehakenSter})}= \bigcup_{G\in \G} \bigcup_{h\in H} \OOOO[G\cup \{h\}]= \bigcup_{G\in \G} \bigcup_{h\in H} (\OOOO[G]\cap \OOO_h) \\
    &= \bigcup_{G\in \G} \left(\OOOO[G]\cap \left(\bigcup_{h\in H} \OOO_h\right)\right)
    =\left(\bigcup_{G\in \G}  \OOOO[G]\right)\cap \left(\bigcup_{h\in H} \OOO_h\right)\\
    &\overset{(\ref{eq:OOOOGdefRondehakenSter}),(\ref{eq:OOOHdefVierkanteHakenSterSter})}=\OOOO(\G) \cap \OOO[H]= \OOOO(\G(\D))\cap \OOO[H]
     \overset{(\ref{eq:OOODisUnieOverGD}),(\ref{eq:OOOOGdefRondehakenSter})}= \OOO(\D)\cap \OOO[H] 
    \\&\overset{(\ref{eq:OOODunieOverOd})}= \bigcap_{H^*\in \D\cup\{H\}} \OOO[H^*]
    \overset{(\ref{eq:OOODunieOverOd})}= \OOO(\D\cup \{H\}) \overset{(\ref{eq:OOODisUnieOverGD}),(\ref{eq:OOOOGdefRondehakenSter})}= \OOOO(\G(\D\cup \{H\})).\qedhere
  \end{align*}
\end{proof}
However, this way of defining \(\G_k\) would simply amount to using \cref{alg:ConjoToDisjNaive}, without any simplifications, yielding \(\G(\HH)\).
Fortunately, however, this iterative approach allows us to apply our simplifications to every intermediate \(\G_k\).
That is, for every \(k\), we can let  \(\G_{k,\text{unsimp}}\coloneqq\{G\cup\{h\}\colon G\in \G_{k-1}, h \in H_k\}\) and then simplify it to obtain \(\G_k=\textsc{Simplify}(\G_{k,\text{unsimp}})\). 
Due to \cref{th:crossprodOrderings,alg:updateCGNaive}, we then still have that \(\OOOO(\G_k)=\OOOO(\G_{k,\text{unsimp}})=\OOOO(\G(\HH_{1:k}))\).
This would however require us  to loop twice over all elements of \(\G_{k,\text{unsimp}}\): once to construct it and once to simplify it.
We can avoid this by constructing \(\G_k\) directly from \(\G_{k-1}\) and \(H_k\) and doing the simplifications along the way:
for every \(G\cup \{h\}\) that we construct, we should not include it in \(\G_{k,\text{unsimp}}\) if we will  remove it anyway in the next step using \cref{th:simp2}, and if we include it, we might as well already replace it with \textsc{MinConeSubset}$(G\cup \{h\})$ if we are going to replace it by that anyway in the next loop in the simplification of \(\G_{k,\text{unsimp}}\).
We have implemented this in \cref{alg:updateCG}.

  \begin{algorithm}
    \caption{From conjunctive to disjunctive generator, with simplifications \label{alg:updateCG}}
    \begin{algorithmic}[1]
    \Require{finite conjunctive generator $\D\Subset \Qe$}
    \Ensure{finite disjunctive generator $\G$ for which \(\OOOO(\G)=\OOOO(\G(\D))\)}
    \Function{ConjunctiveToDisjunctive}{$\D$}
        \Let{$\G$}{$\{\emptyset\}$}
        \ForAll{$H\in \D$}\Comment{\cref{th:crossprodOrderings}}
          \Let{$\G^*$}{$\emptyset$}
          \ForAll{$G\in \G$}
            \ForAll{$h \in H$}
              \If{\textbf{not} \isF\((G\cup \{h\},0)\)} \Comment{\cref{th:simp2}}
                \Let{\(\G^*\)\!}{\!$\G^*\! \cup \{\text{\textsc{MinConeSubset}}(\{G\cup \{h\}\})\}$} \State{}\Comment{\cref{prop:MinConeSubset}}
              \EndIf
            \EndFor
          \EndFor
          \Let{$\G$}{$\textsc{Max}(\G^*,\textsc{G-Ord})$}\Comment{\cref{cor:simpGen}}
        \EndFor
      \State\Return{$\G$}
    \EndFunction
    \end{algorithmic}
    \end{algorithm}

\section{Experiments}\label{sec:experiment}
Having introduced a number of different algorithms, we now proceed to evaluate their performance.
In particular, we will investigate when our algorithms can do the following efficiently: to start from a finite assessment \(\Ass\subseteq \Q\times \Qe\), construct a generator \(\G\) such that \(\OOOO({\G})=\OOOO(\G_{\Ass})\) and use this generator to evaluate the corresponding choice function \(C^{\G}=C_{\Ass}\) for a given option set \(A\in \Q\) containing options from which we want to choose.
First, since we have provided multiple methods to process and simplify the information contained in an assessment, we will investigate to see whether these simplifications are actually useful or whether it is perhaps sometimes faster to skip these preprocessing steps and evaluate \(C_{\Ass}\) directly.
Second, since a lot of variables determine the speed of the algorithms---for both processing the assessments and making the decisions---we want to test how the performance changes based on the type of assessment.
The variables that we will focus on are the size of the assessment and the amount of imprecision in the assessment.

Since an assessment~\(\Ass\) that is not consistent gives rise to an uninteresting operator \(C_{\Ass}\)---in that case, as we know from \cref{prop:niet-leeg}, \(C_{\Ass}(A)=\emptyset\) for all \(A\in \Q\)---we will focus on the case where \(\Ass\) is consistent. 

In our first two experiments, we will investigate how fast the number of option sets in the generator \(\G\) grows as the size of the assessment increases, for different types of assessments.
The first experiment focusses on the type of decision rule that is used to generate the assessments, whereas the second focusses on the amount of imprecision.
The reason this is of interest is that for large \(\G\), evaluating the natural extension~\(C^{\G}\) with \cref{alg:Choose} will require many evaluations of \textsc{IsFeasible}, each of which requires solving a linear program.
Finally, in our third set of experiments, we will measure the actual time it takes to preprocess and then evaluate a choice function.

\subsection{Set-up of the experiments}\label{sec2:set-up}
To test the algorithms, we will start from consistent assessments.
To get such consistent assessments, we will consider a coherent choice function \(C\) and a sequence of random option sets \(A_1,...,A_L\in \Q\), and use these to define the assessment as
\[
    \Ass(C,A_1,...,A_L)\coloneqq \{(C(A_\ell),A_\ell\setminus C(A_\ell))\colon \ell\in \{1,...,L\}\}.
\]
We will also use the shorthand notation \(\Ass\coloneqq\Ass(C,A_1,...,A_L)\) if the choice function and option sets are clear from the context.
To study the effect of the size of the assessment, we will often increase the size of the assessments one by one.
To that end, for all \(\ell\in \{1,...,L\}\), we let 
\[
    \Ass_{1:\ell}\coloneqq \Ass(C,A_1,...,A_\ell)=\{(C(A_k),A_k\setminus C(A_k))\colon k\in \{1,...,\ell\}\},
\]
with then \(\Ass_{1:L}=\Ass\).
Since our assessments come in the form of partial information about a coherent choice function, they can trivially be extended to a coherent choice function and are therefore indeed consistent.
The natural extension of such an assessment will then be a conservative approximation of the choice function \(C\) that is used to construct the assessment, where the approximation becomes less conservative---since it is based on more information---as \(\ell\) increases.

The coherent choice functions that we will use to construct our assessments are instances of  \(C_{\mathcal{E}}\), as introduced in \cref{sec2:examples}, where \(\mathcal{E}\) consists of a finite number of lower expectations~\(\Eu\) each of which is characterised by a finite number of extreme probability mass functions.
If \(p_1,...,p_n\) are the extreme probability mass functions of one such \(\Eu\), then as we explained in \cref{sec2:examples}, we can easily compute \(\Eu(u) = \min_{k\in\{1,...,n\}} \sum_{x\in\X} p_k(x) u(x)\) for every \(u\in \V\).
This implies that 
\[
  C_{\mathcal{E}}(A)=\left\{u \in A \colon (\exists {\Eu}\in \mathcal{E})(\forall v \in A)  {\Eu}(v-u) \leq 0 \text{ and } u\not< v \right\}\]
can be efficiently evaluated as well, for every \(A\in \Q\), enabling us to create the aforementioned consistent assessments.

We will do this for four different types of \(\mathcal{E}\)'s:
\begin{enumerate}
  \item the linear case, where \(\mathcal{E}=\{\E\}\) contains one single linear expectation; we will also use the notation  \(C_{\text{lin}}=C_{\{\E\}}\) and associate it with the colour \textcolor{LimeGreen}{lime $\blacksquare$};
  \item maximality, where \(\mathcal{E}=\{\Eu\}\) contains one single lower expectation; we will also use the notation  \(C_{\text{max}}=C_{\{\Eu\}}\) and associate it with the colour \textcolor{magenta}{magenta $\blacksquare$};
  \item E-admissibility, where \(\mathcal{E}=\{\E_1,\E_2,\E_3\}\) contains three linear expectations; we will also use the notation  \(C_{\text{adm}}=C_{\{\E_1,\E_2,\E_3\}}\) and associate it with the colour \textcolor{blue}{blue $\blacksquare$};
  \item another viable imprecise coherent choice function, where \(\mathcal{E}=\{\Eu_1,\Eu_2,\Eu_3\}\) contains three lower expectation; we will also use the notation  \(C_{\text{imp}}=C_{\{\Eu_1,\Eu_2,\Eu_3\}}\) and associate it with the colour \textcolor{indigo}{indigo $\blacksquare$}.
\end{enumerate}
To generate the (lower) expectations that make up these \(\mathcal{E}\)'s, we use Algorithm~2 in~\cite{nakharutai2018improved}, which works by randomly generating a finite number of probability mass function by means of Algorithm 1 in~\cite{nakharutai2018improved}.
The parameters to be set for this algorithm are the state space (4-dimensional in our case) and the number of probability mass functions for each lower expectation (we use 4).
We also take \(L=30\) and generate option sets \(A_1,...,A_{L}\) that contain a number of options between 2 and 8. 
The number of options in these option sets is drawn uniformly random from \(\{2,3,...,8\}\).
The options themselves are drawn uniformly random from the unit cube \([0,1]^4\).
Note that this choice of domain is not restrictive, as any option set can be rescaled and shifted by the same vector to an option set with options inside the unit cube while still carrying the same information. 
This is because \cref{ax:Oscal,ax:Otransla} gives us that \(w\prec v\) if and only if \(\lambda w+ u\prec \lambda w +u\) for any \(\prec \in \OOO\), \(u,v,w\in \V\) and \(\lambda>0\).
So if we use the notation \(\lambda A +u  \coloneqq \{\lambda a+u\colon a\in A\}\) for any \(A\in \Qe\), \(u\in \V\) and \(\lambda>0\), then we have for a given assessment \(\Ass\) and rescaled assessment \(\Ass'=\{(\lambda V + u, \lambda W + u)\colon (V,W)\in \Ass\}\) that, for any \(\prec\in \OOO\),
\begin{align*}
 (\forall (V,W)\in \Ass)(\forall w\in W)(\exists v\in V)\; w\prec v \;
 &\Leftrightarrow (\forall (V,W)\in \Ass)(\forall w\in W)(\exists v\in V)\; \lambda w+u\prec \lambda v+u\\
 &\Leftrightarrow (\forall (V',W')\in \Ass')(\forall w'\in W')(\exists v'\in V') \; w'\prec v'.
\end{align*}
Therefore, by \cref{prop:asss}, \(\OOO(\Ass)=\OOO(\Ass')\), implying that the consistency and natural extension of \(\Ass\) is the same as that of \(\Ass'\).

For each of the consistent assessments \(\Ass_{1:\ell}\) that are constructed in this way, we will evaluate its natural extension with \cref{alg:Choose}, using 3 different methods to obtain a disjunctive generator \(\G\):
\begin{enumerate}
  \item the straightforward way of constructing the conjunctive generator \(\HH_\times\) from \(\Ass_{1:\ell}\) using \cref{alg:AssToConjAkaPerineumNaive} and then constructing \(\G_\times \coloneqq \G(\HH_\times)\) whenever necessary using \cref{alg:ConjoToDisjNaive}, without saving it in memory; the results are depicted by \myTikzMark{x} in our plots further on,
  \item the intermediate method of constructing a conjunctive generator \(\D_{\triangle}\)  from \(\Ass_{1:\ell}\) using \cref{alg:CreateDer} and then constructing \(\G_\triangle \coloneqq \G(\HH_\triangle)\) whenever necessary using \cref{alg:ConjoToDisjNaive}, without saving it in memory; depicted by \myTikzMark{triangle},
  \item the full simplification method of constructing a conjunctive generator \(\D_{\square}=\D_{\triangle}\) from \(\Ass_{1:\ell}\) using \cref{alg:CreateDer} and then constructing \(\G_{\square}\) using \cref{alg:updateCG} and saving it in memory; depicted by \myTikzMark{square}.
\end{enumerate}
For the first two methods, we take the assessment \(\Ass_{1:\ell}\) and construct the corresponding conjunctive generator \(\D_{\times}\) and \(\D_{\triangle}\).
The size of the corresponding disjunctive generators, i.e. the number of option sets in these generators, can then be calculated using the formula 
\(
  |\G| =  \prod_{H\in \D} |H|,
\)
at least if---as we do---we do not check for duplicates and store these sets as arrays.
To evaluate the choice functions \(C^{\G_{\times}}\) and \(C^{\G_{\triangle}}\) using \cref{alg:Choose}, we loop over the respective generators \(\G_{\times}\) and \(\G_{\triangle}\) without ever fully saving them in memory.
As soon as an option set in the generator has been checked in \cref{alg:Choose}, we can forget about it.  
For the third method, we explicitly construct the disjunctive generator \(\G_{\square}\) and save it in memory, before running \cref{alg:Choose}, because we need the full generator anyway in \cref{alg:updateCG}.

As for the software we used: as programming language we used Julia, and for the linear programs we used the CPLEX solver.

\subsection{Varying the `real' model and size of the assessment}\label{subsec:exp1}
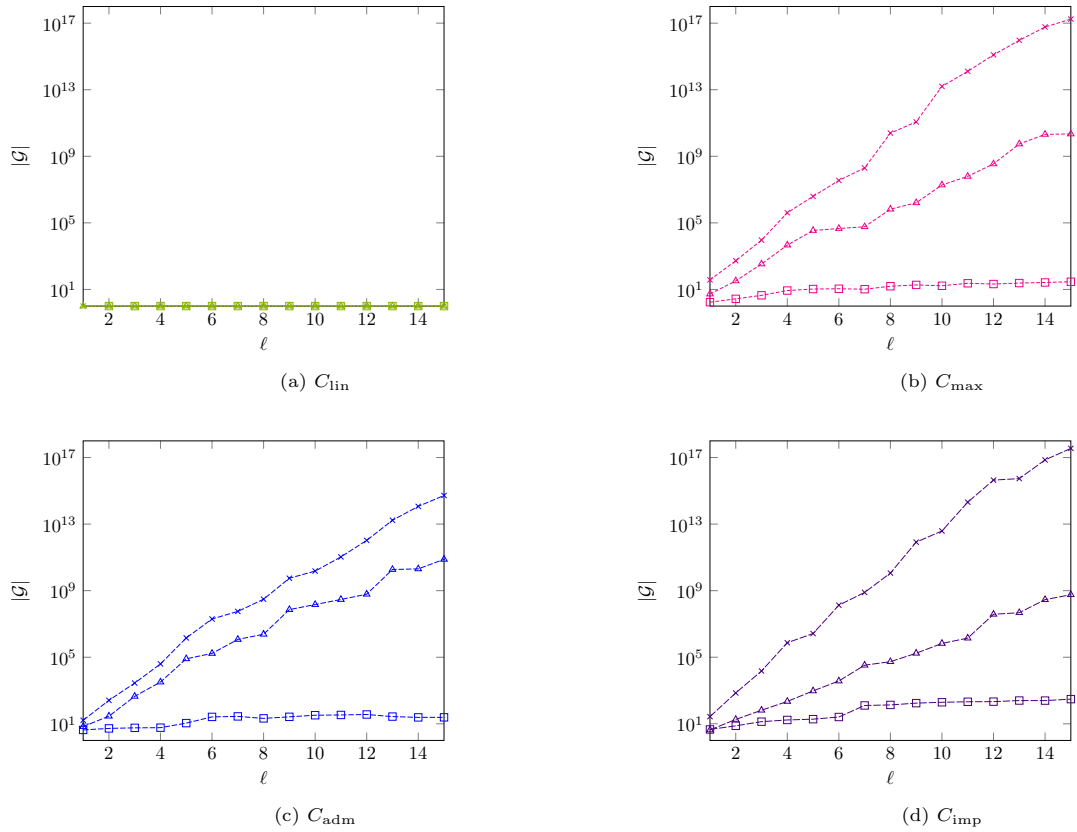
\begin{figure}
  \vspace{1cm}
  \begin{subfigure}[t]{0.5\textwidth}
     \centering
\begin{tikzpicture}[scale=0.8]
  \begin{semilogyaxis}
    [ ymin=1,ymax=10^18,
      xmin=1, xmax=15, domain=1:15,
      xlabel=$\ell$,ylabel=$|\G|$,
    ]
    \addplot[mark=x,mark options={LimeGreen,solid},dash pattern=on 1pt off 1pt,LimeGreen] table[x=l,y=g2,col sep=comma] {_1_-D4.csv};
    \addplot[mark=triangle,mark options={LimeGreen,solid},dash pattern=on 1pt off 1pt,LimeGreen] table[x=l,y=g3,col sep=comma] {_1_-D4.csv};
    \addplot[mark=square,mark options={LimeGreen,solid},dash pattern=on 1pt off 1pt,LimeGreen] table[x=l,y=g1,col sep=comma] {_1_-D4-s.csv};
  \end{semilogyaxis}
\end{tikzpicture}
\caption{\(C_{\text{lin}}\)}
\vspace*{20mm}
\end{subfigure}
  \begin{subfigure}[t]{0.5\textwidth}
     \centering
\begin{tikzpicture}[scale=0.8]
  \begin{semilogyaxis}
    [ ymin=1,ymax=10^18,
      xmin=1, xmax=15, domain=1:15,
      xlabel=$\ell$,ylabel=$|\G|$,
    ]
    \addplot[mark=x,mark options={magenta,solid},dash pattern=on 2pt off 1pt,magenta] table[x=l,y=g2,col sep=comma] {_4_-D4.csv};
    \addplot[mark=triangle,mark options={magenta,solid},dash pattern=on 2pt off 1pt,magenta] table[x=l,y=g3,col sep=comma] {_4_-D4.csv};
    \addplot[mark=square,mark options={magenta,solid},dash pattern=on 2pt off 1pt,magenta] table[x=l,y=g1,col sep=comma] {_4_-D4-s.csv};
  \end{semilogyaxis}
\end{tikzpicture}
\caption{\(C_{\text{max}}\)}\vspace*{20mm}
\end{subfigure}
\begin{subfigure}[t]{0.5\textwidth}
  \centering
  \begin{tikzpicture}[scale=0.8]
    \begin{semilogyaxis}
      [ ymin=1,ymax=10^18,
      xmin=1, xmax=15, domain=1:15,
      xlabel=$\ell$,ylabel=$|\G|$,
    ]
      \addplot[mark=x,mark options={blue,solid},dash pattern=on 3pt off 1pt,blue] table[x=l,y=g2,col sep=comma] {_1,_1,_1_-D4.csv};
      \addplot[mark=triangle,mark options={blue,solid},dash pattern=on 3pt off 1pt,blue] table[x=l,y=g3,col sep=comma] {_1,_1,_1_-D4.csv};
      \addplot[mark=square,mark options={blue,solid},dash pattern=on 3pt off 1pt,blue] table[x=l,y=g1,col sep=comma] {_1,_1,_1_-D4-s.csv};
    \end{semilogyaxis}
  \end{tikzpicture}
  \caption{\(C_{\text{adm}}\)}
  \vspace*{20mm}
  \end{subfigure}
\begin{subfigure}[t]{0.5\textwidth}
  \centering
  \begin{tikzpicture}[scale=0.8]
    \begin{semilogyaxis}
      [ ymin=1,ymax=10^18,
      xmin=1, xmax=15, domain=1:15,
      xlabel=$\ell$,ylabel=$|\G|$,
    ]
      \addplot[mark=x,mark options={indigo,solid},dash pattern=on 4pt off 1pt,indigo] table[x=l,y=g2,col sep=comma] {_4,_4,_4_-D4.csv};
      \addplot[mark=triangle,mark options={indigo,solid},dash pattern=on 4pt off 1pt,indigo] table[x=l,y=g3,col sep=comma] {_4,_4,_4_-D4.csv};
      \addplot[mark=square,mark options={indigo,solid},dash pattern=on 4pt off 1pt,indigo] table[x=l,y=g1,col sep=comma] {_4,_4,_4_-D4-s.csv};
    \end{semilogyaxis}
  \end{tikzpicture}
  \caption{\(C_{\text{imp}}\)}\vspace*{20mm}
  \end{subfigure}
\caption{Comparison  of the number of option sets in the disjunctive generator of an assessment \(\Ass_{1:\ell}\) of size \(\ell\) (horizontal axis), each pair \((V,W)\in \Ass_{1:\ell}\) containing between two and eight options in total, using the first method (\raisebox{-0.3ex}{\protect\myTikzMark{x}}), the second method (\raisebox{-0.3ex}{\protect\myTikzMark{triangle}}) or the third method (\raisebox{-0.3ex}{\protect\myTikzMark{square}}). The results are the average of 7 individual experiments and plot with logarithmic vertical axes.}
\vspace*{10mm}
\label{fig:compchose}
\end{figure}

We start by studying the size of the constructed disjunctive generator as a function of the number \(\ell\) of pairs in the assessment \(\Ass_{1:\ell}\).
The results are depicted in \cref{fig:compchose}.
We consider 7 random assessments \(\Ass_{1:L}\) and depict the average size of the generator over these assessments.
As explained in \cref{sec2:set-up}, for the first two methods, the size of the generators can easily be obtained from the conjunctive generators using the formula 
\(
  |\G| =  \prod_{H\in \D} |H|,
\)
whereas for the third method we explicitly construct the generator \(\G_{\square}\) to determine its size.

In the linear case with a choice function \(C_{\text{lin}}\), we found that every option set \(C_{\text{lin}}(A_k)\) in the assessment was a singleton and the conjunctive generators \(\D_{\times}\) and \(\D_{\triangle}\) were therefore just collections of singletons, resulting in generators \(\G_{\times}\), \(\G_{\triangle}\) and \(\G_{\square}\) consisting of only a single option set.
In the other cases---for \(C_{\text{max}},C_{\text{adm}}\) and \(C_{\text{imp}}\)---the option sets in \(\D\) were not singletons any more.
In these cases, we do however see comparable evolutions in all curves.
As can be seen, for the first two methods (depicted by respectively \myTikzMark{x} and \myTikzMark{triangle}) the size of \(\G\) grows exponentially, but the reduction of each subsequent simplification is considerable.

To compare the different \(\mathcal{E}\)'s we have \cref{fig:compchosev}, where we split the graphs according to the simplification method instead.
The slope for the first method, depicted by \myTikzMark{x}, seems to be slightly lower for \(C_{\text{adm}}\) than for \(C_{\text{max}}\) and \(C_{\text{imp}}\).
Looking at \cref{table1}, the explanation for this is that \(V\) is smaller in that case on average.
So in our experiments, more is rejected by E-admissibility with three randomly chosen mass functions than by maximality with a single lower expectation that is derived from four randomly chosen mass functions.
It is however peculiar that the evolutions of \(C_{\text{max}}\) and \(C_{\text{imp}}\) are so close to each other. 
This seems to indicate that the lower number of rejections of \(C_{\text{imp}}\)---which means less option sets in the conjunctive generator---compensates for the higher number of options in each option set in the conjunctive generator.
\begin{table}
  \centering
  \begin{tabular}{@{}cccc@{}}
    \toprule
     \(\{\E\}\) (lin.) & \(\{\Eu\}\) (max.) & \(\{\E_1,\E_2,\E_3\}\) (E-adm.) & \(\{\Eu_1,\Eu_2,\Eu_3\}\) (imp.) \\
    \midrule
      1.000 & 2.142 & 1.744 & 2.780 \\
    \bottomrule
  \end{tabular}
  \vspace{0.5cm}
  \caption{The average size of \(V_{\ell}=C_{\mathcal{E}}(A_{\ell})\) in the pairs \((V_{\ell},W_{\ell})\in \Ass\), averaged over all pairs and all assessments for different \(\mathcal{E}\).}
  \label{table1}
  \vspace{1cm}
\end{table}



Examining the simplifications of the third method, depicted by the squares (\myTikzMark{square}) in \cref{fig:compchosev}, we see that the number of option sets in the simplified generator (\(\G_{\square}\)) is always relatively small. 
\cref{fig:compchose2} depicts the average size of \(\G_{\square}\) over 7 experiments as a function of the length of the assessment for each of the four generating choice functions.
In this figure we also extend the range to \(\ell=30\) pairs of option sets.
There we see that for \(C_{\text{imp}}\) this seems to increase relatively rapidly while for \(C_{\text{adm}}\) and \(C_{\text{max}}\) it seems to stabilise.
For \(C_{\text{imp}}\), in the worst case of the 7 experiments \(\G_{\square}\) contains a maximum of 45 714 option sets, which was for the full assessment  of length 30, and in the best case it contains a maximum of 4 925 option sets, again for the full assessment, which is still a lot higher than for the other \(\mathcal{E}\)'s.
For the other \(\mathcal{E}\)'s the size of \(\G_{\square}\) stabilises and eventually even starts to decrease with increasing \(\ell\).

\begin{figure}
  \vspace{1cm}

\begin{subfigure}[t]{0.5\textwidth}
  \centering
  \begin{tikzpicture}[scale=0.8]
    \begin{semilogyaxis}
      [ ymin=1,ymax=10^18,
      xmin=1, xmax=15, domain=1:15,
      xlabel=$\ell$,ylabel=$|\G_{\times}|$,
    ]
      \addplot[mark=x,dash pattern=on 1pt off 1pt,mark options={LimeGreen,solid},LimeGreen] table[x=l,y=g2,col sep=comma] {_1_-D4.csv};
      \addplot[mark=x,dash pattern=on 2pt off 1pt, mark options={magenta,solid},magenta] table[x=l,y=g2,col sep=comma] {_4_-D4.csv};
      \addplot[mark=x,dash pattern=on 3pt off 1pt, mark options={blue,solid},blue] table[x=l,y=g2,col sep=comma] {_1,_1,_1_-D4.csv};
      \addplot[mark=x,dash pattern=on 4pt off 1pt, mark options={indigo,solid},indigo] table[x=l,y=g2,col sep=comma] {_4,_4,_4_-D4.csv};
    \end{semilogyaxis}
  \end{tikzpicture}
  \caption{first method}
  \vspace*{20mm}
  \end{subfigure}
  \begin{subfigure}[t]{0.5\textwidth}
    \centering
\begin{tikzpicture}[scale=0.8]
  \begin{semilogyaxis}
    [ ymin=1,ymax=10^17,
      xmin=1, xmax=15, domain=1:15,
      xlabel=$\ell$,ylabel=$|\G_{\triangle}|$,
    ]
    \addplot[mark=triangle,dash pattern=on 1pt off 1pt,mark options={LimeGreen,solid},LimeGreen] table[x=l,y=g3,col sep=comma] {_1_-D4.csv};
    \addplot[mark=triangle,dash pattern=on 2pt off 1pt, mark options={magenta,solid},magenta] table[x=l,y=g3,col sep=comma] {_4_-D4.csv};
    \addplot[mark=triangle,dash pattern=on 3pt off 1pt, mark options={blue,solid},blue] table[x=l,y=g3,col sep=comma] {_1,_1,_1_-D4.csv};
    \addplot[mark=triangle,dash pattern=on 4pt off 1pt, mark options={indigo,solid},indigo] table[x=l,y=g3,col sep=comma] {_4,_4,_4_-D4.csv};
  \end{semilogyaxis}
\end{tikzpicture}
\caption{second method}
\vspace*{20mm}
\end{subfigure}
\begin{subfigure}[t]{0.5\textwidth}
  \centering
  \begin{tikzpicture}[scale=0.8]
    \begin{semilogyaxis}
      [ ymin=1,ymax=10^17,
      xmin=1, xmax=15, domain=1:15,
      xlabel=$\ell$,ylabel=$|\G_\square|$,
    ]
      \addplot[mark=square,dash pattern=on 1pt off 1pt,mark options={LimeGreen,solid},LimeGreen] table[x=l,y=g1,col sep=comma] {_1_-D4-s.csv};
      \addplot[mark=square,dash pattern=on 2pt off 1pt, mark options={magenta,solid},magenta] table[x=l,y=g1,col sep=comma] {_4_-D4-s.csv};
      \addplot[mark=square,dash pattern=on 3pt off 1pt, mark options={blue,solid},blue] table[x=l,y=g1,col sep=comma] {_1,_1,_1_-D4-s.csv};
      \addplot[mark=square,dash pattern=on 4pt off 1pt, mark options={indigo,solid},indigo] table[x=l,y=g1,col sep=comma] {_4,_4,_4_-D4-s.csv};
    \end{semilogyaxis}
  \end{tikzpicture}
  \caption{third method}
  \vspace*{20mm}
\end{subfigure}
\begin{subfigure}[t]{0.5\textwidth}
  \centering
  \begin{tikzpicture}[scale=0.8]
    \begin{semilogyaxis}
      [ ymin=1,ymax=4*10^2,
      xmin=1, xmax=15, domain=1:15,
      xlabel=$\ell$,ylabel=$|\G_\square|$,
    ]
      \addplot[mark=square,dash pattern=on 1pt off 1pt,mark options={LimeGreen,solid},LimeGreen] table[x=l,y=g1,col sep=comma] {_1_-D4-s.csv};
      \addplot[mark=square,dash pattern=on 2pt off 1pt, mark options={magenta,solid},magenta] table[x=l,y=g1,col sep=comma] {_4_-D4-s.csv};
      \addplot[mark=square,dash pattern=on 3pt off 1pt, mark options={blue,solid},blue] table[x=l,y=g1,col sep=comma] {_1,_1,_1_-D4-s.csv};
      \addplot[mark=square,dash pattern=on 4pt off 1pt, mark options={indigo,solid},indigo] table[x=l,y=g1,col sep=comma] {_4,_4,_4_-D4-s.csv};
    \end{semilogyaxis}
  \end{tikzpicture}
  \caption{third method (vertically zoomed)}
  \vspace*{20mm}
  \end{subfigure}
\caption{Comparison  of the number of option sets in the disjunctive generator of an assessment \(\Ass_{1:\ell}\) of size \(\ell\) (horizontal axis), each pair \((V,W)\in \Ass_{1:\ell}\) containing between two and eight options in total, and constructed using \(C_{\text{lin}}\) (\textcolor{LimeGreen}{\raisebox{-0.3ex}{\protect\colorMark{oplus*}{LimeGreen}{1}}}), \(C_{\text{max}}\) (\textcolor{magenta}{\raisebox{-0.3ex}{\protect\colorMark{oplus*}{magenta}{3}}}), \(C_{\text{adm}}\) (\textcolor{blue}{\raisebox{-0.3ex}{\protect\colorMark{oplus*}{blue}{2}}}) or \(C_{\text{imp}}\) (\textcolor{indigo}{\raisebox{-0.3ex}{\protect\colorMark{oplus*}{indigo}{4}}}). The results are the average of 7 individual experiments and plot with logarithmic vertical axes.}
\vspace*{10mm}
\label{fig:compchosev}
\end{figure}

\begin{figure}
   \centering
\begin{tikzpicture}
  \begin{axis}
    [ ymin=0,ymax=1000,
      xmin=1, xmax=30, domain=1:30,
      xlabel=$\ell$,ylabel=$|\G_{\square}|$,
    ]
    \addplot[mark=square,dash pattern=on 1pt off 1pt,mark options={LimeGreen,solid},LimeGreen] table[x=l,y=g1,col sep=comma] {_1_-D4-s.csv};\label{groenkotje}
      \addplot[mark=square,dash pattern=on 2pt off 1pt,mark options={magenta,solid},magenta] table[x=l,y=g1,col sep=comma] {_4_-D4-s.csv};\label{roodkotje}
    \addplot[mark=square,dash pattern=on 3pt off 1pt,mark options={blue,solid},blue] table[x=l,y=g1,col sep=comma] {_1,_1,_1_-D4-s.csv};\label{blauwkotje}
      \addplot[mark=square,dash pattern=on 4pt off 1pt,mark options={indigo,solid},indigo] table[x=l,y=g1,col sep=comma] {_4,_4,_4_-D4-s.csv};\label{paarskotje}
    \end{axis}
  \end{tikzpicture}
  \vspace*{5mm}
\caption{Comparison of the number of option sets in the simplified disjunctive generator~\(\G_{\square}\) of an assessment \(\Ass_{1:\ell}\) of size \(\ell\) (horizontal axis), each pair \((V,W)\in \Ass_{1:\ell}\) containing between two and eight options in total, and constructed using \(C_{\text{lin}}\) (\ref{groenkotje}), \(C_{\text{max}}\) (\ref{roodkotje}), \(C_{\text{adm}}\) (\ref{blauwkotje}) and \(C_{\text{imp}}\) (\ref{paarskotje}). The results are the average of 7 individual experiments and shown in a non-logarithmic plot.}
\label{fig:compchose2}
\vspace*{10mm}
\end{figure}

\subsection{Varying the levels of imprecision}
In the second experiment we look at \(\epsilon\)-contaminations.
These are lower expectations that are mixtures of a precise expectation with the so-called `vacuous expectation', which we then use to choose with maximality.
For any linear expectation \(\E\) and an \(\epsilon \in [0,1]\), we can make an \(\epsilon\)-contamination with the vacuous model as for example described in \cite[Section~4.7.3]{augustin2014introduction}, which leads to the corresponding lower expectation \[\Eu^{\epsilon}\colon \V \to \R \colon u \mapsto (1-\epsilon) E(u) + \epsilon \min {u}.\]
It is not difficult to see that the extreme probability mass functions of \(\Eu\) are \(\{(1-\epsilon)p + \epsilon \mathbb{I}_x \colon x \in \X\}\), where \(\mathbb{I}_x\colon \X\to \R \) is the standard basis vector that is \(0\) everywhere except in \(x\) where it is equal to \(1\).
Since we choose with maximality, the corresponding choice function \(C_{\text{max}}=C_{\{\Eu^\epsilon\}}\) is given by
\begin{align*}
  C_{\text{max}}(A)&=
  C_{\{\Eu^\epsilon\}}(A)
  =\left\{u \in A \colon (\forall v \in A) \; {\Eu^\epsilon}(v-u) \leq 0 \text{ and }u\not< v   \right\}\\
  &=\left\{u \in A \colon (\forall v \in A-u) \; {\Eu^\epsilon}(v) \leq 0 \text{ and }0\not< v  \right\}\\
  &=\left\{u \in A \colon (\forall v \in A-u) \; (1-\epsilon) E(v) + \epsilon \min {v} \leq 0 \text{ and }0\not< v  \right\}\\
  &=\left\{u \in\! A \colon (\forall v \in A-u)\; [ (1-\epsilon)E(v) \leq -\epsilon \min v  \text{ and }  0 \not< v]  \right\}  
\end{align*}
for every \(A\in \Q\) and \(\epsilon\in [0,1]\).
Intuitively \(\epsilon\) can be seen as an indication of the amount of imprecision in the choice function.
We see that for \(\epsilon=0\) we have a precise choice function \(C_{\text{lin}}\), and for \(\epsilon=1\) we have the vacuous choice function\footnote{For a given option set \(A\in \Q\), this choice function only rejects an option \(u\in A\) if it is dominated by some other \(v\in A\) in the sense that \(u<v\).}~\(C_{<}\) because then the first condition \(0\leq -\min v\) is implied by the second condition \(0\not< v\).
In between this \(\epsilon\) induces a continuous transformation between the two.


We are interested in the effect of imprecision, quantified by \(\epsilon\), on the size of the conjunctive and disjunctive generator.
To investigate this, we repeat the following sub-experiment 100 times and take the average of the number of option sets in the resulting generators.
First we generate one probability mass function \(p\) and an array of \aantaloptionsetsexp option sets \(A_1,...,A_{\aantaloptionsetsexp}\).
Then we loop over \(\epsilon\) from \(0.03\) to \(0.99\) in steps of \(0.03\).
For each \(\epsilon\) we created the assessment \(\Ass(C_{\{\Eu^\epsilon\}},A_1,...,A_{\aantaloptionsetsexp})\).
Next, we determine the size of the conjunctive generators \(\HH_{\times}\) and \(\HH_{\triangle}=\HH_\square\) as well as that of the disjunctive generators \(\G_{\times},\G_{\triangle}\) and \(\G_{\square}\) that are constructed by each of our three methods, respectively.
For the first two methods we can still use \(
  |\G| =  \prod_{H\in \D} |H|
\) to determine the size of the disjunctive generator without actually constructing it in memory.
For the third method we need to construct the disjunctive generator explicitly to determine its size.

In \cref{fig:epsilonvar2}, we have plotted the evolution of the size of the conjunctive generators \(\HH_{\times}\) and \(\HH_{\triangle}\) as a function of \(\epsilon\).
On the one hand, in the precise case \(\epsilon=0\), we expect to only retain a single chosen option in the \(V\)-sets of the assessments and for the \(W\)-sets to contain a number of rejected options that ranges from 1 to 7, uniformly so.
Since we have 10 tuples \((V,W)\) in the assessment, we therefore expect (on average) to have approximately \(10 \frac{1+7}2 = 40\) option sets in the conjunctive generator, without simplifications.
On the other hand, in the fully vacuous case, we have some option sets in the unsimplified conjunctive generator from the options that are rejected by \(C_{<}\), but since \(C_{<}\) is the most conservative coherent choice function, these impose no constraints and are therefore uninformative.
We therefore expect to have no option sets in the simplified conjunctive generator.
This is indeed what we see in the plot.
Concretely, this happens because they are removed by \cref{prop:simpDer2} in \cref{alg:CreateDer}.
In between the two extremes, the curves seem to have a more or less linear downward trend.

\begin{figure}
  \centering
\begin{tikzpicture}
  \begin{axis}[
    xlabel={$\epsilon$},
    ylabel={ $|\HH_{\times}|$ \ref{plotzwart3},$|\HH_{\triangle}|$ \ref{plotzwart}}
    ] 
    \addplot[mark=triangle,mark options={magenta,solid},magenta] table[x=epsilon,y=d2,col sep=comma
    ] {epsilons-10.csv};\label{plotzwart}
    \addplot[mark=x,mark options={magenta,solid},magenta] table[x=epsilon,y=d1,col sep=comma
    ] {epsilons-10.csv};\label{plotzwart3}
\end{axis}

\end{tikzpicture}
\vspace*{5mm}
\caption{Total size of the conjunctive generator of an assessment with size \aantaloptionsetsexp  as a function of \(\epsilon\), the amount of contamination of an expectation with the vacuous model, with and without simplifications. The first method, without simplifications, is indicated by \raisebox{-0.3ex}{\protect\myTikzMark{x}}. The second method, with simplifications, is indicated by \raisebox{-0.3ex}{\protect\myTikzMark{triangle}}. (The third method has the same conjunctive generator as the second.) The results are the average of 100 individual experiments.}
\vspace*{10mm}
\label{fig:epsilonvar2}
\end{figure}

\begin{figure}
  \centering
\begin{tikzpicture}
  \begin{semilogyaxis}[
    ymin=10^0,
    xlabel={$\epsilon$},
    ylabel={ $|\G_{\times}|$ \ref{plotzwart3},$|\G_{\triangle}|$ \ref{plotzwart},$|\G_{\square}|$ \ref{plotzwart4}},
    ] 
    \addplot[mark=triangle,mark options={magenta,solid},magenta] table[x=epsilon,y=g2,col sep=comma
    ] {epsilons-10.csv};\label{plotzwart}
    \addplot[mark=x,mark options={magenta,solid},magenta] table[x=epsilon,y=g1_x,col sep=comma
    ] {epsilons-10.csv};\label{plotzwart3}
    \addplot[mark=square,mark options={magenta,solid},magenta] table[x=epsilon,y=g1_y,col sep=comma
    ] {epsilons-10.csv};\label{plotzwart4}
\end{semilogyaxis}

\end{tikzpicture}
\vspace*{5mm}
\caption{Total size of the disjunctive generator of an assessment with size \aantaloptionsetsexp  as a function of \(\epsilon\), the amount of contamination of an expectation with the vacuous model, using three different methods. The first method is indicated by \raisebox{-0.3ex}{\protect\myTikzMark{x}}, the second method is indicated by \raisebox{-0.3ex}{\protect\myTikzMark{triangle}} and the third by \raisebox{-0.3ex}{\protect\myTikzMark{square}}. The results are the average of 100 individual experiments and plot with a logarithmic vertical axis.}
\vspace*{10mm}
\label{fig:epsilonvar}
\end{figure}
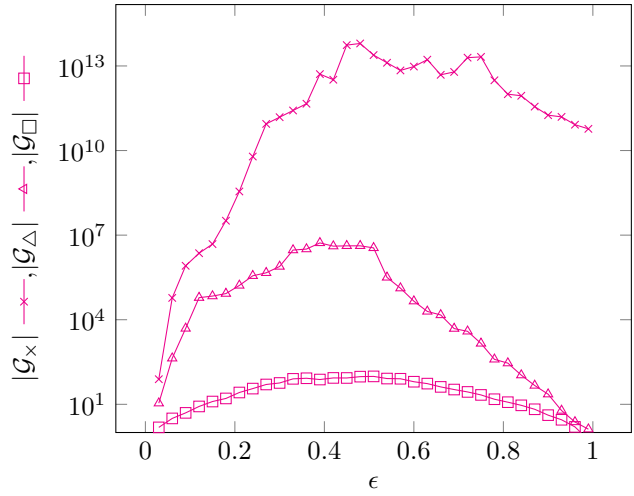

In \cref{fig:epsilonvar}, we have plotted the number of option sets in \(\G_{\times}\), \(\G_{\triangle}\) and \(\G_{\square}\) on the vertical axis as a function of \(\epsilon\) on the horizontal axis. 
The number of option sets in the generators starts low for all three methods, because in the precise case most options are rejected and therefore every \(|H|\) in the product \(
  |\G| =  \prod_{H\in \D} |H|
\) is small.
But when we increase the imprecision, fewer and fewer options are rejected and the number of option sets in the generator \(\G_{\times}\) starts to increase.
It seems that adding imprecision makes things worse, but that this trend does not keep going.
Rejecting fewer options also means that the number of option sets in \(\D_{\times}\) decreases.
At some point the low number of option sets in \(\D_{\times}\) outweighs the fact that the number of options in these option sets increases, and then \(|\G_{\times}|\) decreases.
The generator \(\G_{\triangle}\) starts out similar to \(\G_{\times}\) but differs more from it as imprecision increases.
The reason is that as imprecision increases, the rejections become less and less informative.
For example, it becomes more likely that an option that is rejected, is rejected because of the ordering~\(<\) in which case this option set will be removed from the conjunctive generator \(\D_{\triangle}\) by \cref{prop:simpDer2} in \cref{alg:CreateDer} but not from \(\D_{\times}\).
The generator \(\G_{\square}\) follows a similar trend as \(\G_{\triangle}\), but it is smaller because there is more simplification, and it is also smoother.

\subsection{Time benchmarking: creation vs. evaluation}

In our final experiment, we focus on the actual time it takes to run the algorithms, rather than the size of the generators.
We measured two things:
\begin{enumerate}
  \item the time it takes to `construct' a disjunctive generator for an assessment of a given size for each of the three methods; for the first two methods we mean with `construct' that we find \(\D_{\times}\) and \(\D_{\triangle}\)---since \(\G_\times\) and \(\G_\triangle\) are constructed on the fly without storing them in memory---while 
  for the last method we mean that we find \(\G_{\square}\),
  \item the time to evaluate the natural extension of a given option set, again for a given size of assessment, using \cref{alg:Choose} with either \(\G_{\times}\), \(\G_{\triangle}\) or \(\G_{\square}\).
  For the first two methods this time also includes the time to construct the generator sets on the fly.
\end{enumerate}
The timings are done using the BenchmarkTools Julia package.
Everything else is the same as in \cref{subsec:exp1}.

\begin{figure}
  \begin{subfigure}[t]{0.5\textwidth}
     \centering
\begin{tikzpicture}[scale=0.8]
  \begin{semilogyaxis}
    [ ymin=10^-7,ymax=10^5,
      xmin=1, xmax=15, domain=1:15,
      xlabel=$\ell$,ylabel=time in seconds,
    ]
    \addplot[mark=x,mark options={LimeGreen,solid},LimeGreen] table[x=l,y=d2,col sep=comma] {experiment3/_1_-tijdenchoseD.csv};
    \addplot[mark=triangle,mark options={LimeGreen,solid},LimeGreen] table[x=l,y=d3,col sep=comma] {experiment3/_1_-tijdenchoseD.csv};
    \addplot[mark=square,mark options={LimeGreen,solid},LimeGreen,error bars/.cd, y dir=both, y explicit] table [x=l,y=d1_mean,col sep=comma,y error plus=d1_maxme, y error minus=d1_minme] {experiment3/_1_-tijdenchoseG-stats.csv};
  \end{semilogyaxis}
\end{tikzpicture}
\caption{\(C_{\text{lin}}\)}
\vspace*{20mm}
\end{subfigure}
  \begin{subfigure}[t]{0.5\textwidth}
     \centering
\begin{tikzpicture}[scale=0.8]
  \begin{semilogyaxis}
    [ ymin=10^-7,ymax=10^5,
      xmin=1, xmax=15, domain=1:15,
      xlabel=$\ell$,ylabel=time in seconds,
    ]
    \addplot[mark=x,mark options={magenta,solid},magenta] table[x=l,y=d2,col sep=comma] {experiment3/_4_-tijdenchoseD.csv};
    \addplot[mark=triangle,mark options={magenta,solid},magenta] table[x=l,y=d3,col sep=comma] {experiment3/_4_-tijdenchoseD.csv};
    \addplot[mark=square,mark options={magenta,solid},magenta,error bars/.cd, y dir=both, y explicit] table [x=l,y=d1_mean,col sep=comma,y error plus=d1_maxme, y error minus=d1_minme] {experiment3/_4_-tijdenchoseG-stats.csv};
  \end{semilogyaxis}
\end{tikzpicture}
\caption{\(C_{\text{max}}\)}\vspace*{20mm}
\end{subfigure}
\begin{subfigure}[t]{0.5\textwidth}
  \centering
  \begin{tikzpicture}[scale=0.8]
    \begin{semilogyaxis}
      [ymin=10^-7,ymax=10^5,
      xmin=1, xmax=15, domain=1:15,
      xlabel=$\ell$,ylabel=time in seconds,
    ]
      \addplot[mark=x,mark options={blue,solid},blue] table[x=l,y=d2,col sep=comma] {experiment3/_1,_1,_1_-tijdenchoseD.csv};
      \addplot[mark=triangle,mark options={blue,solid},blue] table[x=l,y=d3,col sep=comma] {experiment3/_1,_1,_1_-tijdenchoseD.csv};
      \addplot[mark=square,mark options={blue,solid},blue,error bars/.cd, y dir=both, y explicit] table [x=l,y=d1_mean,col sep=comma,y error plus=d1_maxme, y error minus=d1_minme] {experiment3/_1,_1,_1_-tijdenchoseG-stats.csv};
    \end{semilogyaxis}
  \end{tikzpicture}
  \caption{\(C_{\text{adm}}\)}
  \vspace*{20mm}
  \end{subfigure}
\begin{subfigure}[t]{0.5\textwidth}
  \centering
  \begin{tikzpicture}[scale=0.8]
    \begin{semilogyaxis}
      [ymin=10^-7,ymax=10^5,
      xmin=1, xmax=15, domain=1:15,
      xlabel=$\ell$,ylabel=time in seconds,
    ]
      \addplot[mark=x,mark options={indigo,solid},indigo] table[x=l,y=d2,col sep=comma] {experiment3/_4,_4,_4_-tijdenchoseD.csv};
      \addplot[mark=triangle,mark options={indigo,solid},indigo] table[x=l,y=d3,col sep=comma] {experiment3/_4,_4,_4_-tijdenchoseD.csv};
      \addplot[mark=square,mark options={indigo,solid},indigo,error bars/.cd, y dir=both, y explicit] table [x=l,y=d1_mean,col sep=comma,y error plus=d1_maxme, y error minus=d1_minme] {experiment3/_4,_4,_4_-tijdenchoseG-stats.csv};
    \end{semilogyaxis}
  \end{tikzpicture}
  \caption{\(C_{\text{imp}}\)}\vspace*{20mm}
  \end{subfigure}
\caption{Time in seconds to `construct' a disjunctive generator \(\G\) as a function of the number \(\ell\) of pairs \((V,W)\) in the assessment \(\Ass_{1:\ell}\) for the three methods:   \raisebox{-0.3ex}{\protect\myTikzMark{x}}, \raisebox{-0.3ex}{\protect\myTikzMark{triangle}} and \raisebox{-0.3ex}{\protect\myTikzMark{square}} respectively. The indicated values are the average of 7 experiments and the error bars on the third method are the maximum and minimum  value observed in the 7 experiments. The vertical axis is logarithmic.}
\vspace*{10mm}
\label{fig:chosetijd}
\end{figure}
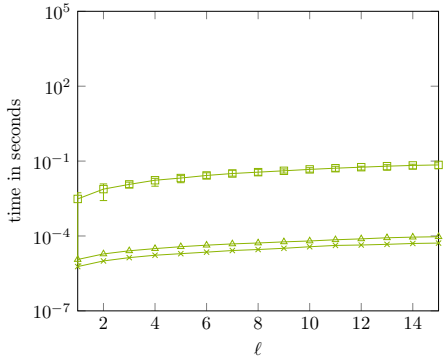
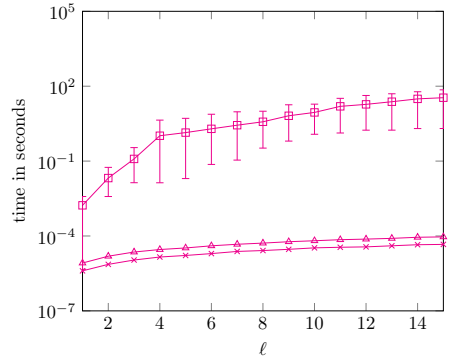
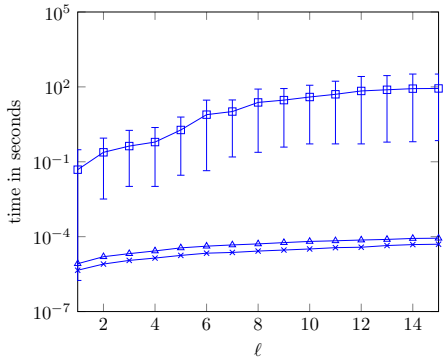
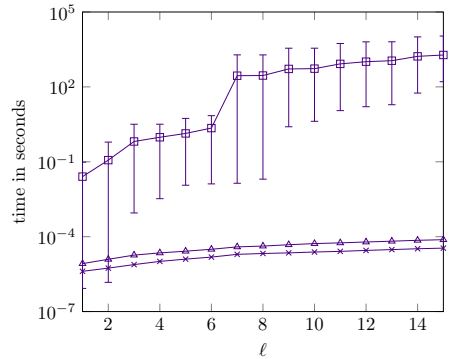

In \cref{fig:chosetijd}, we plot the time in seconds to construct the generator \(\G\) for a given assessment, using our three different methods.
We consider the average over 7 different assessments.
We see that the first method (\myTikzMark{x}) is always the fastest. 
This is because it corresponds to simple subtractions of vectors without additional checks.
At the same time the second method seems to be very close to the first method.
For the third method, it seems to take much longer to construct the generator than for the other two methods, for example for \(C_{\text{imp}}\) for an assessment containing \(15\) pairs of option sets, each containing between two and eight options per pair, it took on average 31 minutes and 36 seconds.
This is because, compared to the second method, for the third method we have to construct \(\G_{\square}\) from \(\D_{\triangle}\) (in memory) while also doing the simplifications that make the difference between \(\G_{\triangle}\) and \(\G_{\square}\). 
For the third method, we also plotted error bars for the minimum and maximum time, but not for the others because they are already very close to each other and the error bars were also very small.
From the error bars it can be seen that there is also a large spread in how long it takes to construct the generator \(\G_{\square}\), except for \(C_{\text{lin}}\).
For \(C_{\text{imp}}\) a large jump can be seen in the number of option sets in the average of \(\G_{\square}\), but this is due to a large jump in the maximum value, as the minimum value does not jump up until later.

In \cref{fig:choosetijd}, we plot the time to choose from a single option set; that is, to evaluate the natural extension \(C_{\Ass_{1:\ell}}\) for a single option set with \cref{alg:Choose}, using either \(\G_{\times}\), \(\G_{\triangle}\) or \(\G_{\square}\).
The points in the figure are an average over the same 7~assessments~\(\Ass_{1:L}\) that we have used in \cref{subsec:exp1}.
Additionally, for each assessment~\(\Ass_{1:L}\), we also generated 7 random options sets to choose from, each containing between 2 and 8 options.
So if the assessments are \(\Ass_{1:L}^1,...,\Ass_{1:L}^7\), we have option sets \(B_1^1,...,B_7^1\) corresponding to \(\Ass_{1:L}^1\) up to \(B_1^7,...,B_7^7\) corresponding to \(\Ass_{1:L}^7\).
Every point on \cref{fig:choosetijd} is the average of the time to calculate \(C_{\Ass_{1:\ell}^r}(B_k^r)\) over \(k\in\{1,...,7\}\) and \(r\in \{1,...,7\}\), for every \(\ell\), using one of our three methods.
The number of options in every option set was chosen uniformly random from \(\{2,...,8\}\), and the options themselves are again generated uniformly random from \([0,1]^4\).
We see in \cref{fig:choosetijd} that our simplifications indeed work, in the sense that more simplifications result in smaller choosing times.
This is what we expected, as a smaller generator will lead to a faster loop over the generator in \cref{alg:Choose}.


Next, in \cref{fig:somtijd}, we plot the sum of the time it takes to construct the generator and the time it takes to use this generator to evaluate the natural extension using \cref{alg:Choose}.
For this plot the data points stop when the calculations take more than 50 minutes on average.
This in fact already occurred in \cref{fig:choosetijd}, but is not visible there because we only plot up to 100 seconds there.
For all methods, we see that it is initially faster to use the second method but that the third method eventually becomes faster for larger assessments.
How fast this happens depends on the type of assessment, and we see that for \(C_{\text{max}}\) and \(C_{\text{adm}}\) it happens faster than for \(C_{\text{lin}}\) and \(C_{\text{imp}}\).
For \(C_{\text{imp}}\), it may seem as if the third method is never better, but this is not the case because from \(\ell=13\) onwards, the results of the second method are not displayed because the calculations took too long; so for these higher values of \(\ell\), the third method is again better.

\begin{figure}
\begin{subfigure}[t]{0.5\textwidth}
   \centering
  \begin{tikzpicture}[scale=0.8]
    \begin{axis}[xlabel=$\ell$,ylabel=$n$, ]
    \addplot[mark=pentagon,mark options={LimeGreen,solid},LimeGreen] table[x=l,y=fraction_result,col sep=comma,only marks] {ratios/ratios-_1_-simplify.csv};
    \end{axis}
  \end{tikzpicture}
  \caption{\(C_{\text{lin}}\)}
  \vspace*{20mm}
\end{subfigure}
  \begin{subfigure}[t]{0.5\textwidth}
    \centering
  \begin{tikzpicture}[scale=0.8]
    \begin{axis}[xlabel=$\ell$,ylabel=$n$, ]
    \addplot[mark=pentagon,mark options={indigo,solid},indigo] table[x=l,y=fraction_result,col sep=comma,only marks] {ratios/ratios-_4,_4,_4_-simplify.csv};
    \end{axis}
  \end{tikzpicture}
  \caption{\(C_{\text{imp}}\)}
  \vspace*{20mm}
  \end{subfigure}
\caption{The vertical axis depicts for how many option sets with (between 2 and 8 options each) you would have to evaluate the natural extension before it is better to use the third method rather than the second method for \(C_{\text{lin}}\) and \(C_{\text{imp}}\). 
For \(C_{\text{imp}}\) the plot only goes up to \(\ell=12\) because for higher \(\ell\) it took too long to evaluate the natural extension using the second method.
}
\vspace*{10mm}
\label{fig:ratio}
\end{figure}
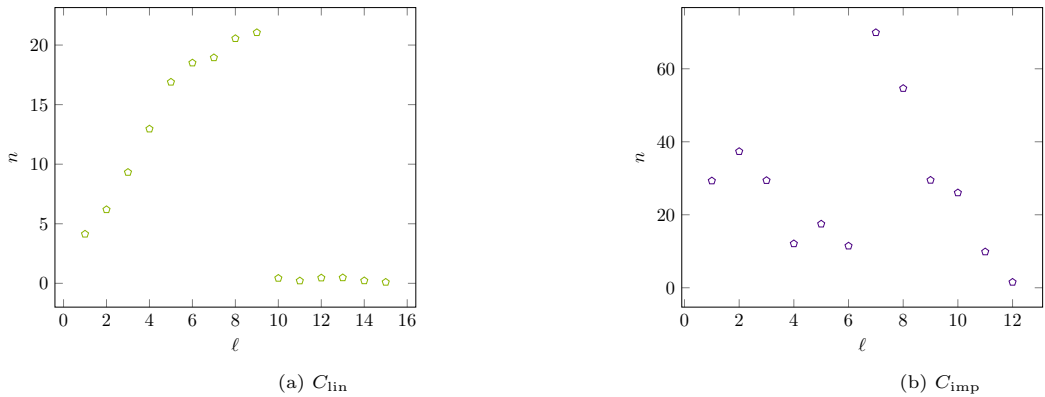

\cref{fig:somtijd} only considers the time for making one choice though.
If we evaluate the natural extension for several option sets, for the same assessment, then there will come a point where the fact that the time investment \(t_{\text{pre},\square}\) for the preprocessing of the third method is higher than to the time \(t_{\text{pre},\triangle}\) for the preprocessing of the second method is compensated by the fact that the time \(t_{\text{choose},\square}\) to evaluate the natural extension for the third method is lower than the time \(t_{\text{choose},\triangle}\) to evaluate the natural extension for the second method.
The number of evaluations \(n\) at which the total times for both methods break even is the smallest \(n\) such that \(t_{\text{pre},\square}+n t_{\text{choose},\square} \leq t_{\text{pre},\triangle}+n t_{\text{choose},\triangle}\).
If \(t_{\text{choose},\triangle}>t_{\text{choose},\square}\)---which is true in all our experiments---solving for \(n\) gives
\[
  n \geq \frac{t_{\text{pre},\square}- t_{\text{pre},\triangle}}{t_{\text{choose},\triangle}- t_{\text{choose},\square}}.
\]
We plot the right-hand side of this inequality in \cref{fig:ratio} for the two most interesting cases: \(C_{\text{lin}}\) and \(C_{\text{imp}}\).
The main conclusion is that even for small assessments, the third method is still faster  than the second method, provided that the number \(n\) of option sets that we need to evaluate is high enough.
For \(C_{\text{lin}}\) we found the highest break-even point for \(\ell=9\), for which the third method is faster if we need to evaluate the natural extension for at least 21 option sets.
For \(C_{\text{imp}}\), we found that the highest value was for \(\ell=7\), for which we would need to evaluate at least 70 option sets before the third method becomes faster.


\begin{figure}
  \begin{subfigure}[t]{0.5\textwidth}
     \centering
\begin{tikzpicture}[scale=0.8]
  \begin{semilogyaxis}
    [ ymin=5*10^-3,ymax=10^2,
      xmin=1, xmax=15,
      xlabel=$\ell$,ylabel=time in seconds,
    ]
    \addplot[mark=x,mark options={LimeGreen,solid},LimeGreen] table[x=l,y=d1,col sep=comma] {kiestijden/_1_-stupid.csv};
    \addplot[mark=triangle,mark options={LimeGreen,solid},LimeGreen] table[x=l,y=d1,col sep=comma] {kiestijden/_1_-simplify.csv};
    \addplot[mark=square,mark options={LimeGreen,solid},LimeGreen] table[x=l,y=d1,col sep=comma] {kiestijden/_1_-xsimplify.csv};
  \end{semilogyaxis}
\end{tikzpicture}
\caption{\(C_{\text{lin}}\)}
\vspace*{20mm}
\end{subfigure}
\begin{subfigure}[t]{0.5\textwidth}
  \centering
\begin{tikzpicture}[scale=0.8]
  \begin{semilogyaxis}
    [ ymin=5*10^-3,ymax=10^2,
    xmin=1, xmax=15,
    xlabel=$\ell$,ylabel=time in seconds, 
    ]
    \addplot[mark=x,mark options={magenta,solid},magenta,domain=1:4,select coords between index={0}{3}] table[x=l,y=d1,col sep=comma] {kiestijden/_4_-stupid.csv};
    \addplot[mark=triangle,mark options={magenta,solid},magenta] table[x=l,y=d1,col sep=comma] {kiestijden/_4_-simplify.csv};
    \addplot[mark=square,mark options={magenta,solid},magenta] table[x=l,y=d1,col sep=comma] {kiestijden/_4_-xsimplify.csv};
  \end{semilogyaxis}
\end{tikzpicture}
\caption{\(C_{\text{max}}\)}\vspace*{20mm}
\end{subfigure}
\begin{subfigure}[t]{0.5\textwidth}
  \centering
  \begin{tikzpicture}[scale=0.8]
    \begin{semilogyaxis}
      [ymin=5*10^-3,ymax=10^2,
      xmin=1, xmax=15,
      xlabel=$\ell$,ylabel=time in seconds, 
    ]
    \addplot[mark=x,mark options={blue,solid},blue] table[x=l,y=d1,col sep=comma] {kiestijden/_1,_1,_1_-stupid.csv};
    \addplot[mark=triangle,mark options={blue,solid},blue] table[x=l,y=d1,col sep=comma] {kiestijden/_1,_1,_1_-simplify.csv};
    \addplot[mark=square,mark options={blue,solid},blue] table[x=l,y=d1,col sep=comma] {kiestijden/_1,_1,_1_-xsimplify.csv};
    \end{semilogyaxis}
  \end{tikzpicture}
  \caption{\(C_{\text{adm}}\)}
  \vspace*{20mm}
  \end{subfigure}
\begin{subfigure}[t]{0.5\textwidth}
  \centering
  \begin{tikzpicture}[scale=0.8]
    \begin{semilogyaxis}
      [ymin=5*10^-3,ymax=10^2,
      xmin=1, xmax=15,
      xlabel=$\ell$,ylabel=time in seconds, 
    ]
    \addplot[mark=x,mark options={indigo,solid},indigo] table[x=l,y=d1,col sep=comma] {kiestijden/_4,_4,_4_-stupid.csv};
    \addplot[mark=triangle,mark options={indigo,solid},indigo] table[x=l,y=d1,col sep=comma] {kiestijden/_4,_4,_4_-simplify.csv};
    \addplot[mark=square,mark options={indigo,solid},indigo] table[x=l,y=d1,col sep=comma] {kiestijden/_4,_4,_4_-xsimplify.csv};
    \end{semilogyaxis}
  \end{tikzpicture}
  \caption{\(C_{\text{imp}}\)}\vspace*{20mm}
  \end{subfigure}
\caption{Time in seconds to evaluate the natural extension for a single option set using \(\G_{\times}\), \(\G_{\triangle}\) or \(\G_{\square}\), as a function of the number \(\ell\) of pairs \((V,W)\) in the assessment. The vertical axis is logarithmic, and the experiments are averaged over 7 separate experiments and 7 option sets to choose from per experiment.}
\vspace*{10mm}
\label{fig:choosetijd}
\end{figure}
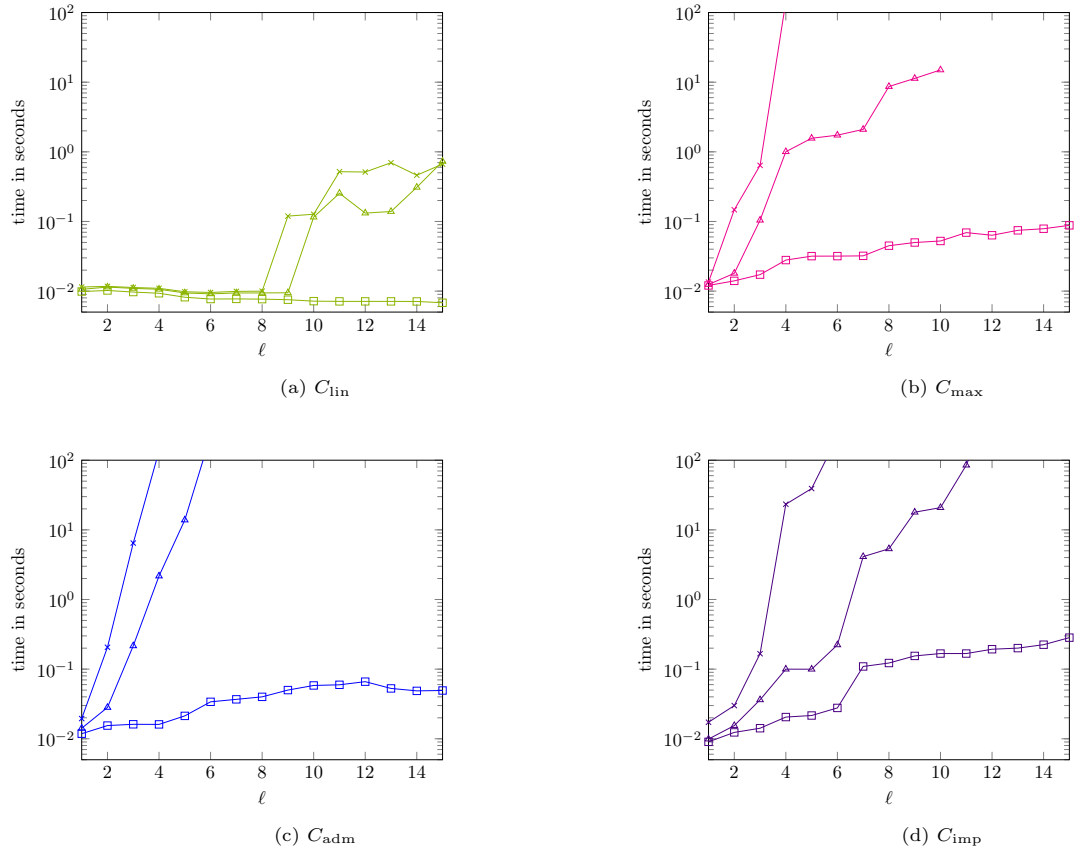

\begin{figure}
  \begin{subfigure}[t]{0.5\textwidth}
     \centering
\begin{tikzpicture}[scale=0.8]
  \begin{semilogyaxis}
    [ymin=5*10^-3,ymax=5*10^3,
      xmin=1, xmax=15, 
      xlabel=$\ell$,ylabel=time in seconds, 
    ]
    \addplot[mark=x,mark options={LimeGreen,solid},LimeGreen] table[x=l,y=d1,col sep=comma] {koos+kiestijden/_1_-stupidcombo.csv};
    \addplot[mark=triangle,mark options={LimeGreen,solid},LimeGreen] table[x=l,y=d1,col sep=comma] {koos+kiestijden/_1_-simplifycombo.csv};
    \addplot[mark=square,mark options={LimeGreen,solid},LimeGreen] table[x=l,y=d1,col sep=comma] {koos+kiestijden/_1_-xsimplifycombo.csv};
  \end{semilogyaxis}
\end{tikzpicture}
\caption{\(C_{\text{lin}}\)}
\vspace*{20mm}
\end{subfigure}
\begin{subfigure}[t]{0.5\textwidth}
  \centering
\begin{tikzpicture}[scale=0.8]
\begin{semilogyaxis}
  [ ymin=5*10^-3,ymax=5*10^3,
  xmin=1, xmax=15, 
  xlabel=$\ell$,ylabel=time in seconds, 
  ]
  \addplot[mark=x,mark options={magenta,solid},magenta,domain=1:4,select coords between index={0}{3}] table[x=l,y=d1,col sep=comma] {koos+kiestijden/_4_-stupidcombo.csv};
  \addplot[mark=triangle,mark options={magenta,solid},magenta] table[x=l,y=d1,col sep=comma] {koos+kiestijden/_4_-simplifycombo.csv};
  \addplot[mark=square,mark options={magenta,solid},magenta] table[x=l,y=d1,col sep=comma] {koos+kiestijden/_4_-xsimplifycombo.csv};
\end{semilogyaxis}
\end{tikzpicture}
\caption{\(C_{\text{max}}\)}\vspace*{20mm}
\end{subfigure}
\begin{subfigure}[t]{0.5\textwidth}
  \centering
  \begin{tikzpicture}[scale=0.8]
    \begin{semilogyaxis}
      [ymin=5*10^-3,ymax=5*10^3,
      xmin=1, xmax=15, 
      xlabel=$\ell$,ylabel=time in seconds, 
    ]
    \addplot[mark=x,mark options={blue,solid},blue] table[x=l,y=d1,col sep=comma] {koos+kiestijden/_1,_1,_1_-stupidcombo.csv};
    \addplot[mark=triangle,mark options={blue,solid},blue] table[x=l,y=d1,col sep=comma] {koos+kiestijden/_1,_1,_1_-simplifycombo.csv};
    \addplot[mark=square,mark options={blue,solid},blue] table[x=l,y=d1,col sep=comma] {koos+kiestijden/_1,_1,_1_-xsimplifycombo.csv};
    \end{semilogyaxis}
  \end{tikzpicture}
  \caption{\(C_{\text{adm}}\)}
  \vspace*{20mm}
  \end{subfigure}
\begin{subfigure}[t]{0.5\textwidth}
  \centering
  \begin{tikzpicture}[scale=0.8]
    \begin{semilogyaxis}
      [ymin=5*10^-3,ymax=5*10^3,
      xmin=1, xmax=15, 
      xlabel=$\ell$,ylabel=time in seconds, 
    ]
    \addplot[mark=x,mark options={indigo,solid},indigo] table[x=l,y=d1,col sep=comma] {koos+kiestijden/_4,_4,_4_-stupidcombo.csv};
    \addplot[mark=triangle,mark options={indigo,solid},indigo] table[x=l,y=d1,col sep=comma] {koos+kiestijden/_4,_4,_4_-simplifycombo.csv};
    \addplot[mark=square,mark options={indigo,solid},indigo] table[x=l,y=d1,col sep=comma] {koos+kiestijden/_4,_4,_4_-xsimplifycombo.csv};
    \end{semilogyaxis}
  \end{tikzpicture}
  \caption{\(C_{\text{imp}}\)}\vspace*{20mm}
  \end{subfigure}
\caption{Total time in seconds to `construct' a \(\G_{\times}\), \(\G_{\triangle}\) or \(\G_{\square}\) and subsequently use it to evaluate the natural extension for a single option set, as a function of the number \(\ell\) of pairs \((V,W)\) in the assessment. The vertical axis is logarithmic, the experiments are averaged over 7 separate experiments (with for each experiment 7 option sets to choose from) and the lines for each method stop when the average was over 50 minutes.}
\vspace*{10mm}
\label{fig:somtijd}
\end{figure}
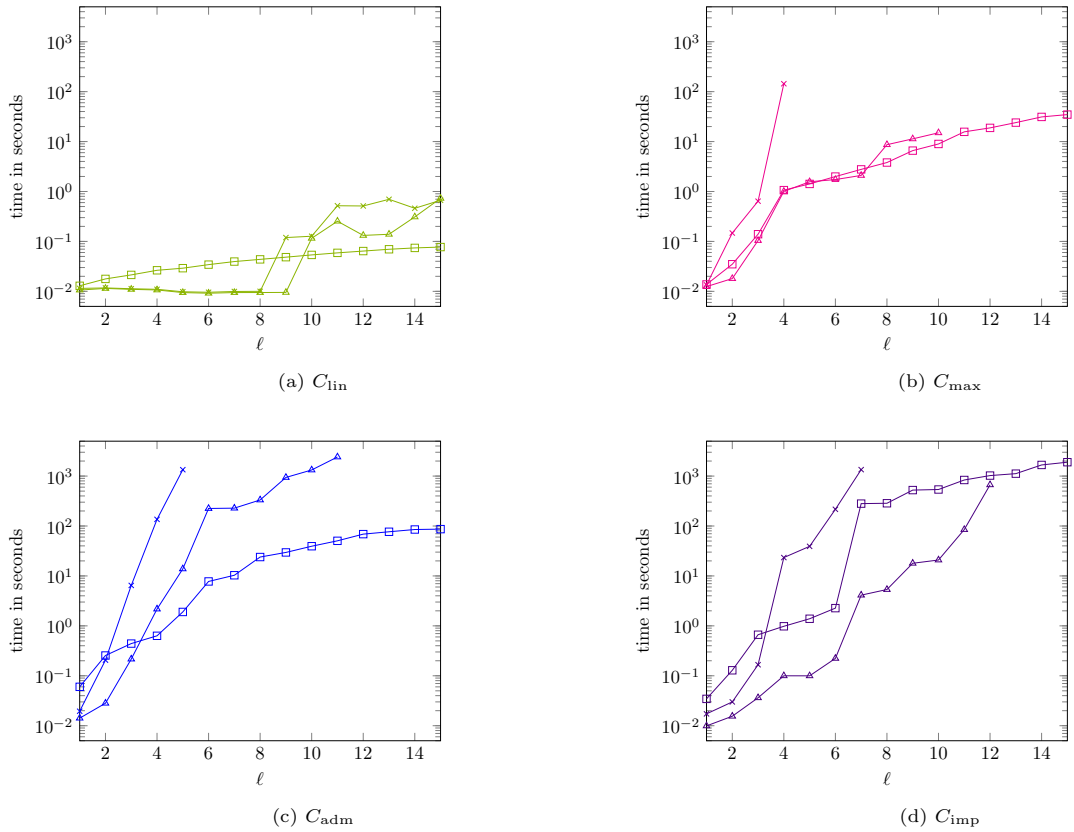

\section{Conclusion and future work}\label{sec:conc}

The main conclusion of this work is that choice functions provide a principled viable framework for inferring new decisions from previous ones.
The two key concepts that we introduced to achieve this were consistency and natural extension.
The former allows one to check if there is at least one preference order that is compatible with an assessment, while the latter allows one to infer new choices based on the assessment by considering all compatible orders.
From a practical point of view, our main contributions are algorithms that are able to execute these tasks.
We have also developed effective methods to simplify the information in the assessment, while still retaining all decisive power.
To demonstrate the performance when including these simplifications, we carried out experiments that measure the space and time efficiency under various scenarios.
Our results show that our simplifications allow for larger assessments to be dealt with, and this for various levels of imprecision.
Related to this, we saw that the third method---which implements all the simplifications---outperforms the other two methods in terms of efficiency for larger assessments.
For smaller assessments, the second method is the most efficient, unless one has to evaluate the natural extension for multiple option sets, in which case the third method eventually becomes faster.
A particularly intriguing observation was that for assessments that are generated with maximality or E-admissibility, the generator of the third method even seemed to stabilise, as seen in \cref{fig:compchose2}.
This is intriguing because it could mean that it is converging towards the `real' model---the model that we used to make the assessment---through simple approximate models.
It would be interesting to investigate further if this trend continues for even larger assessments, and explain why it happens.


Future work could also add onto \cref{sec:simp} by trying to obtain a `simplest' generator for any given assessment, or showing that our methods already achieves this.
Also, how often and in what order one has to apply the simplifications can still be optimised.
Approaches in between method 2 and 3 could for example be considered.
Another direction could be to analyse how efficiency scales with other parameters, both theoretically and experimentally.
This includes the size of the option sets in the assessment, the dimension of the vector space $\V$ and the size of the option set $A$ for which we want to evaluate the natural extension.
One could also consider  alternative forms of assessments, such as bounds on probabilities, bounds on expectations and preference statements, and show how they can be made to fit in our choice functions framework.
Finally, it would be nice to evaluate how our methods perform when applied to real-life decision problems.
\subsubsection*{Acknowledgements}
This work was partially supported by Ghent University, through Jasper De Bock's BOF Starting Grant “Rational decision-making under uncertainty: a new paradigm based on choice functions”, number 01N04819. 
%
%
%
%

\bibliography{ArneDecadt}

\begin{thebibliography}{18}
\expandafter\ifx\csname natexlab\endcsname\relax\def\natexlab#1{#1}\fi
\providecommand{\url}[1]{\texttt{#1}}
\providecommand{\href}[2]{#2}
\providecommand{\path}[1]{#1}
\providecommand{\DOIprefix}{doi:}
\providecommand{\ArXivprefix}{arXiv:}
\providecommand{\URLprefix}{URL: }
\providecommand{\Pubmedprefix}{pmid:}
\providecommand{\doi}[1]{\href{http://dx.doi.org/#1}{\path{#1}}}
\providecommand{\Pubmed}[1]{\href{pmid:#1}{\path{#1}}}
\providecommand{\bibinfo}[2]{#2}
\ifx\xfnm\relax \def\xfnm[#1]{\unskip,\space#1}\fi
\bibitem[{Troffaes(2007)}]{troffaes2007decision}
\bibinfo{author}{M.~C.~M. Troffaes},
\newblock \bibinfo{title}{Decision making under uncertainty using imprecise probabilities},
\newblock \bibinfo{journal}{International Journal of Approximate Reasoning} \bibinfo{volume}{45} (\bibinfo{year}{2007}) \bibinfo{pages}{17–29}.
\bibitem[{{De Bock, Jasper}(2020)}]{de2020archimedean}
\bibinfo{author}{{De Bock, Jasper}},
\newblock \bibinfo{title}{{Archimedean choice functions: an axiomatic foundation for imprecise decision making}},
\newblock in: \bibinfo{booktitle}{{Information Processing and Management of Uncertainty in Knowledge-Based Systems}}, volume \bibinfo{volume}{{1238}}, \bibinfo{publisher}{{Springer}}, \bibinfo{year}{{2020}}, pp. \bibinfo{pages}{{195--209}}.
\bibitem[{De~Bock and De~Cooman(2019)}]{de2019interpreting}
\bibinfo{author}{J.~De~Bock}, \bibinfo{author}{G.~De~Cooman},
\newblock \bibinfo{title}{Interpreting, axiomatising and representing coherent choice functions in terms of desirability},
\newblock in: \bibinfo{booktitle}{Proceedings of the eleventh International Symposium on Imprecise Probabilities: Theories and Applications}, \bibinfo{organization}{PMLR}, \bibinfo{year}{2019}, pp. \bibinfo{pages}{125--134}.
\bibitem[{Seidenfeld et~al.(2010)Seidenfeld, Schervish, and Kadane}]{seidenfeld2010coherent}
\bibinfo{author}{T.~Seidenfeld}, \bibinfo{author}{M.~J. Schervish}, \bibinfo{author}{J.~B. Kadane},
\newblock \bibinfo{title}{Coherent choice functions under uncertainty},
\newblock \bibinfo{journal}{Synthese} \bibinfo{volume}{172} (\bibinfo{year}{2010}) \bibinfo{pages}{157}.
\bibitem[{Van~Camp et~al.(2018)Van~Camp, Miranda, and De~Cooman}]{van2018natural}
\bibinfo{author}{A.~Van~Camp}, \bibinfo{author}{E.~Miranda}, \bibinfo{author}{G.~De~Cooman},
\newblock \bibinfo{title}{Natural extension of choice functions},
\newblock in: \bibinfo{booktitle}{Proceedings of the seventeenth International Conference on Information Processing and Management of Uncertainty in Knowledge-Based Systems}, \bibinfo{organization}{Springer}, \bibinfo{year}{2018}, pp. \bibinfo{pages}{201--213}.
\bibitem[{Sen(1971)}]{sen1971choice}
\bibinfo{author}{A.~K. Sen},
\newblock \bibinfo{title}{Choice functions and revealed preference},
\newblock \bibinfo{journal}{The Review of Economic Studies} \bibinfo{volume}{38} (\bibinfo{year}{1971}) \bibinfo{pages}{307--317}.
\bibitem[{Sen(1977)}]{sen1977social}
\bibinfo{author}{A.~Sen},
\newblock \bibinfo{title}{Social choice theory: A re-examination},
\newblock \bibinfo{journal}{Econometrica: journal of the Econometric Society}  (\bibinfo{year}{1977}) \bibinfo{pages}{53--89}.
\bibitem[{Birkhoff(1940)}]{birkhoff1940lattice}
\bibinfo{author}{G.~Birkhoff}, \bibinfo{title}{Lattice theory}, \bibinfo{edition}{third} ed., \bibinfo{publisher}{American Mathematical Society}, \bibinfo{year}{1940}.
\bibitem[{Augustin et~al.(2014)Augustin, Coolen, De~Cooman, and Troffaes}]{augustin2014introduction}
\bibinfo{author}{T.~Augustin}, \bibinfo{author}{F.~P. Coolen}, \bibinfo{author}{G.~De~Cooman}, \bibinfo{author}{M.~C. Troffaes}, \bibinfo{title}{Introduction to Imprecise Probabilities}, \bibinfo{publisher}{John Wiley \& Sons}, \bibinfo{year}{2014}.
\bibitem[{Quaeghebeur(2013)}]{quaeghebeur2013conestrip}
\bibinfo{author}{E.~Quaeghebeur},
\newblock \bibinfo{title}{The {CONEstrip} algorithm},
\newblock in: \bibinfo{booktitle}{Synergies of Soft Computing and Statistics for Intelligent Data Analysis}, \bibinfo{publisher}{Springer}, \bibinfo{year}{2013}, pp. \bibinfo{pages}{45--54}.
\bibitem[{Matou\v{s}ek and G{\"a}rtner(2006)}]{linprog2006}
\bibinfo{author}{J.~Matou\v{s}ek}, \bibinfo{author}{B.~G{\"a}rtner}, \bibinfo{title}{Understanding and Using Linear Programming}, Universitext, \bibinfo{publisher}{Springer}, \bibinfo{year}{2006}.
\bibitem[{Vaidya(1989)}]{vaidya1989speeding}
\bibinfo{author}{P.~M. Vaidya},
\newblock \bibinfo{title}{Speeding-up linear programming using fast matrix multiplication},
\newblock in: \bibinfo{booktitle}{30th annual symposium on foundations of computer science}, \bibinfo{organization}{IEEE Computer Society}, \bibinfo{year}{1989}, pp. \bibinfo{pages}{332--337}.
\bibitem[{Morgado(1962)}]{morgado1962characterization}
\bibinfo{author}{J.~Morgado},
\newblock \bibinfo{title}{A characterization of the closure operators by means of one axiom},
\newblock \bibinfo{journal}{Portugaliae mathematica} \bibinfo{volume}{21} (\bibinfo{year}{1962}) \bibinfo{pages}{155--156}.
\bibitem[{Fukuda(2020)}]{fukuda2020polyhedral}
\bibinfo{author}{K.~Fukuda}, \bibinfo{title}{Polyhedral computation}, \bibinfo{publisher}{Department of Mathematics, Institute of Theoretical Computer Science ETH Zurich}, \bibinfo{year}{2020}.
\bibitem[{Legat(2023)}]{legat2023polyhedral}
\bibinfo{author}{B.~Legat},
\newblock \bibinfo{title}{Polyhedral computation},
\newblock in: \bibinfo{booktitle}{JuliaCon}, \bibinfo{year}{2023}. \URLprefix \url{https://pretalx.com/juliacon2023/talk/JP3SPX/}.
\bibitem[{Barber et~al.(1996)Barber, Dobkin, and Huhdanpaa}]{barber1996quickhull}
\bibinfo{author}{C.~B. Barber}, \bibinfo{author}{D.~P. Dobkin}, \bibinfo{author}{H.~Huhdanpaa},
\newblock \bibinfo{title}{The quickhull algorithm for convex hulls},
\newblock \bibinfo{journal}{ACM Trans. Math. Softw.} \bibinfo{volume}{22} (\bibinfo{year}{1996}) \bibinfo{pages}{469–483}.
\bibitem[{Fukuda and Prodon(1996)}]{DoubleDes}
\bibinfo{author}{K.~Fukuda}, \bibinfo{author}{A.~Prodon},
\newblock \bibinfo{title}{Double description method revisited},
\newblock in: \bibinfo{booktitle}{Combinatorics and Computer Science}, \bibinfo{publisher}{Springer Berlin Heidelberg}, \bibinfo{address}{Berlin, Heidelberg}, \bibinfo{year}{1996}, pp. \bibinfo{pages}{91--111}.
\bibitem[{Nakharutai et~al.(2018)Nakharutai, Troffaes, and Caiado}]{nakharutai2018improved}
\bibinfo{author}{N.~Nakharutai}, \bibinfo{author}{M.~C.~M. Troffaes}, \bibinfo{author}{C.~C. Caiado},
\newblock \bibinfo{title}{Improved linear programming methods for checking avoiding sure loss},
\newblock \bibinfo{journal}{International Journal of Approximate Reasoning} \bibinfo{volume}{101} (\bibinfo{year}{2018}) \bibinfo{pages}{293--310}.

\end{thebibliography}
\newpage
\appendix



\section{Equivalence of axioms}\label{appsec:equivAx}
\renewcommand{\thesection}{\Alph{section}}
In this appendix we prove the equivalence of our characterisation of coherence and the axiomatic characterisation of \citet{de2019interpreting}.
In particular, with \(\R^2_{>0}\coloneqq \{ (\lambda,\nu)\in \R^2 \colon (\lambda,\nu)>0\} \), we consider the following axioms for a rejection function \(R\colon \Q\to \Q\):
for all \(A,B\in \Q\), \(u\in A\) and \((\lambda_1,\lambda_2)\colon A\times B \to \R^2_{>0}\):
  \begin{enumerate}
    \item[${\mathrm{R}}_0$.] \(u \in {R}(A)\) if and only if \(0\in {R}(A-u)\);
    \item[${\mathrm{R}}_1$.] \({R}(A)\neq A\);
    \item[${\mathrm{R}}_2$.] if \(u>0\) then \(0\in {R}(\{0,u\})\);
    \item[${\mathrm{R}}_3$.] if \(0\in {R}(A\cup\{0\})\) and  \(0\in {R}(B\cup\{0\})\) then \(0\in {R}(\{\lambda_1(v,w) v +\lambda_2(v,w) w \colon v\in A,w\in B\}\cup \{0\});\)
    \item[${\mathrm{R}}_4$.] if \(A\subseteq B\) then \({R}(A)\subseteq {R}(B)\).
  \end{enumerate}
Our main result is then the following.
\begin{theorem}\label{th:hoofdstellingAppendixEchtEcht}
  A choice function \(C\colon \Q\to \Q\) is coherent if and only if its corresponding rejection function \(R_C\) satisfies ${\mathrm{R}}_0$--${\mathrm{R}}_4$.
\end{theorem}
A subtle point is that these are not exactly the axioms used by \citet{de2019interpreting}, as they also add \(\emptyset\) to the domain of their choice and rejection functions.
We will however see that this is a mere technicality with no substantial effect.

To be able to use the results of \citet{de2019interpreting} in our proof for \cref{th:hoofdstellingAppendixEchtEcht}, we too will consider choice and rejection functions whose domain includes \(\emptyset\) and call it them \emph{extended choice and rejection functions}: a map \(\hat{C}\colon \Qe\to\Qe\) is called an extended choice function if it satisfies \(\hat{C}(A)\subseteq A\) for all \(A\in \Qe\).
With any such extended choice function \(\hat{C}\) we also associate a corresponding extended rejection function \(\hat{R}_{\hat{C}}\colon \Qe\to \Qe\), defined by \(\hat{R}_{\hat{C}}(A)=A\setminus \hat{C}(A)\) for all \(A\in \Qe\).
Since we must have that \(\hat{C}(\emptyset)\subseteq \emptyset\), we always have that \(\hat{C}(\emptyset)=\emptyset\) and \(\hat{R}(\emptyset)=\emptyset\), which makes this part of the domain uninteresting.

With any choice function \(C\colon \Q\to\Q\) we associate the extended choice function
\begin{equation}\label{eq:RextDef}
\overline{C} \colon \Qe\to\Qe\colon A \mapsto \begin{cases} C(A) &\text{if }A\neq \emptyset,\\
\emptyset &\text{if }A=\emptyset.
\end{cases}
\end{equation}
This link between extended choice functions and choice functions is bijective because \(\hat{C}(\emptyset)=\emptyset\) for every extended choice function \(\hat{C}\).
For any extended rejection function \(\hat{R}\colon \Qe\to \Qe\), \citet{de2019interpreting} consider the following axioms:
for all \(A,B\in \Qe\), \(u\in A\) and \((\lambda_1,\lambda_2)\colon A\times B \to \R^2_{>0}\):
  \begin{enumerate}
    \item[$\hat{\mathrm{R}}_0$.] \(u \in \hat{R}(A)\) if and only if \(0\in \hat{R}(A-u)\);
    \item[$\hat{\mathrm{R}}_1$.] \(\hat{R}(\emptyset)=\emptyset\) and if \(A\neq \emptyset\) then \(\hat{R}(A)\neq A\);
    \item[$\hat{\mathrm{R}}_2$.] if \(u>0\) then \(0\in \hat{R}(\{0,u\})\);
    \item[$\hat{\mathrm{R}}_3$.] if \(0\in \hat{R}(A\cup\{0\})\) and  \(0\in \hat{R}(B\cup\{0\})\) then \(0\in \hat{R}(\{\lambda_1(v,w) v +\lambda_2(v,w) w \colon v\in A,w\in B\}\cup \{0\});\)
    \item[$\hat{\mathrm{R}}_4$.] if \(A\subseteq B\) then \(\hat{R}(A)\subseteq \hat{R}(B)\).
  \end{enumerate}
  These axioms are very similar to \(\mathrm{R}_0\)--\(\mathrm{R}_4\); the only difference is that they are also imposed for \(A=\emptyset\) and/or \(B=\emptyset\), and that \(\hat{\mathrm{R}}_1\) adds the condition \(\hat{R}(\emptyset)=\emptyset\).
  Due to this similarity, it should not come as a surprise that imposing \(\mathrm{R}_0\)--\(\mathrm{R}_4\) on the rejection function \(R_C\) that corresponds to a choice function \(C\) is equivalent to imposing \(\hat{\mathrm{R}}_0\)--\(\hat{\mathrm{R}}_4\) on the extended rejection function \(\hat{R}_{\overline{C}}\) that corresponds to its extension \(\overline{C}\).
\begin{lemma}\label{lem:equivRhatR}
For any choice function \(C \colon \Q\to \Q\) and \(k\in \{0,1,...,4\}\) we have that \(R_C\) satisfies  ${\mathrm{R}}_k$ if and only if \(\hat{R}_{\overline{C}}\) satisfies  $\hat{\mathrm{R}}_k$.
\end{lemma}
\begin{proof}
It follows immediately from the definitions that for any \(A\in \Q\), \(R_C(A)=\hat{R}_{\overline{C}}(A)\).
Therefore, the implication to the left is immediate.
So now we prove that when at least one of \(A\) and \(B\) are empty, the axioms still hold true.
For \(k=0\), $\hat{\mathrm{R}}_0$ is trivially true for \(A=\emptyset\) as there are no \(u\in A\) then.
For \(k=1\), $\hat{\mathrm{R}}_1$ is true because \(\hat{R}_{\overline{C}}(\emptyset)=\emptyset\) by definition.
\(k=2\) is unrelated to \(A\) and \(B\), so trivially true.
For \(k=3\), $\hat{\mathrm{R}}_3$ is true because if \(A\) and/or \(B\) are empty, then we have \(0\in \hat{R}_{\overline{C}}(\{0\})\) from the antecedent and this is also the implication.
Finally, for \(k=4\), $\hat{\mathrm{R}}_4$ is true because the case where \(A=\emptyset\subseteq B\), for any \(B\in \Qe\), implies that \(\hat{R}_{\overline{C}}(A)=\emptyset \subseteq \hat{R}_{\overline{C}}(B)\) and the case where \(B=\emptyset\) but \(A\neq\emptyset\) is impossible. 
\end{proof}
Similarly to how we associated a choice function \(C_{\OO}\) with any set of preference orders \(\OO\) in \cref{eq:oorsprdef}, we can also associate an extended choice function \(\hat{C}_{\OO}\) with such a set~\(\OO\), by letting
\begin{equation}\label{eq:Rdef2}
  \hat{C}_{\OO}(A)=\{u\in A\colon (\exists \prec \in \OO)(\forall a \in A) u\not\prec a\}\quad\text{ for all }\quad A\in \Qe.
\end{equation}
We call an extended choice function coherent if and only if it is of this form, with \(\OO\) non-empty.
\begin{definition}
  An extended choice function \(\hat{C}\) is coherent if and only if there is a non-empty set of preference orders \(\OO\subseteq \OOO\) such that \(\hat{C}=\hat{C}_{\OO}\).
\end{definition}
\begin{lemma}\label{lem:equivRhatRcoh}
A choice function \(C\) is coherent if and only if \(\overline{C}\) is coherent. 
\end{lemma}
\begin{proof}
The implication to the left is immediate as the definition of coherence is the same for both and \(C\) has a smaller domain than \(\overline{C}\).
For the implication to the right, assume that \(C\) is coherent.
By definition of coherence, there is a set of preference orders \(\OO\) such that \(C=C_{\OO}\).
Then for all \(A\in \Q\) we have \(\overline{C}(A)=C(A)=C_{\OO}(A)=\hat{C}_{\OO}(A)\).
For \(A=\emptyset\) we also have trivially that \(\hat{C}_{\OO}(\emptyset)=\emptyset=\overline{C}(\emptyset)\).
\end{proof}

Before we get to the proof of \cref{th:hoofdstellingAppendixEchtEcht}, we now first use the results of \citet{de2019interpreting} to establish a similar result for extended choice functions.
\begin{proposition}\label{th:hoofdstellingAppendixEcht}
  An extended choice function \(\hat{C}\colon \Qe\to \Qe\) is coherent if and only if its corresponding extended rejection function \(\hat{R}_{\hat{C}}\)  satisfies $\hat{\mathrm{R}}_0$--$\hat{\mathrm{R}}_4$.
\end{proposition}

Our proof makes use of the following technical lemmata.

\begin{lemma}\label{lem:verschuivenR}
  Consider any set of preference orders \(\OO\subseteq \OOO\).
  Then for all \(A\in \Qe\), 
   \(
    0\in \hat{R}_{\hat{C}_\OO}(A-u) \Leftrightarrow u\in \hat{R}_{\hat{C}_\OO}(A)
   \).
\end{lemma}
\begin{proof}
  \(u\in \hat{R}_{\hat{C}_\OO}(A)\) is by definition equivalent to \((\forall \prec \in \OO)(\exists a \in A) u\prec a\).
  This is by \cref{ax:Otransla} equivalent to \((\forall \prec \in \OO)(\exists v \in A-u) 0\prec v\), which is equivalent to \(0\in \hat{R}_{\hat{C}_\OO}(A-u)\).
\end{proof}
\begin{lemma}\label{lem:extendedRejectionFunction}
Consider an extended rejection function \(\hat{R}\colon \Qe\to\Qe\) that satisfies $\hat{\mathrm{R}}_0$ and let \(K\coloneqq \{ A\in \Qe \colon 0\in \hat{R}(A\cup \{0\})\}\).
Then 
\begin{equation}\label{eq:Kdef}
  (\forall u \in \V)(\forall A\in \Qe)\; u \in \hat{R}(A\cup \{u\}) \Leftrightarrow A-u\in K.  
\end{equation}
\end{lemma}
\begin{proof}
  \(A-u\in K\) is equivalent to \(0\in \hat{R}((A-u)\cup \{0\})=\hat{R}((A\cup\{u\})-u).\)
  By $\hat{\mathrm{R}}_0$ this is equivalent to \(u\in \hat{R}(A\cup\{u\})\).
\end{proof}

\begin{proofof}{\cref{th:hoofdstellingAppendixEcht}}
  First we prove the implication to the right.
  Consider any non-empty set \(\OO\subseteq \OOO\).
  We need to prove that \(\hat{R}_{\hat{C}_\OO}\) satisfies $\hat{\mathrm{R}}_0$ to $\hat{\mathrm{R}}_4$.
  Axiom $\hat{\mathrm{R}}_0$ follows immediately from \cref{lem:verschuivenR}. 
  Since \(\hat{R}_{\hat{C}_\OO}(\emptyset)=\emptyset\) follows from Definition~(\ref{eq:Rdef2}) and, for any \(A\in \Q\), we have \(\hat{R}_{\hat{C}_\OO}(A)=A\setminus \hat{C}_{\OO}(A)\), $\hat{\mathrm{R}}_1$ follows immediately from \cref{prop:niet-leeg}.
  $\hat{\mathrm{R}}_2$ follows immediately from \cref{ax:Oext} and Definition~(\ref{eq:Rdef2}).
  For $\hat{\mathrm{R}}_3$, take any \(\prec\in \OO\).
  From Definition~(\ref{eq:Rdef2}) and \cref{ax:Oantisym}, we know that there are \(a\in A\) such that \(0\prec a\) and \(b\in B\) such that \(0\prec b\).
  Without loss of generality, assume that \( \lambda_1(a,b)>0\).
  By \cref{ax:Oscal}, we have \(0\prec \lambda_1(a,b) a\) and either \(\lambda_2(a,b) b=0\) or \(0\prec \lambda_2(a,b) b\).
  In the former case we find that \(0\prec \lambda_1(a,b) a =\lambda_1(a,b) a+\lambda_2(a,b) b\). 
  In the latter case, it follows from \cref{ax:Otransla} that  \( \lambda_1(a,b) a\prec\lambda_1(a,b) a+\lambda_2(a,b) b\) and therefore, since \(0\prec \lambda_1(a,b) a\), from \cref{ax:Otransi} that \(0\prec \lambda_1(a,b) a+\lambda_2(a,b) b\).
  Hence, in both cases, this implies that \(0\prec \lambda_1(a,b) a+\lambda_2(a,b) b\).
  Since \(\lambda_1(a,b) a+\lambda_2(a,b) b\in \{\lambda_1(v,w) v +\lambda_2(v,w) w \colon v\in A,w\in B\}\cup \{0\}\) and \(\prec \in \OO\) was arbitrary, it therefore follows from Definition~(\ref{eq:Rdef2}) that 
  \[0\in \hat{R}_{\hat{C}_\OO}(\{\lambda_1(v,w) v +\lambda_2(v,w) w \colon v\in A,w\in B\}\cup \{0\}).\]
  For $\hat{\mathrm{R}}_4$, 
  if \(u\in \hat{R}_{\hat{C}_\OO}(A)\), then for all \(\prec \in \OO\) there is some \(a\in A\subseteq B\) such that \(u\prec a\), whence also \(u\in \hat{R}_{\hat{C}_\OO}(B)\).

  Next we prove the implication to the left making heavy use of the results in  \cite{de2019interpreting}.
  In Ref. \cite{de2019interpreting}, the main tool are so-called `coherent sets of desirable option sets' \(K\subseteq \Qe\), where coherence is characterised through a number of axioms \cite[Definition~2]{de2019interpreting} that are not important for our purposes.
  Since \(\hat{R}_{\hat{C}}\) satisfies $\hat{\mathrm{R}}_0$, we know from \cref{lem:extendedRejectionFunction} that \(K\coloneqq \{ A\in \Qe \colon 0\in \hat{R}_{\hat{C}}(A\cup \{0\})\}\) satisfies \cref{eq:Kdef}. 
  Since \(\hat{R}_{\hat{C}}\) satisfies \(\hat{\mathrm{R}}_0\)--\(\hat{\mathrm{R}}_4\), it therefore  follows from \cite[Proposition~4]{de2019interpreting} that \(K\) is a `coherent set of desirable option sets'. 
  Therefore, it follows from \cite[Theorem~9]{de2019interpreting} that there is a non-empty set \(\mathscr{D}\subseteq \overline{\mathrm{\mathbf{G}}}\) of coherent sets of desirable options such that 
  \[K=\bigcap_{G\in \mathscr{D}}\{A\in \Qe\colon A\cap G \neq \emptyset\}.\]
  Let \(\OO\coloneqq\{\prec_G\colon G\in \mathscr{D}\}\) and recall from \cref{lem:axD} that this is a set of preference orders, which is furthermore non-empty because \(\mathscr{D}\) is.
  Observe also that, for all \(G\in \mathscr{D}\), 
  \[\{A\in \Qe\colon A\cap G \neq \emptyset\}=\{A\in \Qe \colon (\exists a \in A) 0\prec_G a\}\] by definition of \(\prec_G\).
  We then see that
  \begin{align*}
    K&=\bigcap_{G\in \mathscr{D}} \{A\in \Qe \colon (\exists a \in A) 0\prec_G a\}
    =\{A\in \Qe \colon (\forall G\in \mathscr{D})(\exists a\in A) 0\prec_G a\} \\
    &= \{A\in \Qe \colon (\forall \prec \in \OO) (\exists a\in A) 0\prec a\} 
    = \{A\in \Qe \colon (\forall \prec \in \OO) (\exists a\in A\cup \{0\}) 0\prec a\} \\
    &= \{A\in \Qe \colon 0\in \hat{R}_{\hat{C}_\OO}(A\cup \{0\})\},
  \end{align*}
  where the second to last equality follows from \cref{ax:Oantisym} and the last from \cref{eq:Rdef2}.
  By the definition of \(K\), this implies that for any \(A\in \Qe\), \(0\in \hat{R}_{\hat{C}}(A\cup\{0\})\) if and only if \(0\in \hat{R}_{\hat{C}_\OO}(A\cup \{0\}) \).
  Now, by $\hat{\mathrm{R}}_0$, the previous observation and \cref{lem:verschuivenR} we have for any \(A\in \Qe\) and \(u\in A\) that
  \begin{align*}
    u\in \hat{R}_{\hat{C}}(A) &\Leftrightarrow 0\in \hat{R}_{\hat{C}}(A-u)\Leftrightarrow 0\in \hat{R}_{\hat{C}}((A-u)\cup\{0\})\\ & \Leftrightarrow 0\in \hat{R}_{\hat{C}_\OO}((A-u)\cup\{0\}) \Leftrightarrow 0\in \hat{R}_{\hat{C}_\OO}(A-u) \Leftrightarrow u\in \hat{R}_{\hat{C}_\OO}(A),
  \end{align*}
  since \(0\in A-u = (A-u)\cup \{0\}\).
    Therefore, \(\hat{R}_{\hat{C}}= \hat{R}_{\hat{C}_\OO}\).
\end{proofof}

\begin{proofof}{\cref{th:hoofdstellingAppendixEchtEcht}}
By \cref{lem:equivRhatRcoh}, we have that coherence of \(C\) is equivalent to coherence of \(\overline{C}\).
By \cref{th:hoofdstellingAppendixEcht}, this is equivalent to \(\hat{R}_{\overline{C}}\) satisfying $\hat{\mathrm{R}}_0$--$\hat{\mathrm{R}}_4$.
Finally, by \cref{lem:equivRhatR}, this is equivalent to \(R_C\) satisfying ${\mathrm{R}}_0$--${\mathrm{R}}_4$.
\end{proofof}

\renewcommand{\thesection}{Appendix \Alph{section}}

\end{document}